%% file: main.tex
\documentclass[twocolumn]{aastex631}

\usepackage[T1]{fontenc}

\newcommand{\newtwo}[1]{{\color{black} #1}}

\newcommand{\tatiana}[1]{{\color{black} #1}}
\newcommand{\round}[1]{{\color{black} #1}}

\newcommand{\diff}{{\it diff}}

\newcommand{\nodia}{{\it noDIA-based}}
\newcommand{\diabased}{{\it DIA-based}}
\newcommand{\temp}{{\it tmpl}}
\newcommand{\search}{{\it srch}}

\newcommand{\new}[1]{{\color{black} #1}}
\newcommand{\eg}{{\it e.g.}}
\newcommand{\ie}{{\it i.e.}}

\newcommand{\idiff}{\ensuremath{I_\mathrm{diff}}}
\newcommand{\itempl}{\ensuremath{I_\mathrm{tmpl}}}
\newcommand{\isearch}{\ensuremath{I_\mathrm{srch}}}

\usepackage{amsmath}	% Advanced maths commands
\usepackage{amssymb}	% Extra maths symbols
\usepackage{mathtools}
\usepackage{subfigure}
\usepackage{natbib}
\usepackage{multirow}
\usepackage{tikz}
\usetikzlibrary{shapes.geometric, arrows}

\usepackage{hyperref}

\usepackage{newtxtext,newtxmath}
\usepackage{graphicx}%Include figure files
\usepackage{mathtools}
\usepackage{ulem}
\usepackage{scrextend}
\usepackage{chngcntr}
\definecolor{mygreen}{RGB}{25, 196, 139}
\shorttitle{noDIA}
\shortauthors{TAC et al.}
\graphicspath{{./}{figures/}}
\begin{document}

% \title{There's no difference: The potential for Convolutional Neural Networks for transient detection without template subtraction}

\title{What's the Difference? The potential for Convolutional Neural Networks for transient detection without template subtraction}

\author{Tatiana Acero-Cuellar}
\affiliation{University of Delaware,
Department of Physics and Astronomy,
Newark, DE USA}
\affiliation{Universidad Nacional de Colombia,
Observatorio Astronómio Nacional,
Bogotá, Colombia}

\author{Federica Bianco}
\affiliation{University of Delaware,
Department of Physics and Astronomy,
Newark, DE USA}
\affiliation{
University of Delaware,
Joseph R. Biden, Jr. School of Public Policy and Administration, 
 Newark, DE USA}
\affiliation{
University of Delaware,
Data Science Institute,
Newark, DE, USA}
\affiliation{
Vera C. Rubin Observatory, Tucson, AZ, USA}
\author{Gregory Dobler}

\affiliation{
University of Delaware,
Joseph R. Biden, Jr. School of Public Policy and Administration, 
Newark, DE USA}
\affiliation{University of Delaware,
Department of Physics and Astronomy,
Newark, DE USA}
\affiliation{
University of Delaware,
Data Science Institute,
Newark, DE, USA}

\author{Masao Sako}
\affiliation{
Department of Physics and Astronomy, University of Pennsylvania, Philadelphia, PA, USA}

\author{Helen Qu}
\affiliation{
Department of Physics and Astronomy, University of Pennsylvania, Philadelphia, PA, USA}

\collaboration{6}{and The LSST Dark Energy Science Collaboration}

% \author{Butler Burton}
% \affiliation{Leiden University}
% \affiliation{AAS Journals Associate Editor-in-Chief}

% \author{Amy Hendrickson}
% \altaffiliation{AASTeX v6+ programmer}
% \affiliation{TeXnology Inc.}

% \author{Julie Steffen}
% \affiliation{AAS Director of Publishing}
% \affiliation{American Astronomical Society \\
% 1667 K Street NW, Suite 800 \\
% Washington, DC 20006, USA}

% \author{Magaret Donnelly}
% \affiliation{IOP Publishing, Washington, DC 20005}



\begin{abstract}
We present a \new{study of the potential for} Convolutional Neural Networks (CNNs) to enable separation of astrophysical transients from image artifacts, a task known as ``real-bogus'' classification without requiring \new{a template subtracted (or difference) image which requires a computationally expensive process to generate}, involving image matching on small spatial scales in large volumes of data. \new{Using data from the Dark Energy Survey, we explore the use of CNNs to (1) automate the ``real-bogus'' classification, (2) reduce the computational costs of transient discovery. We compare the efficiency of two CNNs with similar architectures, one that uses ``image triplets'' (templates, search, and difference image)} and one that takes as input the template and search only. We measure the decrease in efficiency associated with the {loss of information} in input finding that the testing accuracy is reduced from $\sim96\%$ to $\sim91.1\%$. We further investigate how \new{the latter model} learns the required information from the template and search by exploring the saliency maps. Our work (1) \new{confirms that CNNs are excellent models for ``real-bogus'' classification that rely exclusively on the imaging data and require no feature engineering task; (2) demonstrates that high-accuracy (>90\%) models can be built without the need to construct difference images, but some accuracy is lost}. Since once trained, neural networks can generate predictions at minimal computational costs, we argue that future implementations of this methodology could dramatically reduce the computational costs in the detection of transients in synoptic surveys like Rubin Observatory’s Legacy Survey of Space and Time by bypassing the Difference Image Analysis entirely.
\end{abstract}

% \keywords{Classical Novae (251) --- Ultraviolet astronomy(1736) --- History of astronomy(1868) --- Interdisciplinary astronomy(804)}


\section{Introduction} \label{sec:intro}
\input{intro.tex}

\section{state of the art and traditional solutions to detection of transients}\label{sec:stateofart}
\input{traditionalSolutions.tex}

%\newpage
\section{Data}\label{sec:data}
\input{data.tex}

% \newpage
\section{Methodology}\label{sec:method}

\input{model.tex}
\section{Results}\label{sec:results}
\input{result.tex}

\section{Future work and limitation of this work}\label{sec:futurework}
\input{limitations.tex}
%\clearpage
\section{Conclusions}\label{sec:conclusion}

\input{conclusion.tex}

\section{Acknowledgment}\label{sec:thanks}

\input{acknowledge}

% \onecolumngrid
\bibliography{main}
\bibliographystyle{aasjournal}
% \bibliographystyle{mnras}
% \bibliography{refs} 
\appendix 
\onecolumngrid 
%\counterwithin{figure}{section}
%\counterwithin{table}{section}
\section{Preprocessing}
\subsection{Scaling astrophysical images for input to Neural Networks}
\label{sec:appendixa}
\input{appendixa.tex}

\tatiana{\subsection{Organization of the image components\\}{We have chosen to organize the three elements of the input data, \diff, \temp, and 
\search, horizontally as a 51$\times$153 input array, instead of a more traditional depth stack that makes the data shaped 51$\times$51$\times$3. We made this choice because it allows a more intuitive and clear saliency analysis. We have inspected the impact of this choice and found no  performance degradation. Here, we present a confusion matrix for a \diabased\ model with input 51$\times$51$\times$3 tensors: compared to the performance of the \diabased\ model presented in \autoref{sec:results} and \autoref{fig:confusiomatrix_models}, there is a small decrease in TN and a corresponding increase in FN rates. Similarly, we trained a \nodia\ version of this model with identical architecture except for the shape of the input layer ($51\times51\times2$) and found no significant improvement (and slightly more imbalance in the TP TN classes).}

\label{sec:appendix3channel}\input{appendix_3channel.tex}}
\clearpage
% \appendix 
% \newpage
\section{Detailed architecture of \diabased\ and \nodia\ models}
\label{sec:appendixb}

\input{appendixb.tex}
\clearpage

% \appendix 
\section{Saliency maps for various transients}

\label{sec:appendixc}

\input{appendixc.tex}

\clearpage
\newpage
\label{sec:appendix-relabel}

\input{appendix_relabel}

\end{document}

%% file: intro.tex
% \question{questions are in red}

% \new{new text is purple}

% \ggd{Greg's comments look like this}

% \greg{New text written or modified by Greg in this color violet}

Modern observational astronomy has shown us that the Universe is not static and immutable: to the contrary, it is a lively and dynamic system. We now know and understand a variety of different phenomena that can give rise to variations in the brightness and color of astrophysical objects, including explosive stellar death (supernovae, kilonovae, Gamma Ray Bursts, etc), less dramatic and powerful stellar variability (flares, pulsations), and variability arising from geometric effects (such as planetary transients and microlensing). The time scales for these phenomena range from seconds to years. When the variations are terminal –like supernovae– or stochastic –like stellar flares– they are \new{often} referred to as ``transients''. 
Detecting optical astrophysical transients characteristically requires sequences of images across a significant temporal baseline. Surveys \new{designed to study the ever-changing skies}, like the Dark Energy Survey \citep[DES]{DES} or the future Rubin Observatory Legacy Survey of Space and Time \citep[LSST]{lsst} \new{and many others}, search for spatially localized changes in brightness in patches of sky previously observed. 

Due to the rarity of astrophysical transients, a long baseline in conjunction with the observation of a large area of the sky is typically required to detect a statistically significant sample of transients like supernovae and individual examples of rare events, such as kilonovae.  The process itself is arduous and requires considerable human intervention at multiple stages. The first step is typically the creation of high quality templates of each region of the sky that is to be searched; the templates are then subtracted from nightly images, a process known in astrophysics as Difference Image Analysis (DIA) that was initially pioneered \new{by \cite{1992ApJ...399L..43C}} and \cite{Tomaney_1996} and then formalized by \cite{Alard_1998} and there is a rich history of subsequent improvements in the efficiency and accuracy of DIA models. Templates  (\temp) are typically constructed as stacks of high quality (favorable observing conditions) sky images. These high quality images must then be aligned with the ``today'' image, typically called the ``search image'' (\search), and degraded to match its Point Spread Function (PSF, which we note may vary across a single large field of view image) and scaled to match its brightness. The product generated by the subtraction of the template from the search image is the so-called ``difference image''   (\diff; see for example a description of the DIA processing pipeline for DES in \citealt{Kessler_2015}).
Transients can then be detected as clusters of adjacent pixels deviating significantly from the \new{background}.
% The current way of classifying optical images is to construct the image called template, which, in the case of  uses (READ \cite{Kessler_2015}), then the difference image is the residual obtained by subtracting the template and the science images. Many problems behind this steps to get the final classification and taking into account the large amount of data collected every day, the process of labeled objects by human inspection turns out to be very inefficient.))))\\
% Furthermore, miss-classifications due to human or computational error could lead to miss new astrophysical objects. \\
However, even the best existing DIA algorithms produce difference images with large pixel value deviation from an ideal 0-average that would be expected if no changes occurred in a patch of sky. Transients, variable stars, and moving objects will result in detections, but also a typically large number of artifacts will be detected by these thresholding schemes.

Machine learning models offer an excellent opportunity to improve the efficiency of transient detection at this stage, automating the classification between ``real'' astrophysical transients and ``bogus'' artifacts: these models are often referred to as ``Real-Bogus'' (hereafter, RB). Generally, the applications of machine learning to this problem are based on the extraction of features from the (difference) images that are then fed to models like Random Forests, k-Nearest Neighbors, or Support Vector Machines \citep{Goldstein_2015, S_nchez_2019, Mong_2020}. These models have achieved high accuracy and enabled the discovery of transients at scale in larger synoptic surveys, for example in the Palomar Transient Factory \citep[PTF][]{bloom2012automating} and Zwicky Transient Facility \citep[ZTF][]{2019PASP..131c8002M}. 

The process of engineering features for machine learning models allows the experts to embed domain knowledge in models. In the case of RB classification, the features engineered to be used by models such as the ones described above usually rely on visual inspection of a subset of the data. However, these features may overlook abstract associations between image properties that can be effective for classification, and in fact, may be biased towards human perception and theoretical expectations. An alternative approach involves the use of models that can learn features directly from the data, such as Convolutional Neural Networks \citep[CNNs]{lecun1989generalization}.  Here, it is possible to train a model using the images themselves, skipping the step of feature-design entirely.

RB CNN-based models appeared in the literature as early as 2016 \citep{Cabrera_2016}, and they are typically based on the analysis of the difference images arising from the DIA. While neural networks may be computationally demanding in the training phase, the "feed-forward" classification that arises from a pre-trained model is typically rapid and computationally light,\footnote{although the model may require a large amount of memory to cache.} leaving DIA as the computational bottle-neck in the process of astrophysical transients' detection. Yet in principle, the entirety of the information content embedded in the \diff-\temp-\search\ image triplet is also contained in the \temp-\search\ image pair.  In this paper, we explore the use of CNNs as RB models %(models that separate ``real'' astrophysical transients from ``bogus'' artifacts), 
concentrating on the potential for building high-accuracy models that \textit{do not} require the construction of difference images.  

 This paper represents a first, critical step in the process of conceptualizing and realizing a DIA-free model for astrophysical transients. Here we \new{invesigate if} Neural Networks can discriminate between astrophysical transients and artifacts without a difference image, while still relying on DIA for detection. This is the necessary premise for the complete elimination of DIA and \new{estimating the impact of the information lost by dropping the \diff\ will enable future work toward the development of} models that will not rely on DIA for detection, and that will detect and characterize transients (measure magnitude) from the \temp-\search\ image pair only.

% \subsection{Template}

% \subsection{Difference image analysis}

% \subsection{Convolutional Neural Network}

% SHOR DESCRIPTION ??

% In this paper we compare two CNN models trained on the same DES Y1 data \citep{Goldstein_2015}, one using as input data the template, search, and difference images, and the other using only the template and the search images. The main purpose is to demonstrate that relative simple CNN models are able to classify as ``real'' and ``bogus'' with high accuracy transient observations, without the necessity of the difference image.In \autoref{stateofart} we discuss the history, approaches, and details of DIA in the literature, and in \autoref{sec:data} we present the DES data that we will use to build our ``real''/``bogus'' classification models.  In \autoref{sec:results} we show the results of building a model that does not use the difference image as input and compare its performance to one that does, and we conclude with a discussion of the broader implications for this work in light of upcoming surveys in \autoref{sec:conclusion}.\\ 

This paper is organized as follows: in %\autoref{sec:intro}, the introduction, we briefly describe DIA process \new{and contextualize our work by reviewing literature examples of deep learning architectures developed for astrophysical transients' detection and characterization and the associated datasets}. In 
\autoref{stateofart} we discuss the \new{DIA history and methodology}. In \autoref{sec:data} we present the DES data that we used to build our RB classification models and the pre-processing steps. In \autoref{sec:method} \new{we discuss our methodology, illustrating the CNNs architectures used and, in \autoref{subsec: saliency}, presenting and discussing the use of saliency maps to gain insights into the models}. In \autoref{sec:results} we show the results of building a model that does not use the difference image as input, named \nodia -based model and compare its performance to one that does, named \diabased\ model;  we outline future work and conclude with a discussion of the broader implications for this result in light of upcoming surveys in \autoref{sec:futurework}.

This study is reproducible and all the code that supports the analysis presented here is available on a dedicated \texttt{GitHub} repository.\footnote{\url{https://github.com/taceroc/DIA_noDIA}.}
% in \autoref{sec:conclusion}.
\begin{figure*}

    \centering
    \includegraphics[width=0.52\linewidth]{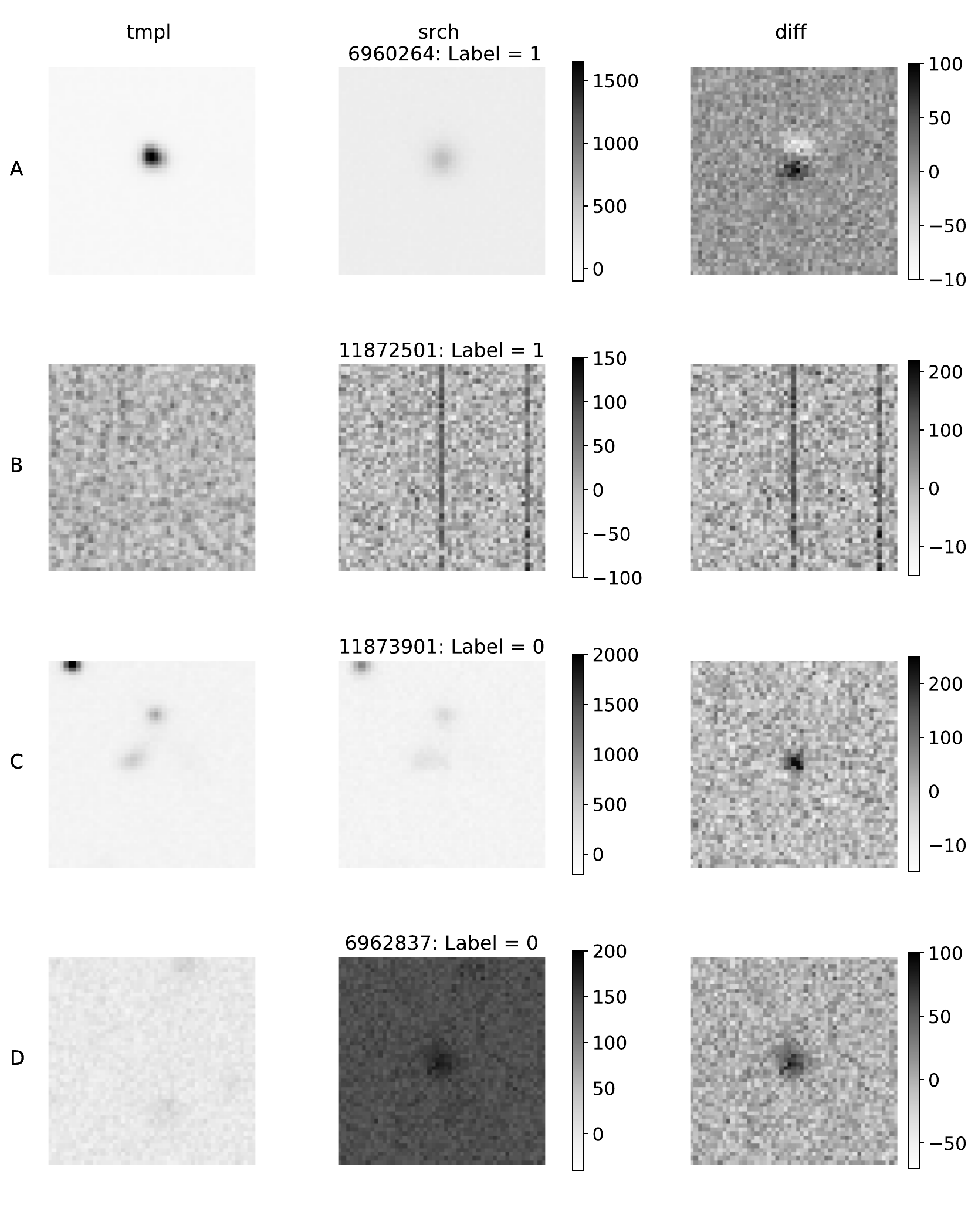}
    \caption{Example of four DIA-sets from the first season of the Dark Energy Survey (DES): from left to right the images correspond to the template (\temp), search (\search), and difference (\diff) image; the \diff\ is generated as the subtraction of \temp\ and \search. Each pair of \temp\ and \search\ images is mapped to the same color range. We refer to these 3-images sets as ``image triplets'' or DIA-sets. A and B show artifacts, human-labeled as ``bogus'' (\texttt{label = 1}). C and D show transients labeled as ``real'' (\texttt{label = 0}).
     Above each triplet is the unique ID of the transient (see \citealt{Goldstein_2015})
    }
    \label{fig:examples_no_normalization}
\end{figure*}

%  in \autoref{sec:DI} we describe in more detail the Difference Imaging processes, in \autoref{sec:autoscan} we present relevant information about the precursor of our work \citep{Goldstein_2015},
\begin{figure*}
    \centering
    \includegraphics[width=0.6\linewidth]{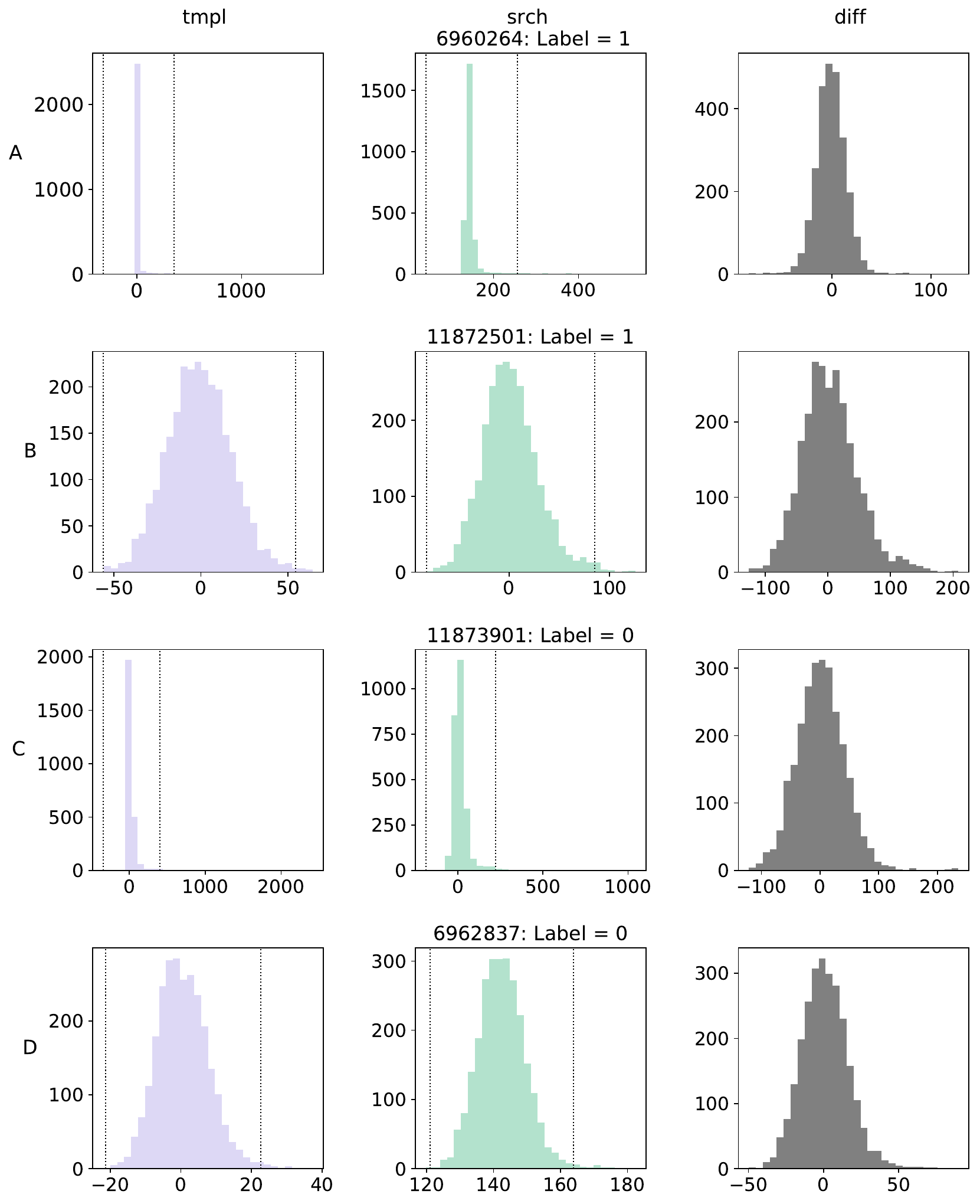}
    \caption{Histogram of pixel values for 
    the data shown in \autoref{fig:examples_no_normalization} (before scaling and normalization). While all difference images (right) show a bell-shaped distribution, \new{template and search images \round{show} different behaviors regardless of the ``real'' or ``bogus'' label. In this image, that the $x$ range for each subplot extends to cover the full range of values in the distribution. While this display choice decreases one's ability to discern details in the core of the distribution, it highlights the information on skewness and asymmetry.} For example, B and D have similar pixel values distribution, however, B is ``bogus'' and D is ``real''. \new{Similarly, A and C, a ``real'' and a ``bogus'' transient respectively, both show right-skewed distributions for \temp\ and \search. } The vertical lines show the $\mu \pm 3\sigma$ interval for the \search\ and \temp\ images. The \diff\ images (last right column) are standardized individually to a mean of $0$ and a standard deviation of $1$. The \search\ and \temp\ are instead scaled, setting the pixel contained inside the 3$\sigma$ interval (vertical lines on the histograms) to the range 0-1. This allows to retain negative values or as well values above $1$ while keeping the core of the distributions to within a homogeneous range. \new{In Appendix A, \autoref{sec:appendixa} , we provide more details about the data preprocessing and include additional plots showing the distribution of data before and after the preprocessing tasks.}} \label{fig:histobeforenormalizarion}
\end{figure*}

\begin{figure*}
    \centering
    \includegraphics[width=0.7\linewidth]{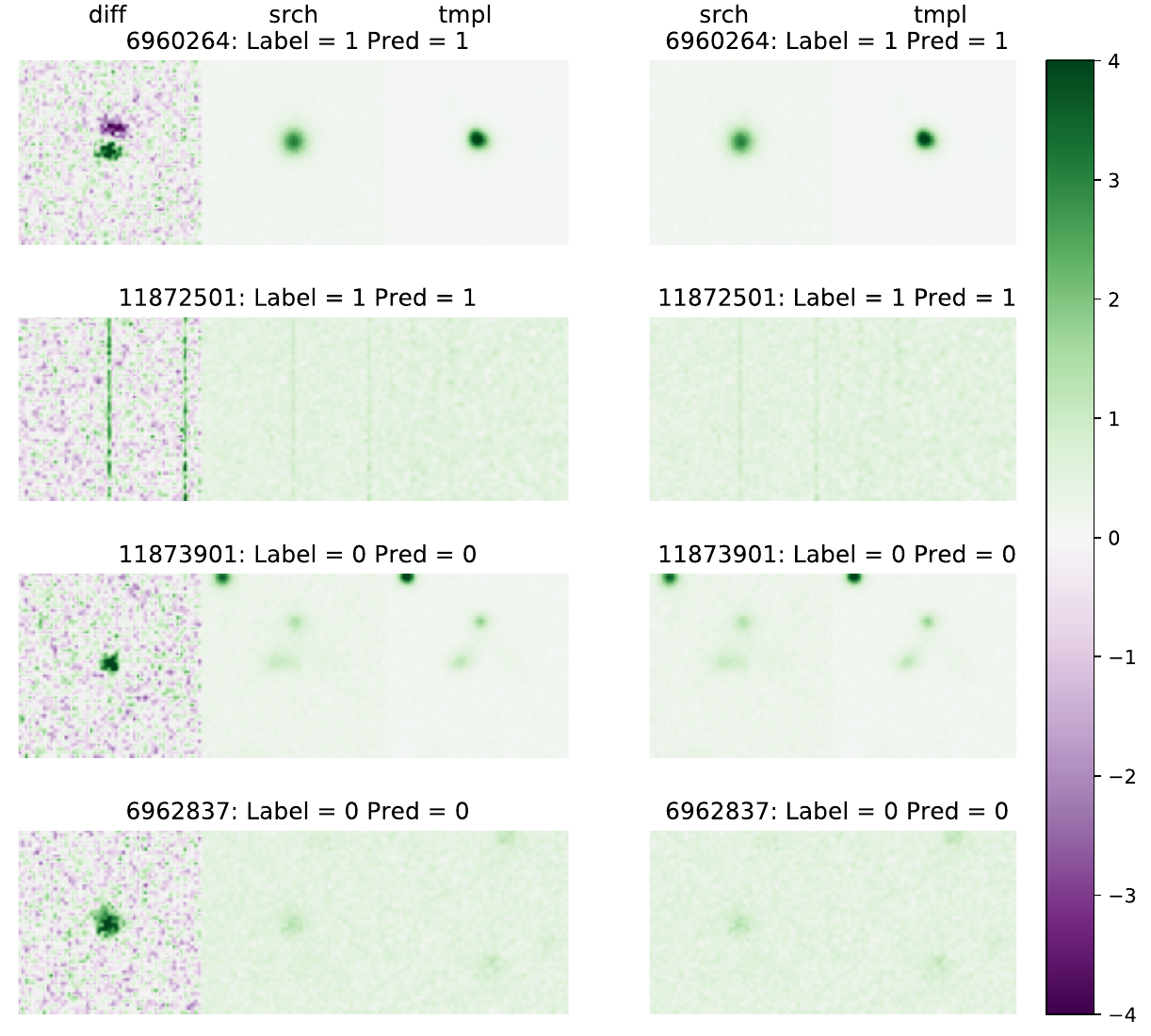}
    \caption{Horizontally stacked data for the examples shown in \autoref{fig:examples_no_normalization}. On the {\it left}, the composite images used as input to the \diabased\ CNN model: the composite follows the order, from left to right: \diff, \search, \temp. On the {\it right}, the composite images used for the \nodia\ model, composed of \search\ and \temp. Each images element was scaled or normalized following the description given in \autoref{sec:data} and \autoref{fig:histobeforenormalizarion} before combining them into a single image. Above each composite are the unique transient ``ID'', the original label, and prediction made by our model. The four transients were classified correctly by both \diabased\ and \nodia\ models. Purple shades indicate negative and green  positive pixel values.
%     \underline{Left}: Horizontally stacked normalized data for the examples shown in \autoref{fig:examples_no_normalization}. The stacked follow the order, from left to right: (\diff, \search, \temp). Each composite images is normalized as follows, the \diff\ are standarized (subtracting the mean and dividing by the standard deviation); \search\ and \temp\ are preprocessed using the min-max scheme. This is the format of the data input in the neural network (see section \ref{subsection: model_neural_network}).
% \underline{Right}: Horizontally stacked normalized data for the examples shown in \autoref{fig:examples_no_normalization}. The stacked follow the order, from left to right: (\search, \temp). Each composite images is normalized as follows, \search\ and \temp\ are preprocessed using the min-max scheme. This is the format of the data input in the neural network (see section \ref{subsection: model_neural_network}). The divergent color map used is \texttt{PRGn}, given by python
}
    %The color map used is ``viridis'', given by python. This images are the final data used for the Neural Network model.  }
    \label{fig:examples_hstack_normalization}
\end{figure*}

%% file: traditionalSolutions.tex
\label{stateofart}
\subsection{Difference Imaging}
\label{sec:DI}
Difference images are produced subtracting a template, an image generated coadding multiple images \citep[\eg][]{Kessler_2015}, from a sky image and they are currently the basis for most astrophysical transient search algorithms. The difference image allows brightness changes to be detected even if embedded in Galaxy light, for example, in the case of extra-galactic explosive transients. Great efforts have been made to improve the quality and effectiveness of the difference images. Although the name may suggest the process simply entails subtracting images from each other, the procedure is in fact riddled with complications because of the following reasons. First, the images used to build the template and the search images are taken principally within different atmospheric conditions \citep{Zackay_2017} generating variations in the quality of the images. The construction of a proper template is also a delicate task; typically templates are built by stacking tens of images taken under favourable sky conditions at different times. This improves the image quality but also mitigates issues related to variability in the astrophysical objects captured in the image \citep{Hambleton_2020}: one wants to capture each variable source at its representative brightness. Typically then, the template image is of higher quality than the search image, and it is degraded to match the search image PSF and scaled to match its brightness. Yet, the scaling and PSF may vary locally in the image plane, especially for images from large field of view synoptic surveys such as the 2.2 sq. degrees DES or $\sim10$ sq. degrees Rubin images. Finally it is important that the images are perfectly aligned, both in the creation of the template and in the subtraction process to create the difference image. This implies accounting for rotation as well as potentially different warping effects on the images and template. Once the PSF match and the alignment is done, it is possible to subtract the degraded template from the search images to obtain the difference image. To degrade the image quality of the template to match the search image, a convolution kernel that must be applied to the template needs to be determined \citep{Alard_1998},
\begin{eqnarray}
    \text{\temp}(x,y) \otimes \text{Kernel}(u,v) = \text{\search}(x,y),
\end{eqnarray}

\noindent 
where \temp\ is the high-quality image, the template image, \search\ is the one night image or search image and $\otimes$ is the convolutional operation. The arguments $x$ and $y$ represent the coordinates of \new{the pixel} matrix that compose the images; $u$ and $v$ the coordinates of the kernel matrix. 

To solve the computationally expensive problem of matching PSFs, the kernel can be decomposed in terms of simple functions, for instance Gaussian functions, and the method of least squares can be used to determine the best values for the kernel. The fitted solution of one search image can be determined in a short computational time. However, the computational cost scales with the image size and resolution. Surveys and telescopes constructed with the goal of discovering new transients are generally designed to collect tremendous amounts of data to maximize event rate (detection of astrophysical transients).  For the DES, the computational cost per 2.2 sq. degree image is $\sim 15.5$ CPU hours (with roughly 2/3 of that time spent on PSF matching)\footnote{\new{DES team, private communication.}}.  The upcoming Rubin LSST will collect more than 500 images every night each with 3.2 Gigapixels. This process thus is bound to turn out to be very expensive.
In this paper, we trained our models on postage stamps where transients were detected or simulated (see \autoref{sec:data}), thus a direct comparison of the computational cost is not trivial. \new{For comparison, a} detailed discussion of the computational cost of our models is included in \autoref{sec:computationcost} and the CPU Node hours required to train and generate predictions from our models are reported in \autoref{tab:acc results}. We note here that, with a Deep Neural Network approach to this problem, the computational cost is high in training, but the predictions  \new{require minimal computational time}.

A bad subtraction can occur either because of poor PSF matching, poor alignment, or poor correction of image warping. In all of these cases, the subtraction would lead to artifacts or ``bogus'' alerts, like the one shown in the difference image in  \autoref{fig:examples_no_normalization}A and B. In particular: in \autoref{fig:examples_no_normalization}A, the difference image shows a so-called ``dipole'', where one side of a suspected transient is dark and the other side bright: this typically arises in case of mis-alignments, but it might also be caused by moving objects in the field, or differential chromatic refraction \citep{carrascodavis2021alert}. Conversely, \autoref{fig:examples_no_normalization}B shows a ``bogus'' alert caused by an image artifact: a column of bad pixels in the search image. At this location there is no astrophysical object in the image thumbnail: no host galaxy or star that could give rise to variations.

Panels \autoref{fig:examples_no_normalization}C and \autoref{fig:examples_no_normalization}D show genuine transients in our DES training data: in both of these two examples, there is a clear transient in the \diff\  images (high pixel values at the center of the image).

\subsection{Autoscan and other feature-based Real Bogus models}
\label{sec:autoscan}

We developed our models on data collected in the first year of DES. Thus, a direct precursor of our work is \cite{Goldstein_2015}, in which the authors created an automated RB based on a Random Forest (RF) supervised learning model \citep{ho1995random} to detect transients, and particularly supernovae, in the DES data, hereafter refered to as \texttt{autoscan}. For these kinds of models, the process of selecting and engineering features is pivotal. %\citep{2011AAS...21720507S} (\masao{I suggest removing this reference}).
% \footnote{The data can be found at \url{https://portal.nersc.gov/project/dessn/autoscan/\#}} 
\texttt{Autoscan} is based on 38 features derived from the \diff, \search, and \temp\ images. The selection and computation of these features was done attempting to represent quantitatively what humans would leverage in visual inspections. For instance, \texttt{r\_aper\_psf}  distinguishes a bad subtraction of \search\ and \temp\ that would lead to a \diff\ qualitative similar to  \autoref{fig:examples_no_normalization}A; the feature \texttt{diffsum} measures the significance of the detection by summing the pixel values in the center of the \diff\ image; the feature \texttt{colmeds}, indicating the CCD used for the detection, is designed to identify artifacts specific of a CCD, like bad rows/columns of pixels.

In other RB models, like in \cite{S_nchez_2019},
the feature selection is performed purely statistically: features were initially selected based on variance thresholds. In the same work, different techniques are explored to reduce the number of features, and thus the complexity of the classification problem. For example, a RF model was trained using all features. Then \new{a feature importance analysis enabled a reduction of the} dimensionality of the problem by removing possible redundant or irrelevant features. Examples of models based on features closer to the data include \cite{Mong_2020}, where the features are simply the flux values of the pixels around the center of the image.

\subsection{Deep Neural Netowrk approaches}\label{sec:AI}

CNNs have demonstrated enormous potential in image analysis including object detection, recognition, and classification across domains \citep{5206848}. Examples of astrophysics applications of CNNs include \citet{Dieleman_2015} for galaxy morphology prediction, \citet{Kim_2016} for star-galaxy classification, \citet{PhysRevLett.120.141103} for signal/background separation for Gravitational Waves (GW) searches, where the GW time series are purposefully encoded as images to be analyzed by a CNN, and many more.

CNNs are particularly well-suited to learning discriminating features from image input data. CNNs can work on high dimensional spaces (here the dimensionality of the input is as large as the number of pixels in the image) due to the generalizability of the convolution operation to $n$ dimensions while preserving relative position information. Vectors of raw pixel values can theoretically be used to train traditionally feature-based models, such as RFs, but pixel-to-pixel position data in higher dimensions is unequivocally lost. Previous studies already compared feature-based supervised models, like RF, and supervised CNNs for RB, demonstrating that CNN\new{s} generally lead to increased accuracy.  In \cite{Gieseke_2017} an accuracy of $\sim0.984$ is achieved with a RF model in the RB task, and it increases to $\sim0.990$ when applying a CNNs to the same data. In \cite{Cabrera_2016} \new{an}  accuracy of $\sim 0.9889$ \new{is} achieved with a RF \new{and} is increased to $\sim0.9932$ with a CNN. In \cite{Cabrera_Vives_2017}, an RF gave 0.9896 and a CNN 0.9945. In \citet{Liu_2019} the accuracy improves from $\sim0.9623$ with a RF to $\sim0.9948$ with a CNN.
% added here because is DIA (3 images)

\new{In \citet{Duev_2019} a CNN RB classifier called \texttt{braai}, developed for the Zwicky Transient Facility \citep[ZTF]{Bellm_2018}, achieves a $\sim$98\% accuracy for training and validation data set (Area Under the Curve, AUC =  0.99949). \texttt{braai} implementes a custom VGG16 \citep{VGG16} architecture.}

\new{Going beyond RB, a model for image-based transient classification through CNNs has been prototyped by the Automatic Learning for the Rapid Classification of Events team \citep[ALeRCE][]{carrascodavis2021alert}. The model classifies between AGNs (Active Galactic Nuclei), SNe (SuperNovae), variable stars, asteroids, and artifacts in ZTF \citep[]{Bellm_2018} survey data with a reported accuracy exceeding 95\% for all types, except SNe (87\%). This CNN model was trained using a combination of the \search, \temp, and \diff.}%, where rotational invariance was considered by training the model with rotated version of the images and features that characterize each astronomical object type were passed through a layer and concatenated with the output layer of the images. %They results have shown that AGN can be miss-classify for VS and vice versa; SN for asteroids or bogus; asteriods for SN or VS, and bogus for all of them except AGN.

Here we explore the potential and the intricacies of leveraging AI to bypass the DIA step. 
\new{The CNN RB models mentioned above} differ not only in their architecture (for example, a single or multiple sequences of convolutional, pooling, dropout, and dense layers) but also in the choice of input. For instance, \cite{Gieseke_2017} used the \temp, \search, and \diff\ images in combination for training their CNN; \cite{Cabrera_2016} augmented this image set with an image generated as the \diff\ divided by an estimate of the local noise;  \cite{Cabrera_Vives_2017} trained an ensemble of CNNs on different rotations of this four-fold image set. Conversely, \citet{Liu_2019} used the \diff\ as the sole input. Although, all these attempts have shown good results with accuracy higher than $90\%$, these models all rely on DIA to construct the \diff. Taking into account that \diff\ are built from the \temp\ and \search, and in principle, should carry no additional information content than the search-template pair alone, a logical step to follow would be to only consider the two latter images.

\tatiana{A first attempt in this direction is presented in \cite{Sedaghat_2018}, where the authors develop a Convolutional Autoencoder (encoder-decoder) named \texttt{TransiNet}. The model is developed and tested on both real and synthetic data. Synthetic data was created by using background images from the Galaxy Zoo data set in Kaggle \citep{galaxy-zoo-the-galaxy-challenge}, then simulated transients were implanted in the search images. Template and search images from the Supernova Hunt project, Catalina Real-time Transient Survey, \citep[CTRS]{Drake_2009}, were also used. Data were fed to the autoencoder to generate a difference image that contains only the transient (the CNN do not generate background noise). 
The CNN model was trained and tested only on synthetic data and separately trained on a combination of synthetic and real data and tested on the real data. The former model achieves scores (precision and recall) of 100\%; the latter model a precision of 93.4\% and recall of 75.5\%, and establish a precedent for the possibility of avoiding the construction of the DIA-\diff\ to reliably detect optical transients.

Another notable work where difference images are not used is \cite{Carrasco_Davis_2019}. The authors implement Recurrent convolutional
Neural Network (RCNN) to train a sequence of images (instead of the classical template and search images) to classify 7 types of variable objects. The model was trained using synthetic data and tested using data from the High cadence Transient Survey \citep[HiTS]{F_rster_2016}. The average performance recall of the model is 94\%.}

\cite{wardega2020detecting} trained a model that could distinguish between optical transient and artifacts using a search image from Dr. Cristina V. Torres Memorial Astronomical Observatory (CTMO, a facility of the University of Texas Rio Grande Valley\footnote{\url{https://www.utrgv.edu/physics/outreach/observatory/index.htm}}) and a template image from the Sloan Digital Sky Survey \citep[SDSS]{Gunn_2006}. They trained two Artificial Neural Network models, a CNN and a Dense Layer Network on simulated data and tested the models using data from CTMO and SDSS. The data used for training and testing had specific characteristics: transients were a combination of a source in the CTMO images (or \search\ image) and background in the SDSS image (or \temp\ image); artifacts were a combination of a source in both the CTMO and SDSS images. Within this dataset, both models yield high accuracy ($>95\%$).  However, studies based on more diverse and realistic data, \eg, sources near galaxies, or embedded in clusters, are needed to demonstrate the feasibility of this approach. The data used in this paper fulfills this condition and is described in \autoref{sec:data}.

%% file: data.tex
This study is designed as a detailed comparison \new{RB} of CNN-based models, with and without \diff\ in input. \new{Our starting point is the well known \texttt{autoscan} random-forest-based RB \citep[][see \autoref{sec:autoscan}]{Goldstein_2015}, which supported the DES thousands of discoveries since its first season: we train our model on the data that \texttt{autoscan} was trained on, and benchmark our results to the performance of \texttt{autoscan}}. The choice of \texttt{autoscan} as our point of reference \new{and benchmark} is motivated by its application to the discovery of transients in a state of the art facility, the DES \citep{DES}, which can be considered a precursor of upcoming surveys like the Rubin Legacy Survey of Space and Time. The latter,  expected to start in \new{2025}, will deliver $\sim 20~\mathrm{Tb}$ of high resolution sky image data each night, covering a footprint of $\sim20,000$ sq. deg. every $\sim 3$ nights, with expected millions of transients per night, demanding rapid methodological and technical advances in accuracy and efficacy of transient detection models \citep[LSST]{ivezic2019lsst}. \new{In particular, the properties of the DES images are expected to be similar to those of Rubin LSST given the similar image resolution (0.26''/pixel and 0.2''/pixel for DES and LSST respectively, which results in seeing-limited images taken from nearby sites in Chile with similar sky properties) and similar imaging technologies (both cameras employ similar  chips, wavefront sensing, and adaptive optic systems, \citealt{2016SPIE.9906E..4JX}), although the field of view of Rubin LSST is much larger and the overall image quality is expected to be superior to precursor surveys}.

%\section{
% The data used in this project consists of images collected by The Dark Energy Survey during its first observational season, August 2013 through February 2014. The data corresponds to observations of $898,963$ transients labeled by human inspection in two groups (\textit{OBJECT\_TYPE}): $454,092$ are labeled as real astrophysical transients ($1$) and $444,871$ as artifacts, or ``bogus'', as it is customary in the field,. Three images are made available for each transient, each of size $51 \times 51$, divided in three groups: the template (\temp) images, science images (\search) and the difference (\diff) images, correspond to the result of subtraction of template and science images [\cite{Goldstein_2015}]. Every group of three images (\diff,\search,\temp) correspond to a unique 'ID' and each one of the images has the same label information encoded in a binary 1 or 0 for ``real'' and ``bogus'' respectively.
% %}
% All the information about this training instances can be found in \url{https://portal.nersc.gov/project/dessn/autoscan/#}.\\

The data used in this work consists of postage stamps of images collected by the DES during its first observational season (Y1), August 2013 through February 2014\footnote{The data can be found in \url{https://portal.nersc.gov/project/dessn/autoscan/\#}} \citep{Abbott_2018}. The data corresponds to 898,963 DIA-sets, a template (\temp) image, search (\search) image, and their difference (\diff). The construction of the templates images for the DES Y1 leveraged the data collected in season two (Y2) as well as the Science Verification images (observations collected prior to survey start in order to evaluate the performance of the instrument). More information on the DES DIA pipeline can be found in \citep{Kessler_2015}.  Of these DIA sets, 454,092 contain simulated SNe Ia, which constitute the ``real'' astrophysical transients set (\texttt{label = 0}) and 444,871 are \new{human-labeled images} from DES, \ie, the ``bogus'' set (\texttt{label = 1}).\footnote{\new{Each image is offered in two formats: ``.gif'', and ``.fits''. The former is an 8-bit compressed format (convenient for visual inspection as it can be opened with commonly available software); the latter, the ``Flexible Image Transport System'', is a common data format for astronomical data sets which enables high precision with a large dynamic rage. The data in ``.fits'' format was used in this work.} More information related to how to manipulate this astronomical data format is available in \url{https://docs.astropy.org/en/stable/io/fits/}.} Each image is $51 \times 51$ pixels - corresponding to approx \new{180} 
arcseconds square of sky. 

 Some examples of the data are \new{shown} in \autoref{fig:examples_no_normalization}. Each transient is identified by a unique ``ID''. The metadata, includes the labels associated with each image as well as the the $38$ features used for classification in \citet{Goldstein_2015}.
 Because we only analyze postage stamps with detections, we implicitly still rely on the DIA to enable the detection step at this stage of our work. The \temp\ images in our postage stamps, however, are not PSF matched.
% \\
% \vspace{-10.8mm}

 %The other two columns corresponded to the ``ID'' and the ``OBJECT$\_$TYPE'': ``real'' (0) and ``bogus''(1); these two were used for this work.
\subsection{Scaling and normalization}\label{subsec:scaling}

A word about data preparation and normalization is in order as  astrophysical images are inherently very different from the images upon which CNNs have been built. When training CNN models for image analysis, each image is typically simply scaled to a common range ($0-1$). However, the dynamic range of an astrophysical image is typically large and the distribution of pixel values is generally very different from Gaussian, with the majority of pixels sitting at low values (the sky) and a few pixels at or near saturation (which in some cases may carry the majority of the information content). 
Furthermore, in the DIA-set case, the pixel-value distribution of the \diff\ differs qualitatively from the \temp\ and \search\ ones. While the \temp\ and \search\ are typically naturally positive valued with a long tail at the bright end (right-skewed because of the presence of bright astrophysical sources such as galaxies that host transients or stars that vary), the \diff\ image is, in absence of variable or transient sources, symmetric around 0 \new{(see \autoref{fig:histobeforenormalizarion})}.

%To gave more relevance to the 
The \diff\ images %and for their Gaussian behavior, as seen in \autoref{fig:histobeforenormalizarion} (first left column); each of them was
were standardized % The \diff\ images would 
to have a mean $\mu = 0$ and a standard deviation $\sigma = 1$. The \search\ and \temp\ images, instead, were scaled %minimum and maximum value for each image and considering pixel values inside the 
to map the $\mu\pm 3\sigma$ interval of the original image to $0-1$. %Meaning that $mean \pm 3\sigma$ would be $0$ or $1$ respectively. However, 
This scheme allows us to retain resolution in the shape of the core of the distribution while also retaining extreme pixel values. % allows  negative values and values above $1$ because extreme (outside the $3\sigma$ clip) values were scaled too. 
    \autoref{fig:histobeforenormalizarion} shows the distribution of pixel values for four DIA sets, for the same data as in \autoref{fig:examples_no_normalization}, which include two ``real'' and two ``bogus'' labels. \new{The distribution moments used for standardization are shown}. \hyperref[sec:appendixa]{Appendix A} shows the pixel distributions before and after scaling for the same data in some more detail.

\begin{figure*}

   \centering
   \includegraphics[width=0.45\linewidth]{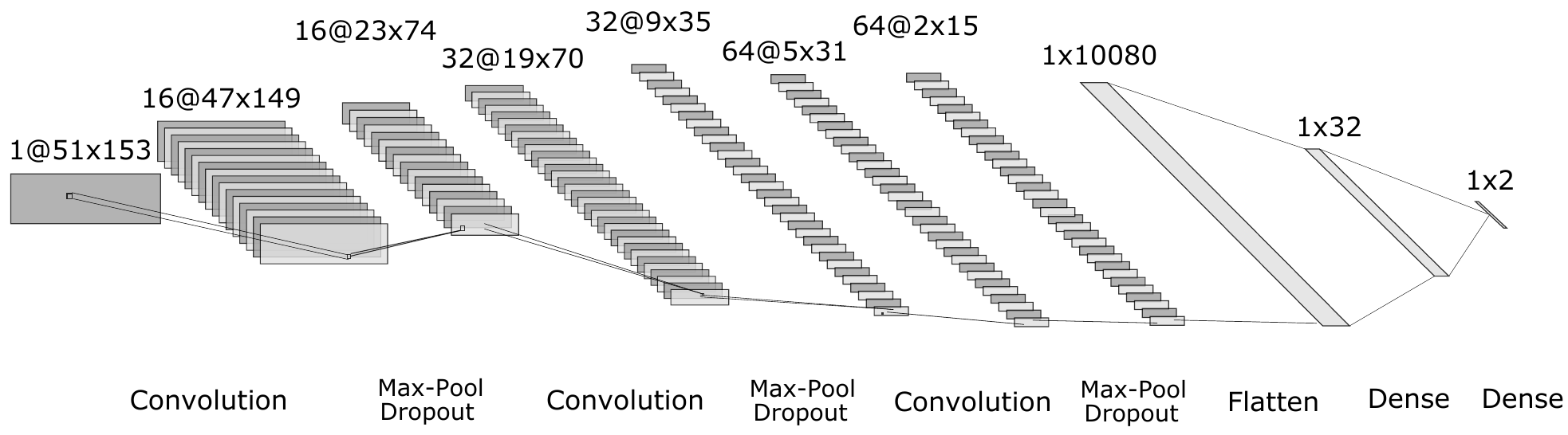}
    \includegraphics[width=0.45\linewidth]{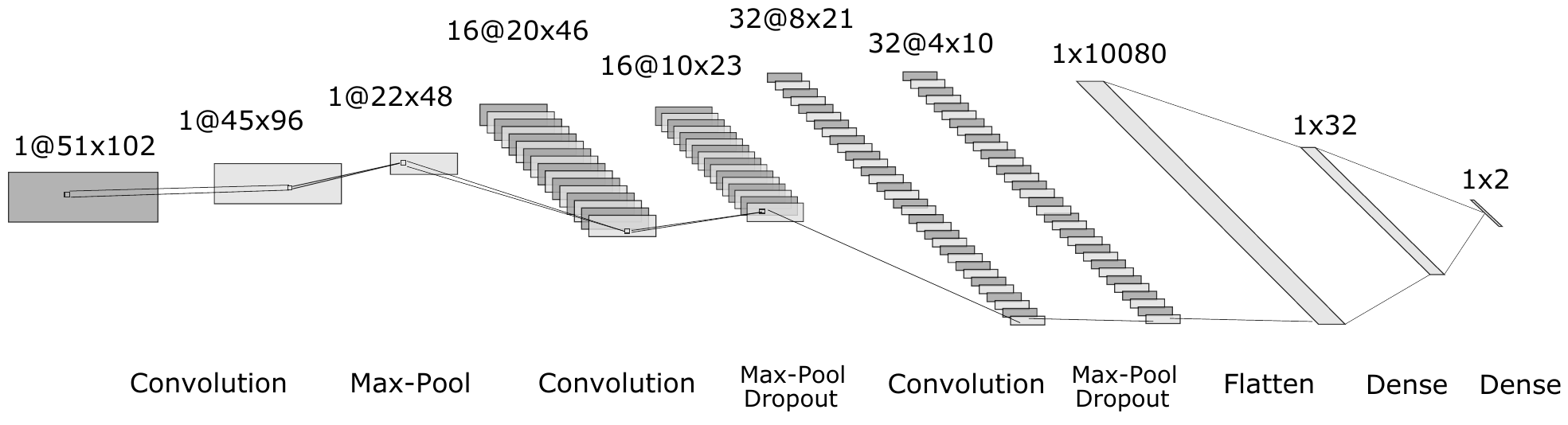}
   \caption{Architecture of the Neural Networks used in this project to classify ``real'' and ``bogus'' transients. {\it Left}: the \diabased\ model that uses image triplets as input (\diff, \temp, \search). The input layer is $51 \times 153$ (see left \autoref{fig:examples_hstack_normalization}); a convolution layer $(5\times 5)$ learns $16$ filters; max pooling $(2\times2)$ and dropout; convolution $(5\times 5)$ learns $32$ filters;  maximum pooling $(2\times2)$ and dropout; convolution $(5\times 5)$ learns $64$ filters;  maximum pooling $(2\times2)$ and dropout; flatten layer, Dense $(32)$ and the output is a Dense $(2)$-class layer. {\it Right}: the \nodia\ model that uses the \temp\ and \search\ images only. The input layer is $51 \times 102$ (see right \autoref{fig:examples_hstack_normalization}); a convolution layer $(7\times 7)$ learns $1$ filter; maximum pooling $(2\times2)$; convolution $(3\times 3)$ learns $16$ filters;  maximum pooling $(2\times2)$ and dropout; convolution $(3\times 3)$ learns $32$ filters;  maximum pooling $(2\times2)$ and dropout; flatten layer, Dense $(32)$ and the output was a Dense $(2)$-class layer.The illustrations were made using NN-SVG tool by \citet{LeNail2019}.}
\label{fig:architecturesCNN}
\end{figure*}

One further decision has to be made in combining the three images in the DIA set to feed them to the CNN. \new{While commonly the images would be stacked depth-wise,} we stacked the scaled \diff, \search\ and \temp\ horizontally. %  the three each image (already standardized and scaled) per ``ID'', following the order (\diff, \search, \temp), and using \textit{np.concatentate}; 
Thus the size of the data in input to our CNN is ($N_{tr}\times51\times153$), where $N_{tr}$ is the number of transients to be considered. Four examples of the data ``triplets'' in input to our \diabased\ CNN are in the left panel of \autoref{fig:examples_hstack_normalization}. Following this horizontal structure, we mimic closely the way that human scanned this type of data for classification, and we will take advantage of this scheme when examining the models' decisions in \autoref{subsec: saliency}. \tatiana{We inspected the impact of this choice by comparing the accuracy  of a model that was given the  images stacked in depth ($51\times51\times3$) with the model with $51\times153$ images in input, with otherwise identical architecture after the first hidden layer, and found that our choice does not affect the overall performance (both models achieved 96\% accuracy as will be discussed in \autoref{sec:results}; 
see also \hyperref[sec:appendix3channel]{Appendix A}, \autoref{fig:confusionechannel}).} %this acc is for CAVINESS, plots here are NERSC, acc is down by a ~1% using CAVINESS, NERSC is a lit less than 97%

\new{Since our goal is to measure the impact of \new{reducing the information passed to the model the input} by not using the DIA, for our \nodia\ models the training and testing data set were constructed in the same way as the previous triplets with the \temp\ and \search\ side-by-side}, but without the \diff\ image. Some examples are in the right panel of \autoref{fig:examples_hstack_normalization}.

% \onecolumngrid

%% file: model.tex
\subsection{Basics of Neural Networks}\label{subsection: intro_neural_network}
Neural Networks are models that learn important features and feature associations directly from data. % by training with a portion of the total data. 
They \new{can be used as} supervised learning models  for classification or regression. They consist of a series of layers of linear combinations of the input data each with real value parameters known as weights and biases, combined with activation functions that enable learning {\it non-linear}, potentially very complex, relationships in the data. The weights tell us the relevance of the input feature with respect to the output, and the biases are the \new{offset} value that determine the output. %(e.g. if a final output value is greater than a certain threshold then is A otherwise is B). 
Fitting these quantities to the data minimizes the %loss These quantities are the ones that guarantees the minimum
loss,
meaning the prediction is as close as possible to the original target/label \citep{nielsen_NN}. For Deep Neural Networks (DNNs), ``deep'' refers to the fact that there are multiple hidden layers, \new{between} the input and output layer.  %that follow a layer structure, where each layer learns new and complex features of the data, architecture of these models usually consist on a input layer + hidden layers + output layer. Each hidden layer has weights and biases that is connected with the other hidden layers and both input and output data. 
With this architecture the features learned by each layer do not follow a human selection, rather, the features arise in the analysis of the data \citep{LeCun_Bengio_Hinton_2015}. Among DNNs, CNN models have layers that represent convolutional filters.
%The layers used in this work are:  \mintinline{c}/Conv2D/, \mintinline{c}/Dropout/, \mintinline{c}/Flatten/, \mintinline{c}/Dense/, etc... - a description of the structure and use of these standard layers can be found in 
More information on DNNs and CNNs can be found in \cite{DBLP:journals/corr/abs-1803-08375, 10.5555/2999134.2999257, Dieleman_2015}, as well as \cite{Gieseke_2017}, among many others.

We used the \texttt{Keras}\footnote{\url{https://keras.io/api/layers/convolution\_layers/convolution2d/}.} implementation of the CNN \citep{chollet2015keras}.

\subsection{DIA and noDIA based models architecture}

The input of our Neural Networks are the horizontally stacked images of size $51 \times 153$ for the \diabased\ model (\diff, \search, \temp), and $51 \times 102$ for \nodia\ model (\search, \temp). \tatiana{For both the \diabased\  and the \nodia, $100,000$ images were used to build the model: $80,000$ images for training  and $20,000$ for validation. An additional  set of $20,000$ images is used for testing, \emph{i.e.} the predictions on this test are only done after the model hyperparameters are set, and the result reported throughout are based on this set.} The images were selected randomly from the $898,963$ and while certainly training with a larger set can lead to higher accuracy, the amount of images was sufficient for this comparison of \diabased\ and \nodia\ models while being conservative with limited computational resource. The data is composed of $50,183$ images labeled as ``bogus'' and $49,817$ labeled as ``real''. %Each model was tested using $20,000$ divided as  $9922$ ``bogus'' and $10,078$ ``real''.\\

The network architectures used for this work are shown in  \autoref{fig:architecturesCNN}, left panel, for the \diabased\ model, and right  panel for the \nodia\ model. More details about the architecture can be found in \hyperref[sec:appendixb]{Appendix B}. Both architectures follow a similar structure. 

In designing the neural networks, we started with the \diabased\ model and developed an architecture that would match the performance of \citet{Goldstein_2015}.  %The combination tested and their respective performances are given in APPENDIX.}
\new{While examples exist in the literature of RB models with higher measured performance (see \autoref{sec:stateofart} and \citealt{Gieseke_2017},  \citealt{Cabrera_2016}, \citealt{Cabrera_Vives_2017}, \citealt{Liu_2019}, \citealt{Duev_2019}, \emph{etc.}), we emphasize that each one of these models is applied to a different datasets, such that their performance cannot be treated \round{as a} benchmark. Furthermore, our goal here is not to replace the DES RB model with a higher performing one, but to measure the impact of \new{ the loss of information content in the input} caused by removing the \diff. Matching the performance of the accepted model for RB separation within the DES is a sufficient result for our purpose. In fact, more complex architectures have been implemented on these data without a significant performance improvement \round{(see \hyperref[sec:appendixb]{Appendix B.2})}. This leads us to believe that at least a fraction of the 3\% incorrect predictions are associated with noisy and incorrect labels (see \autoref{subsec:performance}).}

With this model in hand, we created a \nodia\ with a similar structure in order to enable direct comparison and measure the effect of the change in the input data. 

\begin{figure*}
    \centering
    \includegraphics[width=0.45\linewidth]{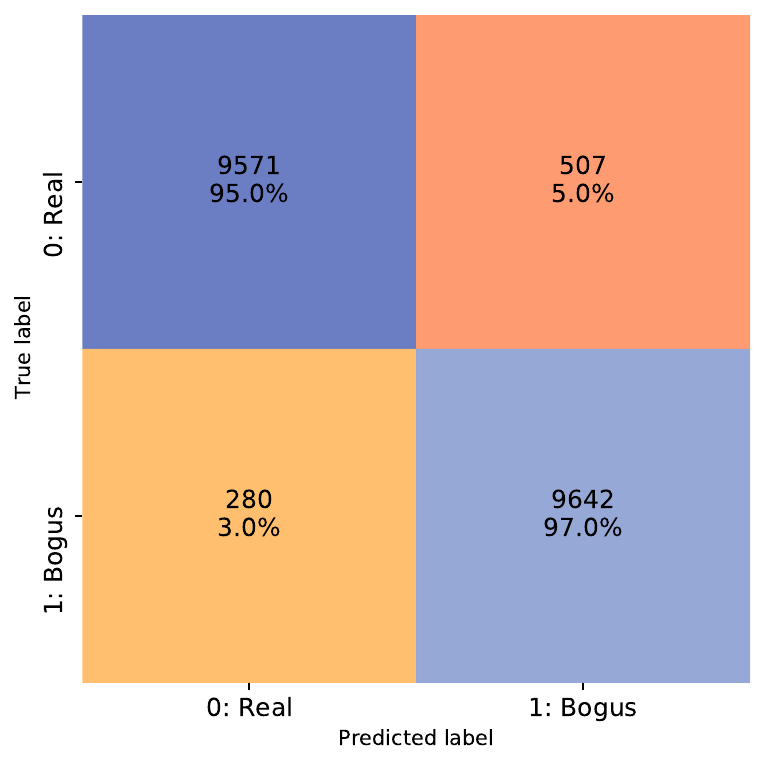}
    \includegraphics[width=0.45\linewidth]{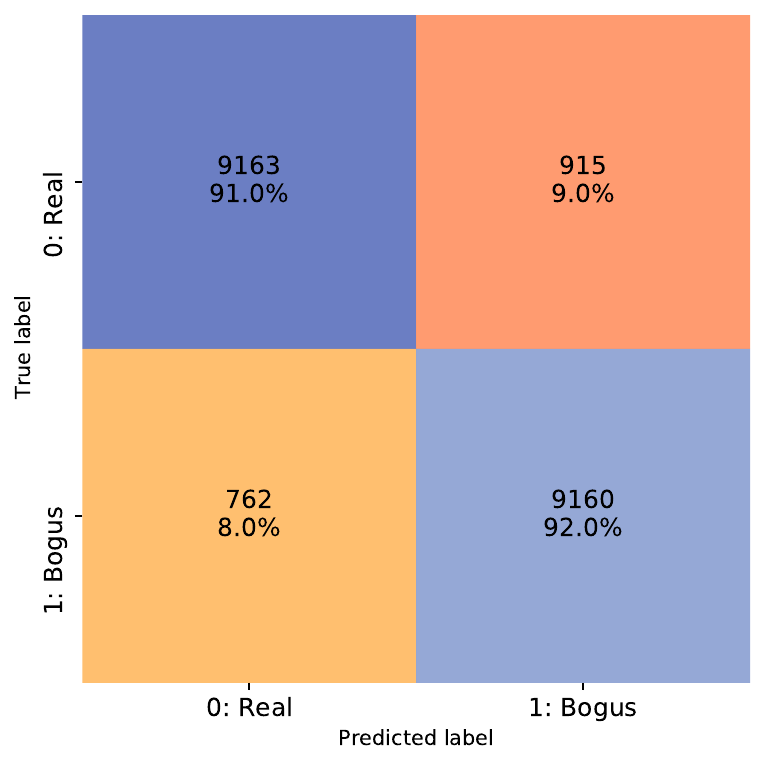}
    \caption{Confusion matrix (CM) for our testing data, a set composed of $10,078$ objects labeled as ``real'' and $9,922$ as ``bogus''.  Squares of the matrix, from the top left in the clockwise direction, indicate: True Positive (TP, \texttt{label = 0}, \texttt{prediction = 0} ), False Negative (FN, \texttt{label = 0}, \texttt{prediction = 1}), True Negatives (TN, \texttt{label = 1}, \texttt{prediction = 1}), and False Positive (FP, \texttt{label = 1}, \texttt{prediction = 0}). We note that here, somewhat unusually, 0 corresponds to ``real'' and ``positive'', and 1 to ``bogus'' and ``negative'', as we chose to remain consistent with the original labeling of the data presented in \citep{Goldstein_2015}. 
    {\it Left}: The CM of our \textbf{\diabased } model shows that from the $10,078$ transients labeled as ``real'', $9,571$ (\ie, 95\%) were correctly classified and the \tatiana{rate} for the TN objects is even higher, at $97\%$. {\it Right}: The CM for the \nodia\ model shows that of the $10,078$ transients labeled as ``real'', $9,163$ (\ie, 91\%) were predicted correctly and the TN \new{rate} is $92\%$. }
    \label{fig:confusiomatrix_models}
% \end{figure}
\end{figure*}

In the development of our models we followed two general guidelines: 
\begin{enumerate}
\item When designing the models, our goal was to push the accuracy of the  \diabased\ CNN model to match the accuracy of \texttt{autoscan}. While a more exhaustive architectural exploration or a hyperparameter grid search may well lead to increased efficacy, matching the accuracy of \texttt{autoscan} (at $\sim$ 97\%) is sufficient for our demonstration. 
  The final architecture used is one that reached the same accuracy as \citet{Goldstein_2015}, and  False Positive and False Negative Rates most similar to the ones obtained in \citet{Goldstein_2015}  (see \autoref{fig:confusiomatrix_models} and \autoref{sec:results}). Once we matched the \texttt{autoscan} performance, we focused on the potential for removing the \diff\ image from the input.
\item While the architecture of the \diabased\ CNN was designed with the attempt to achieve a specific target performance, the architecture of \nodia\ is deliberately kept as close as possible to that of the \diabased\ model. This enables a direct comparison of the effects of the removal of the \diff\ image. \new{ 
The architecture of \nodia, shown on the right in \autoref{fig:architecturesCNN} was not optimized explicitly for RB classification, rather was inherited from the \diabased\ model, only modifying the original design to adapt for the different dimensionality of the input data. 
One further deliberate modification is implemented in the choice of a single-filter first layer for \nodia. \new{The \diff\ image is produced by matching the PSF of the science image in the template (and scaling the brightness with a trivial scaling factor). Everything else needed for the RB classification is contained in the \temp-\search\ pair as is in the \temp-\search-\diff\ triplet}. Thus the CNN that is not offered the \diff\ image needs to learn the image PSF, which is constant across the postage stamp size image and it should be possible to model it with a single filter.} 
%} \new{Our goal with this work is to assess the impact of the information loss due to dropping the \diff\ image in input and to examine the behavior of the CNN in absence of a \diff\ image. We reserve the specific optimization of a CNN-based model to the detection and classification of transients without leveraging the DIA to future work and we note that there is likely significant space for improvement in this direction with architectural choices specifically designed for the search and template input pairs}.

\end{enumerate}
\subsection{Performance assessment}\label{subsec:performance}
\tatiana{Although our goal was to measure the performance impact of the {loss of input}, as part of our performance assessment tasks, we conducted an extensive hyperparameter search on the existing \round{\nodia}\ architecture and tested alternative pre-packaged architectures \round{for the \diabased}\ known to perform well on image classification. 

We performed a grid searches varying the kernel size of the convolutional layers and the batch size \round{(see \hyperref[sec:appendixb]{Appendix B.2}, \autoref{tab:noDIAgridsearch})}. The optimal parameters are reported for each model in \hyperref[sec:appendixb]{Appendix B}. 
% \round{We point out that some of the combinations in the grid search provide slightly better accuracy for the testing data, $\sim 0.911$, than the one reported in this work. However the True Negative and True Positive rates (see \autoref{sec:results}) were in those cases unbalanced, with a decrease in the True Positive rate}. 
We re-trained two pre-made deep learning models, VGG16 \citep{VGG16} and ResNet 50-V2 \citep{RESNET50V2}, \round{(see \hyperref[sec:appendixb]{Appendix B.1}, \autoref{tab:DIACNNresnet_architecture})}  with the same data used to train the \diabased\ presented here. Neither outperformed our \diabased\ model, while both presented issues with overfitting. Pre-made architectures force constraints on the size of the images used to train the model, %To implement these architectures the data must have the form of: (height, width, 3), thus the horizontal stacking strategy we chose for our models, which will be helpful in the interpretation of the CNN (see \autoref{subsec: saliency}) is no longer possible. 
thus these models were trained with input images stacked in depth ($51\times51\times3$, see \autoref{subsec:scaling}). The small size of our postage stamps also limits the available architectures to models with relatively few layers (due to the repeated application of pooling layers). %Furthermore, training the model with the \nodia\ approach is only possible by changing the shape of the images to match the require third depth axis. 
VGG16  achieves a performance similar to our model's in terms of the testing accuracy with 0.9603, but trains faster and begins overfitting after $\sim20$ epochs (\round{where overfitting is diagnosed visually from the loss curves identifying the epoch at which the validation loss starts increasing in spite of continued improvements in the training loss}). ResNet 50-V2 shows signs of overfitting throughout the entire training process and only achieves a test accuracy of 0.9416. 
We conclude that this historical data set 
%is likely to% %we know it contains inaccuracy because "bogus" are the real data and "real" are the simulations%
contains some level of label inaccuracy such that surpassing the performance of \texttt{autoscan} may be effectively impossible. %In fact, when reviewing the results of our models, we noted that %

\round{We remind the reader that in this dataset, simulated supernovae are by default labeled as Real. However, we found that many images labeled as Bogus, upon visual inspection could be re-classified as transients (we estimate between 3\% and 10\% of the Bogus labels could be reclassified (see \hyperref[sec:appendix-relabel]{Appendix D})}.
% Several images classified as false positives
%based on the original labels looked, upon visual inspection, as genuine transients.%\fed{should we show an example here}. 
This dataset has now been archived and cannot be further validated by, for example, assessing the recurrence of transients at a sky position to validate the {\it bogus} nature of a non-simulated detection. Thus, while CNN models do exist in the literature with higher accuracy, our model's performance is considered optimal at $0.961$.}
\round{We performed a $k$-fold cross-validation with $k=6$ for the \nodia\ model. The average accuracy for the test data set is $0.911 \pm 0.005$.}

%\question{Can you show me the loss plot?}. 
\subsection{Saliency maps}\label{subsec: saliency}

% \masao{This is a very interesting section!  For refeers like me who don't know much about how saliency maps are produced, can you write a couple of sentences about how they are made?  I think you said something about removing each pixel and looking at how much the predictions change.  A short description is helpful for the uneducated reader like me.}

Saliency maps quantify the importance of each pixel of an image in input to a CNN in the training process. They provide some level of interpretability through a process akin to feature importance analysis by enabling  an assessment of which pixels the model relies on the most for the final classification. If the task were, for example, to identify cats and dogs in images, the expectation will be to find that the most important pixels are located within the dog or cat bodies, and not in the surrounding, while if the task were to identify activities performed by cats and dogs, one may find the important pixels both within the dogs and in the surrounding, particularly in objects associated with the performed tasks. Furthermore, some portions of the subject's body may be most distinctive (ears, nose) and we would expect more importance given to those pixels. The ``importance'': of a pixel would be simply measured by the weight the trained NN assigns to that pixel in the case of a single-layer perceptron, but in the case of Deep Neural Networks, a highly non-linear operation is performed on the input data, and the importance, or saliency, of an input feature, or pixel, is harder to assess. \tatiana{Saliency maps have been extracted from CNNs and studied in earlier works, chiefly in \cite{Lee_2016_CVPR}. Subsequently, saliency maps have been used as a tool for improving the efficiency of the performance of the CNNs (see for example \citealt{DBLP:journals/corr/abs-2105-00937}).  Within the field of transient detection,   \cite{Reyes_2018}, have  leveraged saliency maps to identify the most relevant pixels and improve performance on transient classification. We go one step further and use the saliency maps to investigate how the model leverages the DIA image and what information the model uses in its absence.}

We follow \cite{simonyan2014deep}'s definition of saliency: denoting the class score with $S_c$, such that the NN output on image $\mathcal{I}$ is represented by $S_c(\mathcal{I})$, the saliency map is given by :
\begin{eqnarray}
 S_c(\mathcal{I}) \approx w^T \mathcal{I} + b,\\
 w = \frac{\partial S_c}{\partial \mathcal{I}} \biggr\rvert_{\mathcal{I}_0}.
 \end{eqnarray}
 
That is: $w^T$ is the weight and $b$ the bias of the model. $w$ is the derivative of $S_c$ with respect to $\mathcal{I}$ calculated specifically in the local neighborhood of pixel $\mathcal{I}_0$, and the approximation sign indicates a first order Taylor expansion has been used to approximate the solution. Each pixel in an image in the training set is associated with the corresponding pixel in the saliency map, and the saliency score (the importance) of that pixel measures the change in model output as a function of changes in the value of that input pixel by back-propagation. The higher the value of a pixel in a saliency map, the more influence that pixel has in the final classification. In this work we also refer to these maps as maps of pixel importance.

\begin{figure}
    \centering
    \includegraphics[width=1\linewidth]{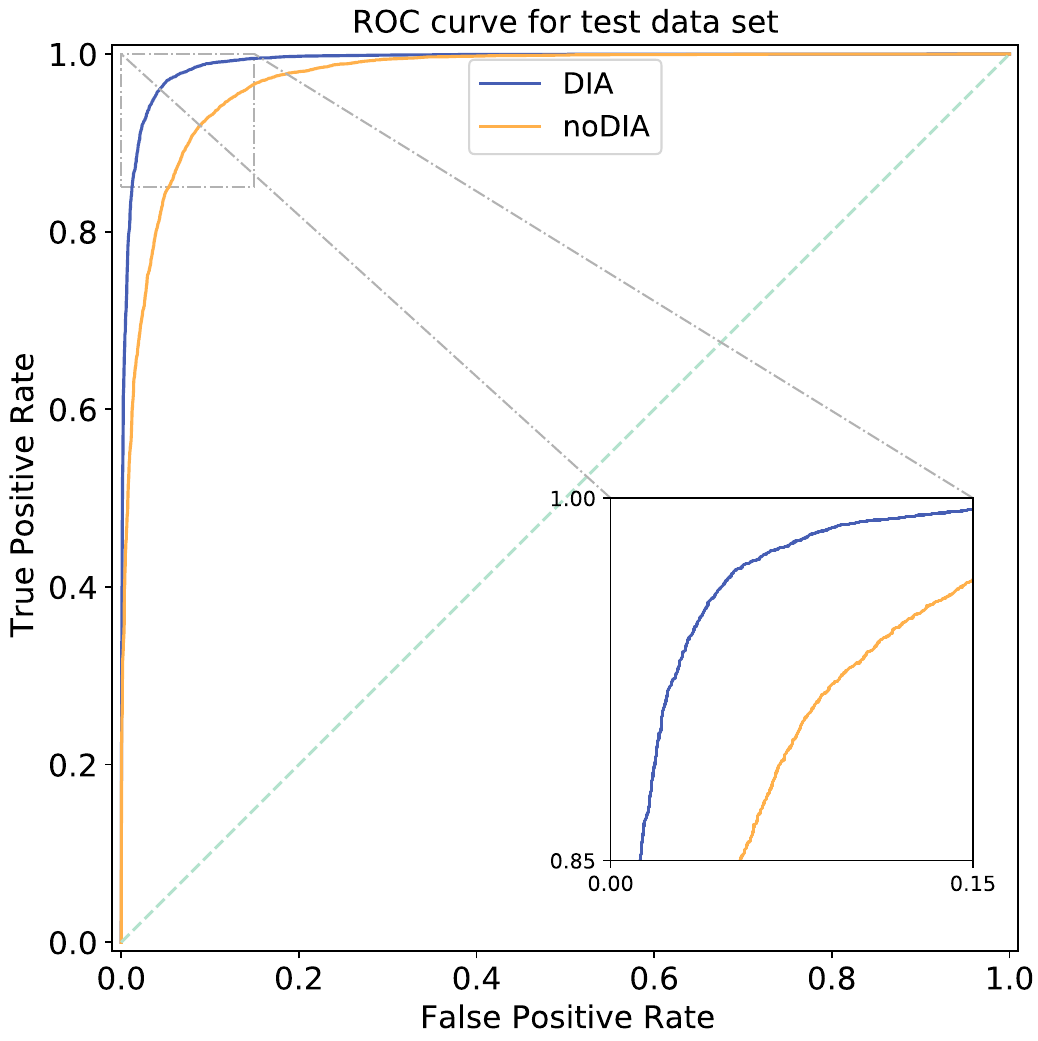}
    \caption{ROC curve for the $20,000$ images used for testing, Blue line for the  \diabased\ model, the area under the curve is $0.992$. Orange line, for the \nodia\ model, the area under the curve is $0.973$. This figure is discussed in \autoref{sec:results}.}
    \label{fig:roc_models}
\end{figure}

\begin{figure*}
    \centering
    \includegraphics[width=0.9\linewidth]{
    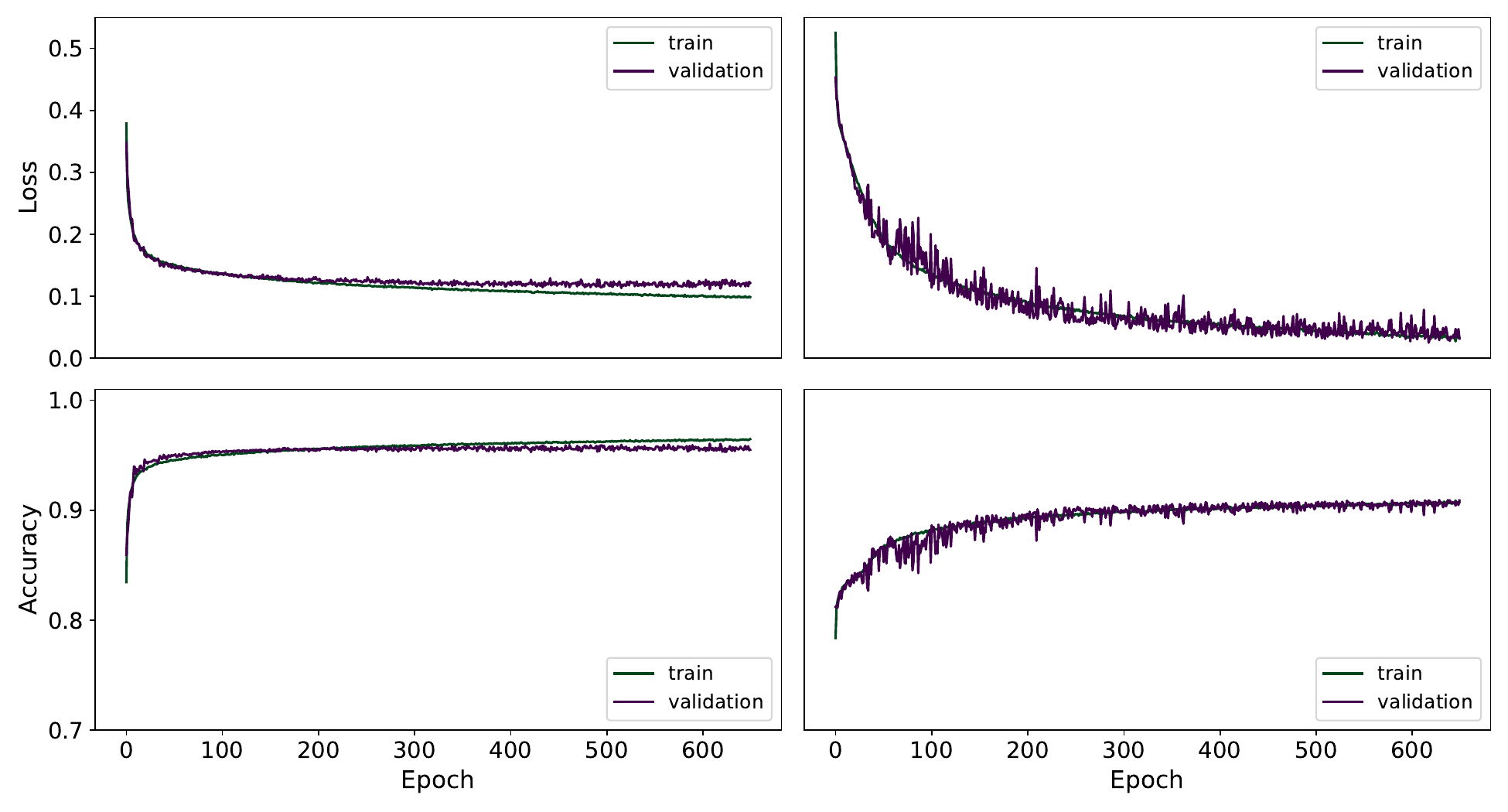}
    \caption{Loss (\emph{top}) and accuracy (\emph{bottom}) for both models presented in this work. \tatiana{Details of the loss function, early stop and saving strategies, and optimization method are available in \hyperref[sec:appendixb]{Appendix B}}. Purple lines correspond to the validation data (20\% or 20,000 images) and green to the training data (80\% or 80,000 images). {\it Left}:  results for our \diabased\ model  (\autoref{fig:architecturesCNN}). {\it Right}: results for the \nodia\ model (\autoref{fig:architecturesCNN}). %This figure is discussed in \autoref{sec:results}.
     \new{Both models show convergence at a >90\% accuracy with no major overfitting. This figure is discussed in detail in \autoref{sec:results}.}}.
    \label{fig:loss_models}
% \end{figure}
\end{figure*}

In our case, given the side-by-side organization of the three elements of the input image set, %(\diff, \search, \temp), 
the saliency maps can help us assess how much the \diabased\ model relied on the \diff\ to enable correct classification, and, thus, provide some intuition in the difficulty of the challenge offered to the \nodia\ model.

For the \diabased\ model we have an expectation guided by intuition: that a greater concentration of important pixels should be found in the \diff\ image. In \autoref{sec:results},  we will consider the veracity of this hypothesis both qualitatively by visually inspecting the saliency maps, and by designing a saliency-based metric that enables a quantitative approach. We calculated the normalized sum of the saliency pixel values for each third of the image-triplet, corresponding to \diff, \search, \temp. \round{Indicating with $I$ the importance of the segment of an image, ($\idiff, \isearch, \itempl$), with $p$ a pixel and $s_p$ its corresponding saliency value, and utilizing the subscript $d$, $s$ and $t$, to refer to pixels respectively in the \diff, \search\ and \temp, we have: }

\begin{eqnarray}
\begin{split}
\idiff=\frac{\sum_{p_d}{s_{p_d}}} {\sum_{p}{s_p}};\\
\isearch=\frac{\sum_{p_s}{s_{p_s}}} {\sum_{p}{s_p}};\\
\itempl=\frac{\sum_{p_t}{s_{p_t}}} {\sum_{p}{s_p}}.
\label{eq: saliencymetric}
\end{split}
\end{eqnarray}

% in the Where $I$ indicates the importance of a segment of the input image. 
\round{The numerators capture the importance of each third of an image while
%, $p_d$ denotes the pixels in the leftmost third of the image, corresponding to \diff; $p_s$ the middle third corresponding to the \search.
%and $p_t$ the right most third, corresponding to the \temp. $s_{p_x}$ indicates the value of pixel $p_x$ in the saliency map in the $x$ segment (\diff\ , \search\ or \temp). 
the denominator normalizes each metric by the total sum of the saliency pixel values, so that 
$\idiff + \isearch + \itempl = 1$.}

% $p$ is an index running over all pixels and $s_p$ indicates the value of pixel $p$ in the saliency map (the concatenation of the three segments).}

% \begin{eqnarray*}
% \begin{split}
% \round{\idiff = \frac{\sum_{p_d \in P_d}s_{p_d}}{\sum_{p\in P}s_p};}\\
% \isearch=\frac{\sum_{p_s \in P_s}s_{p_s}}{\sum_{p\in P}s_p};\\
% \itempl= \frac{\sum_{p_t \in P_t}s_{p_t}}{\sum_{p\in P}s_p};
% \label{eq: saliencymetric_refereesugg}
% \end{split}
% \end{eqnarray*}
% \begin{eqnarray*}
% \begin{split}
% \round{\idiff = \frac{\sum_{p \in D}s_{p\in D}}{\sum_{p}s_p};}\\
% \isearch=\frac{\sum_{p_s \in P_s}s_{p_s}}{\sum_{p\in P}s_p};\\
% \itempl= \frac{\sum_{p_t \in P_t}s_{p_t}}{\sum_{p\in P}s_p};
% \label{eq: saliencymetric_refereesugg}
% \end{split}
% \end{eqnarray*}

% \round{where $P_d$ is the set of indexes in all pixels in the left-most third of the image, and $P$ is the set of indexes of all pixels in the saliency map.
% }

% where $I$ indicates the importance of a segment of the input image, $p$ is an index running over all pixels and $s_p$ indicates the value of pixel $p$ in the saliency map. The denominator normalizes each metric by the total sum of the saliency pixel values. In the numerators, $p_d$ denotes the pixels in the leftmost third of the image, corresponding to \diff; $p_s$ the middle third corresponding to the \search; and $p_t$ the right most third, corresponding to the \temp. 
This metric allows us to assess the relative importance of the \diff\ (\search\ or \temp) component of the image in performing RB classification. Results from these metrics are discussed in detail in \autoref{subsec:results_seliancy}.

%% file: result.tex
% \subsection{Neural Network}

The accuracies of our models \newtwo{and their respective errors, calculated as the standard deviation,}  %the model for DIA and no\diabased\  using the $80,000$ for training and the $20,000$ for testing 
are presented in \autoref{tab:acc results}. The \diabased\ model reached, by design, the accuracy of our benchmark model \texttt{autoscan}: 97\% on True Negative (TN) rate and 95\% on True Positive (TP) rate (see \autoref{sec:method}). We remind the reader that we use the \texttt{autoscan} convention for the definition of TN and TP: Positive is a ``real'' transient ($\texttt{label = 0}$), negative is a ``bogus'' ($\texttt{label = 1}$). There is a drop of $\sim 4\%$ between the accuracy of the \diabased\ and the \nodia\ models. Along with the accuracy, in \autoref{tab:acc results} we present computational costs of training on 20,000 images measured in CPU Node hours,\footnote{The computational cost was calculated by using the formula given in \url{https://docs.nersc.gov/jobs/policy/} and the allocations information given by the system Iris \url{https://iris.nersc.gov}. Training time is reported in CPU hours. Prediction time is reported per “postage stamp” (1ms) and the classification operation is trivially parallelizable (can be run independently on each postage stamp).} and the clock time for the prediction for one single image.

\begin{table}
\footnotesize{
    \centering
    \begin{tabular}{|c|c|c|c|c|c|}
    \hline
    \multirow{2}{*}{\bf{Model}}& \multicolumn{3}{c|}{\centering \bf{Accuracy}}
 & \bf{Training}$^8$ & \bf{Prediction}$^8$\\
    \bf{ }&\bf{Train} & \bf{  Test} & \bf{Val}& \bf{CPU } & \bf{Clock-time}\\
    \bf{ }&\bf{} & \bf{  } & \bf{}& \bf{(hours)} & \bf{(ms)}\\
    \hline
       \multirow{2}{*}{ \diabased}&\multirow{2}{*}{0.965}&$0.961\pm$&\multirow{2}{*}{0.960}&\multirow{2}{*}{$\sim$35}&\multirow{2}{*}{$1.00\pm0.03$}\\&&$0.004$&&&\\ \hline
      \multirow{2}{*}{ \nodia}&\multirow{2}{*}{0.920}&$0.911\pm$&\multirow{2}{*}{0.914}&\multirow{2}{*}{$\sim$56}&\multirow{2}{*}{$0.30\pm0.01$}\\&&$0.005$&&&\\ \hline
    \end{tabular}
    \caption{Comparative table of the accuracy and computational cost for training, testing and validation data for the \diabased\  (\autoref{fig:architecturesCNN}, left) and \nodia\  (\autoref{fig:architecturesCNN}, right) models. }
    \label{tab:acc results}}
\end{table}

Confusion matrices for the testing set are shown for the \diabased\ model in left panel of  \autoref{fig:confusiomatrix_models} and for the \nodia\ model in the right panel of the same figure. In this figure, as in the following confusion matrices and histograms that we will present in \autoref{subsec:results_seliancy}, correct predictions are indicated in shades of blue, incorrect in shades of orange, and darker shades are associated with the true labels. The percent accuracy for each class, True Positives (TP), True Negatives (TN), False Positives (FP), and False Negatives (FN), as well as the number of images in each class are reported within the figure.

The Receiver Operating Characteristic (ROC) curve shows the relation between the True Positive Rate (TP / (TP + FN)) also know as recall, and the False Positive Rate (FP / (FP + TN)) when changing the threshold value (\eg, a threshold of 0.5 will indicated that values greater than 0.5 would be classified as ``bogus''). The ROC for the testing data for the \diabased\ and \nodia\ models are presented in \autoref{fig:roc_models}. The Area Under the Curve (AUC), which can be used as a comprehensive metric of the aggregated classification performance of a model \citep{hanley1983method, hernandez2012unified}, is 0.992 and 0.973 for the \diabased\  and \nodia\ models respectively. %, demonstrating good performance for both models \tatiana{and within \texttt{autoscan} performance}.

The loss and the accuracy curves in the left panel of \autoref{fig:loss_models} for the \diabased\  model shows some evidence of overfitting (the validation curve flattens compared to the training curve) starting just a few epochs before training came to end of the 350-epochs; meanwhile, for the \nodia\ model in right \autoref{fig:loss_models}, after $650$ epochs there was no visual evidence of overfitting indicating that the model is still learning generalized information from the data; yet the accuracy improvements from epoch $350$ to $650$ were small.

The nature of the \nodia\ model leads to hypothesize that because the input data contains less information, this model takes longer to learn features from the data to be able to classify them. The \nodia\ model took in fact longer (more epochs) to come to stable accuracy and loss values. The loss and accuracy curves are also noisier for the validation of the \nodia\ model in right \autoref{fig:loss_models} compared to the \diabased\ model. This also can be explained with the same argument: the \nodia\ model had a harder problem to solve and this is reflected in a noisier path to minimization. We conclude that this $\sim$ 4\% loss in accuracy is directly related to \new{the loss of information in the input} caused by dropping the \diff\ in input. In addition, we tested if longer training or slightly richer architectures could make up for the loss of \diff\ and found that neither extending the training  beyond 650 epochs or adding convolutional layers improved performance \round{(see \hyperref[sec:appendixb]{Appendix B.1}, \autoref{tab:noDIACNNextracn_architecture}, \autoref{tab:noDIACNNextracndense_architecture}}).
%is less linear. 
% We cannot however rule out, and in fact we suspect, that the hyperparameters or the architecture selected for the \nodia\ CNN model could be improved --- as this is a proof of concept paper and the architecture was chosen to enable a direct comparison with the \diabased\ model, future work will include experimenting with different architectures and an extensive grid-search to optimize hyperparameters.
\begin{figure*}
    \centering
    \includegraphics[width=0.9\linewidth]{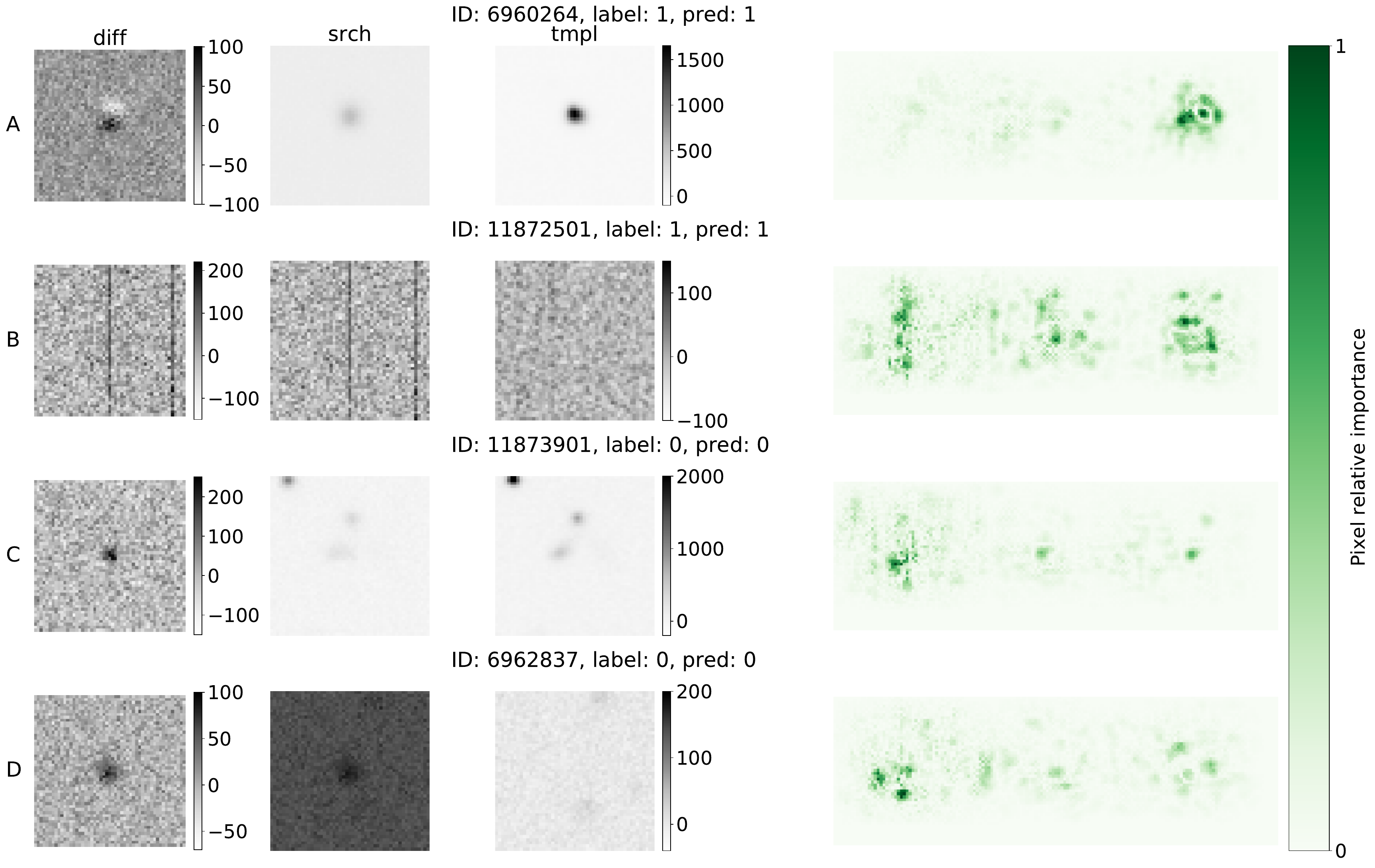}
    \caption{Saliency map for the four transients in \autoref{fig:examples_no_normalization} for the \diabased\ model. On the left, in grey color, the original \diff, \search, and \temp\ images are plotted in their natural flux scale (before normalization). On the right, the saliency map for the combined image. The intensity of a pixel color in the white-to-green scale indicates the pixel relative importance: the maps are normalized to 1 individually, such that dark green corresponds to high saliency score, with 1 corresponding to the most important pixel in the image triplet. With the side-by-side organization of the input data, these maps enable a visual understanding of the importance of each element of the combined image in the real-bogus classification. We note how in some cases (panel A) the decision is largely based on the \temp, rather than the \diff\ image, and in some cases all three image elements contribute similarly to the decision (panel B). This figure is discussed in more detail in \autoref{subsec: saliency}.}
    \label{fig:saliency_4id}
\end{figure*}

\begin{figure*}
    \centering
    \includegraphics[width=0.84\linewidth]{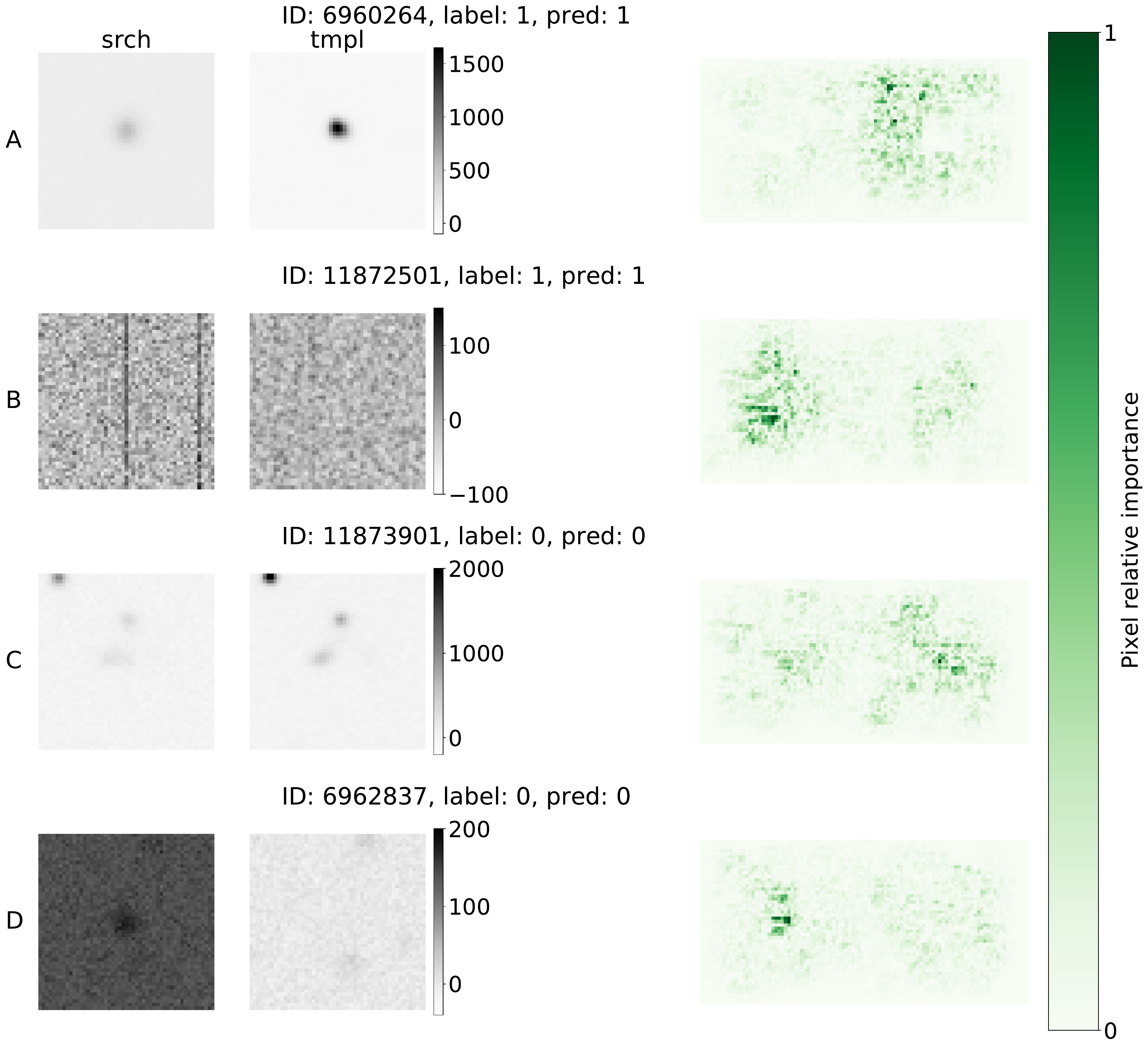}
    \caption{As \autoref{fig:saliency_4id} but for the \nodia\ model: saliency map for the four transients in \autoref{fig:examples_no_normalization}. This figure is discussed in more detail in \autoref{subsec: saliency}.}. 
    \label{fig:saliency_4id_noDIA}
\end{figure*}

% \begin{figure*}
% \begin{minipage}[b]{0.45\linewidth}
%     \centering
%     \includegraphics[width=1\linewidth]{
%     figures/ROC3D.pdf}
%     \caption{ROC curve for the $20,000$ images used for testing the \textbf{\diabased\ }model. The area under the curve was $0.992$.}
%     \label{fig:roc_modelAAA}
% % \end{figure}
% \end{minipage}
% \begin{minipage}[b]{0.45\linewidth}
% % \begin{figure}[H]
%     \centering
%     \includegraphics[width=1\linewidth]{
%     figures/ROC.pdf}
%     \caption{ROC curve for the $20,000$ images used for testing the \textbf{\nodia\ }model. The area under the curve was $0.973$.}
%     \label{fig:roc_modelCCC}
%     \end{minipage}
% \end{figure*}
\subsection{A peek into the model decisions through saliency maps}\label{subsec:results_seliancy}

% \masao{This is a very interesting section!  For readers like me who don't know much about how saliency maps are produced, can you write a couple of sentences about how they are made?  I think you said something about removing each pixel and looking at how much the predictions change.  A short description is helpful for the uneducated reader like me.}

\begin{figure*}
        \centering
    \includegraphics[width=0.45\linewidth]{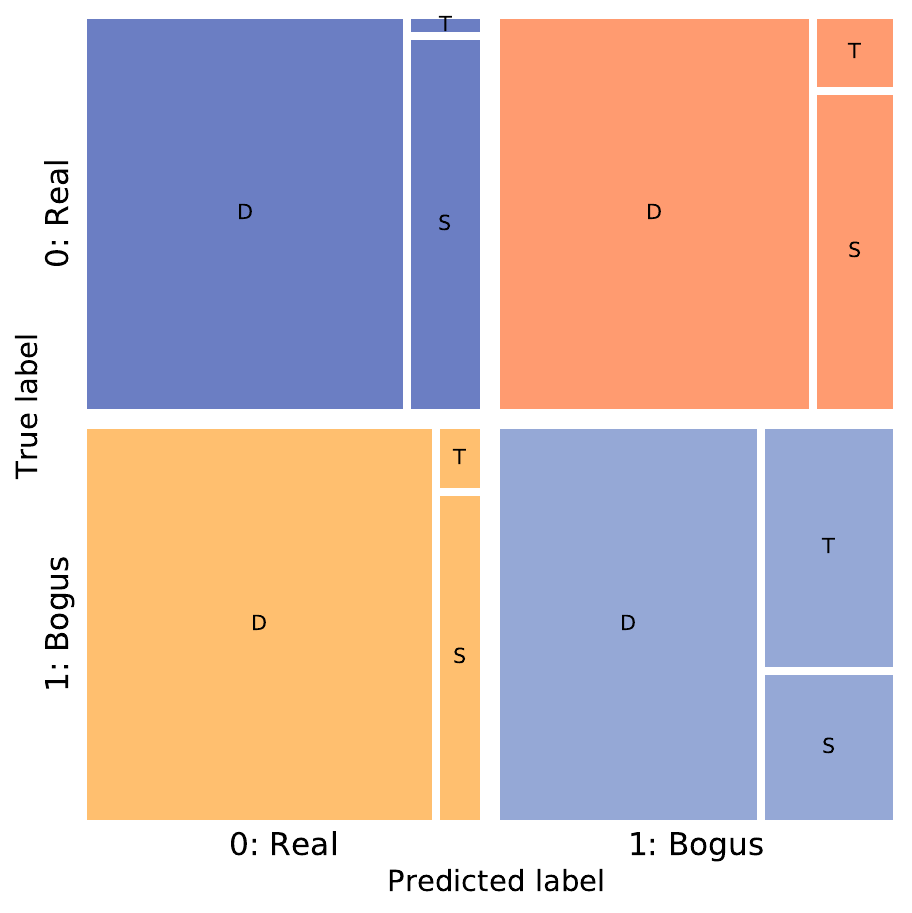}
     \includegraphics[width=0.45\linewidth]{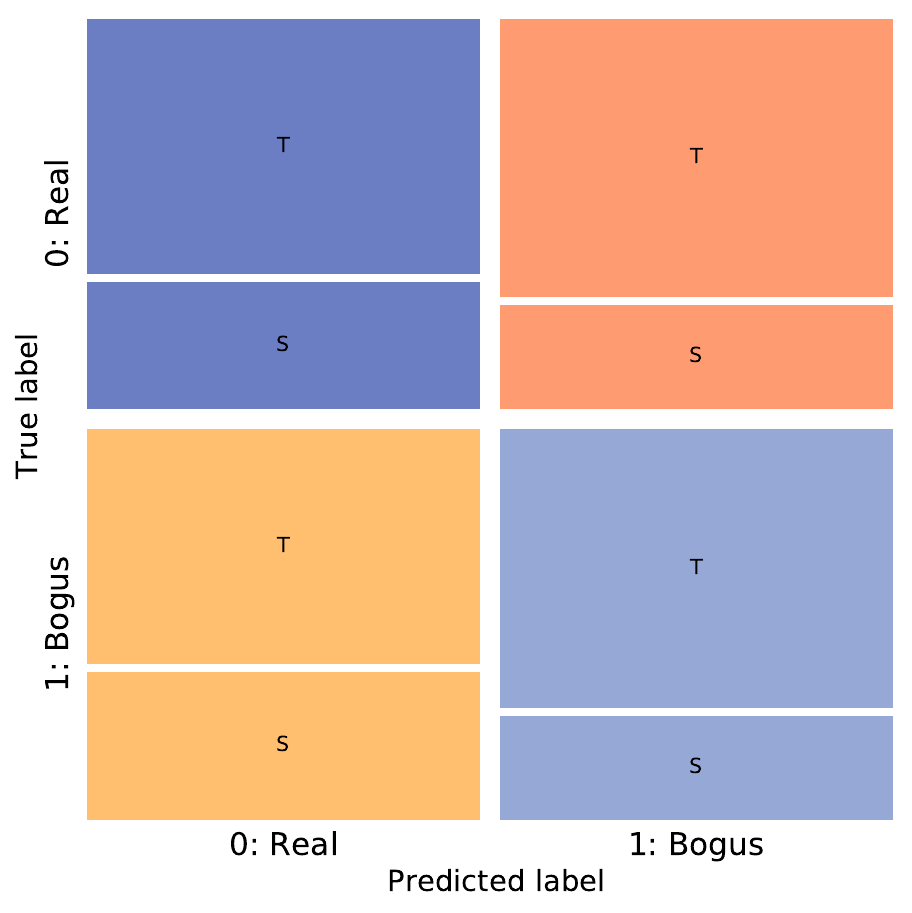}
    \caption{Confusion matrix reporting the proportion of transients for which the highest concentration of important pixels is found in the \diff, \search, or \temp\ portion of the input image for the \diabased\ model results (left) and \nodia\ model (right), {\it Left}: \diabased. For 80\% ($7683$) of the $9571$ transients classified correctly as ``real'', the classification principally relied on the the \diff\ image ($I_\diff > 1/3$); for 18\% ($1758$) on the the \search ;  and for 1\% ($130$) on the the \temp . For incorrect ``real'' classifications 88\% of the images relied principally on \diff\,10\% on \search, and 2\% on \temp. For incorrect ``bogus'' classifications 79\% of the images relied principally on \diff\, 17\% on \search, and 4\% on \temp. For correct ``bogus'' classifications 66\% of the images relied principally on \diff\, 13\% on \search, and 21\% on \temp. {\it Right}: \nodia. Of the $9163$ transients classified correctly as ``real'', for 66\% of them, the classification relied principally on the \temp\ image. For the incorrect ``real'' classifications 61\% of the cases were principally based on the \temp. For the incorrect ``bogus'' classifications 71\% of the cases were principally based on the \temp. For correct ``bogus'' classifications 72\% of the cases were principally based on the \temp. }
    \label{fig:saliency_confusions}
\end{figure*}

%A gradient of the final output with respect to the input data is the operation used to calculate the score. 
%A saliency map provides a score to each one of the pixels that form the image used for training the CNN base on how important is that pixel to the final classification of the CNNs. A gradient of the final output with respect to the input data is the operation used to calculate the score. 
%For the purpose of this work the score range is not relevant, we focus only on the value given by the code and that a higher value implies a more relevant pixel for the classification. A higher score of a pixel would indicated that a small change in that flux value would affect the final output, and consequently the final classification decision. 
%Our expectation is that pixels scored highly are located in the object that we want to classify. 

\begin{figure*}
    \centering
    \includegraphics[width=0.45\linewidth]{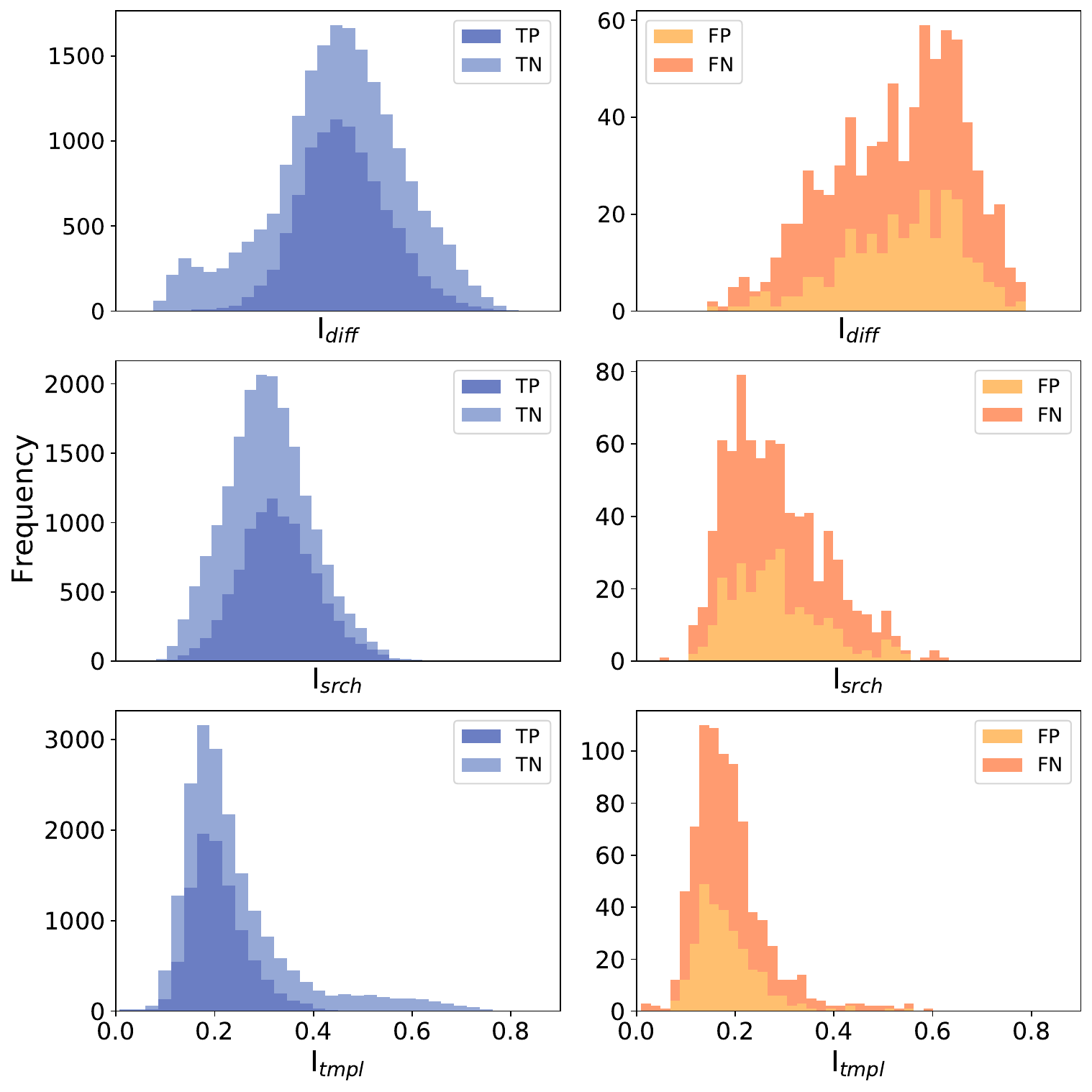}
    \includegraphics[width=0.45\linewidth]{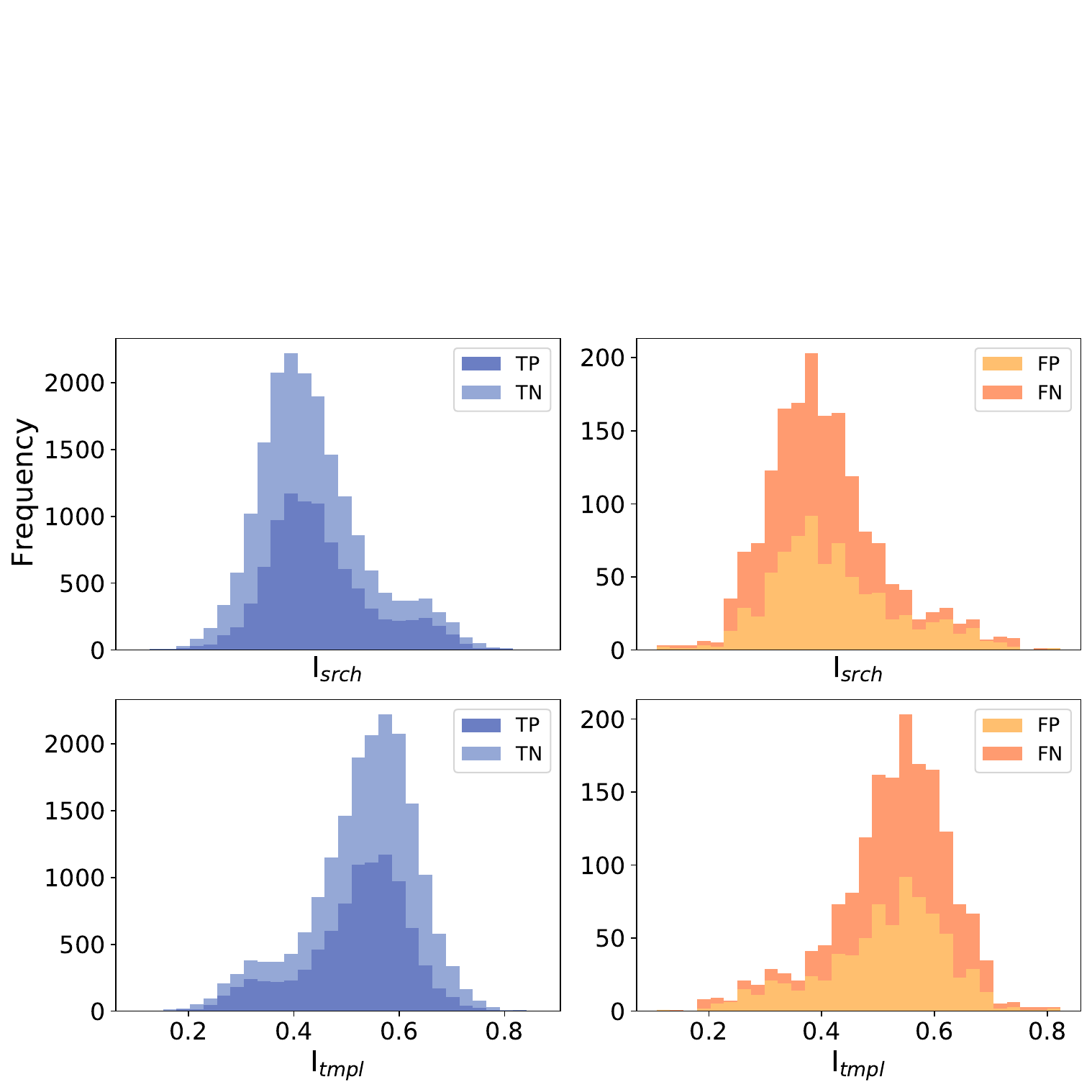}
    \caption{The distribution of  $\idiff$ (top), \isearch\ (middle), and \itempl\ (bottom) value for the 20,000 transients in the test data (see \autoref{eq: saliencymetric}). The colors corresponds to the quadrants of the confusion matrix to which the transient belong according to the model prediction, \diabased\ predictions in the first and second column from the left to the right and \nodia\ in the third and fourth. Blue shades correspond to correct predictions (TP and TN) and orange to FP and FN. Note that the $y$ axis values are different in each plot and the FP and FN histograms contain far fewer observations. On the {\itempl}, for the \diabased\ model (\emph{left}), the TP classification shows a preference for the \diff\ images and the distribution peaks at $\idiff \sim 0.5$. For TN, however, small $\idiff$ values are more common, with an significant fraction of observations in the $0<$\idiff$<0.33$ region. This behaviour is complemented by a long \itempl\ tail in the TN distribution ($0.4<\itempl<0.8$). The FP and FN distributions are qualatively similar to the TP and TN, but noisier, as they contain fewer than 10\% of the objects. {\it Right}: \nodia\ model. All four classes have qualitative similar distributions in \itempl and \isearch. Classifications rely mostly on the \temp\ in all cases ($\itempl > 0.5$). \new{The mathematical meaning of the plotted quantities is described in detail in \autoref{subsec: saliency}. This figure, is discussed in \autoref{subsec:results_seliancy}}.}
    \label{fig:histo_saliency}
\end{figure*}

In \autoref{subsec:scaling}, we described how the importance of individual image pixels in the RB prediction performed by our models can be measured, and the design of a saliency-based metric to assess which component of the image is most important to perform the RB classification. Here we inspect the saliency maps, both visually and quantitatively through the measured values of \idiff, \isearch, and \itempl.

In \autoref{fig:saliency_4id} and \autoref{fig:saliency_4id_noDIA} we show the four transients we considered as examples throughout this work, the same images used in \autoref{fig:examples_no_normalization} and \autoref{fig:examples_hstack_normalization}, and the corresponding saliency maps for the \diabased\ model and \nodia\ model respectively. \autoref{fig:saliency_confusions} and \autoref{fig:histo_saliency} report the results of \autoref{eq: saliencymetric} for the objects in our training set.

Lets start with some considerations about the saliency maps for the four image examples for the \diabased\ model (\autoref{fig:saliency_4id}), and specifically from  panels C and D were the transients were correctly predicted as ``real''. We observe that the greatest concentration of important pixels for both these images is found in the left-most third of the image: %the \diff\ section, \question{approx the 50\% of all the pixels that form the triple in the saliency map}, meaning that
$\idiff \sim 0.5 $ for both. 
%, according to the definition in \autoref{eq: saliencymetric}.
We speculate, from our experience in labeling real/bogus by visual inspection and consulting with some of the human scanners that labeled the original \texttt{autoscan} images, that this behaviour is similar to what a human scanner would do: if in the \diff\ image there is a clearly real transient, the scanner would not need to study in detail the \search\ and \temp\ images. 

\autoref{fig:saliency_4id}A and B show correctly classified ``bogus'' transients. In \autoref{fig:saliency_4id}A, a ``bogus'' produced likely by a moving object displaying a classical dipole, the majority of the important pixels are located in the \temp\ ($\itempl \sim 0.54 $), concentrated around the location of the central source and the location  where its ``ghost'' image is (the coordinates corresponding to the location the bright patch of pixels in the \diff, but in the \temp\ portion of the composite image).
In \autoref{fig:saliency_4id}B, there is no central source and the detection is triggered by an image artifact. The important pixels are found in all three image segments and are spread around a large area of each image: the model has inspected the image in its entirety to decide the classification. 
Following our considerations about the similarity between the CNN and human decision process, \new{having discussed the typical visual inspection process with members of the DES team that human-labeled this data set, we find that, here too, the CNN  mimics closely what a human scanner would do}: because there is not a clear central source (a ``real'' object) in the \diff, the scanner \new{wouldn't simply draw a conclusion based on the \diff\, but instead would} analyze the \search\ and \temp\ images to extract more information from the context and enable a \new{robust} classification. However, it should be noted that no quantitative studies of the features the human scanners use to classify transients has been done, thus this remains simply an intriguing suggestion.

For the case of the \nodia\ model, the expectation was less clear: both the \search\ and the \temp\ images are necessary to ``reconstruct'' the information contained in the \diff, and while the pixels overlapping with the central transient are obviously expected to be important, the pixels that surround it are necessary to essentially reproduce the scaling and PSF-matching operations between \temp\ and \search\ that the DIA performs. 
Accordingly, the saliency maps presented in \autoref{fig:saliency_4id_noDIA} are more difficult to interpret: in all four cases, important pixels are found all over the composite images.
%the results, that are presented graphically in right \autoref{fig:histo_saliency} and right \autoref{fig:saliency_confusions}, are more difficult to interpret.

To explore how the choice of important pixels may depend on the image label and on the correct classification we report  the fraction of images for which the \diff\ (\search, \temp) is the dominant source of important pixels {\it within} the confusion matrix in \autoref{fig:saliency_confusions}. To do this, we use a rough but intuitive cutoff: if the normalized sum of the saliency pixels in a third of the image is larger than $\frac{1}{3}$, then we deduce that the model principally used that component for its decision. For example, where $\idiff > 0.33$, we conclude that the model principally relied on \diff\ to make the RB classification. With this cut-off we can assess if there are differences in the model behavior when classifying objects as a function of their labels or their classification.
In all four cases (all combinations of ``real'' and ``bogus'' label and prediction) the concentration of important pixels is largest in the portion of the image corresponding to the \diff\  in the \diabased\ model. It is however interesting to note that, in order to correctly classify the ``bogus'', the \diabased\ model uses the template and search image more heavily than in all other cases (\diff, \search, \temp,  = 66\%, 13\%, 21\% for TN, while $I_\diff>80\%$ for TP, FP, and FN). 

The cut-off method described above does not allow us to distinguish between cases where multiple sections of the images were used jointly, perhaps with similar importance, from cases where the model truly only relied on one section of the image. For that, we take a closer look at the distribution of saliency values. 
In \autoref{fig:histo_saliency} we show the distribution of values of the three metrics defined in \autoref{eq: saliencymetric} for each of the four cases: TP and TN, in shades of blue, FP and FN, in shades of orange, following the color-scheme adopted in \autoref{fig:confusiomatrix_models} and \autoref{fig:saliency_confusions}. For the \diabased\ model (the two left-most columns), for the majority of the 20,000 images in the training set $\idiff > 0.33$, but there is a secondary pick in the $\idiff$ distribution near $\idiff \sim0.1$ populated entirely by TN cases, complementary to a long right tail in the \itempl distribution ($\itempl > 0.4$). This confirms that the correct classification in the presence of real transients relies on \diff, but \temp\ and \search\ become important to correctly classify bogus transients, just like we had seen in the exemplary cases in \autoref{fig:saliency_4id}. We note also that the general shape of each distribution (\idiff, \isearch, \itempl) is similar for the TP and TN case (blue) and for the FP and FN cases (orange). 
 
For the \nodia\ model the important pixels are concentrated in the \temp\  for most images ($\itempl > 0.5$) for both correct and incorrect classifications.
This is somewhat counter-intuitive since the \temp\ {\it does not} contain the transient itself! However, one may speculate that this is because the \temp, a higer quality image, contains more accurate information about the context in which the transient arises: \eg, if it is located near a galaxy or not. This information is important to the classification. It is also interesting to note that for the transients predicted as ``real'' in both TP and FP, the fraction of images that leveraged primarily the \temp\ is approximately $2/3$, and for images predicted as ``bogus'', in both TN and FN, it is approximately $3/4$.

To help guide the interpretation of the saliency maps, a few more maps are plotted in \hyperref[sec:appendixc]{Appendix D}. Where we provide 6 examples per each class of the confusion matrix, for both the \diabased\ and the \nodia\ models. 

\subsection{Computational cost of our models}\label{sec:computationcost}
The computational cost of our models, reported in CPU Node hours in \autoref{tab:acc results}, confirms that while training a CNN model for RB can be computationally expensive, and significantly more so if the \diff\ is not used in input (\nodia), the model prediction \new{only takes a few seconds even on large datasets}. Using a NN-based platform, the computational costs are front-loaded. In transient detection this could mean that the observation-to-transient discovery process \new{can be rapid}, while computation time can be spent during off-sky hours (principally to build templates). Furthermore, while the training time is longer for the \nodia\ model than for the model that uses the \diff\ in input (\diabased) the computational cost of the forward-pass (prediction) scales superlinearly with the size of the feature set (pixels), so that our \nodia\ model takes less than half the time than the \diabased\ one to perform the RB task. With a clock time of 0.3 ms per 51$\times$51 pixels postage stamp, predicting over the full DES focal plane would take $\sim 1$ minute. However, we note that at this stage of our work we still rely on the DIA in several ways: while the \temp\ and \search\ in input to \nodia\ are not PSF matched, this proof of concept is performed on transients that were detected in \diff\ images, and we leverage the alignment of \temp\ and \search\ and centering of the postage stamp that arose from the DIA (see \autoref{sec:futurework}). Conversely, predictions would only need to be done in correspondence of the sources detected in the \temp\ and \search\ images, and not for the entire CCD plane.

% \begin{figure*}
% \begin{minipage}[b]{0.45\linewidth}
%     \centering
%     \includegraphics[width=0.8\linewidth]{
%     figures/confusionmatrix_saliencyDIA(1).pdf}
%     \caption{Confusion matrix and the proportion of the most important pixels given by the saliency map of each of the images in the triple after the training using the \textbf{\diabased\ }model. i.e., from the $9564$ transients classified correctly as ``real'', 90\% ($8630$) of them, the classification was done based principally on the \diff\ image due to the fact that the sum of pixels in the \diff\ section in the saliency map was the greatest of the three. }
%     \label{fig:saliency_confusionDIA}
% \end{minipage}
% \begin{minipage}[b]{0.45\linewidth}
%     \centering
%     \includegraphics[width=0.8\linewidth]{
%     figures/confusionmatrix__saliencynoDIA.pdf}
%     \caption{Confusion matrix and the proportion of the most important pixels given by the saliency map of each of the images in the duplex after the training using the \textbf{\nodia\ }model. i.e., from the $9147$ transients classified correctly as ``bogus'', 24\% ($2197$) of them, the classification was done based principally on the \search\ image due to the fact that the sum of pixels in the \search\ section in the saliency map was the greatest of the two. }
%     \label{fig:saliency_confusionnoDIA}
%     \end{minipage}
% \end{figure*}

%% file: limitations.tex
This work is targeted to the investigation of CNN RB model performance, with and without \diff\ in input, on a single data set, the same dataset upon which the  development of the random-forest-based \texttt{autoscan} was based (see \autoref{sec:data}). This approach enables a straightforward comparison, but it comports some limitations. 

The labels in our data set come from simulations of SNe ($\texttt{label = 0}$) and visual inspection that classifies artifacts and moving objects ($\texttt{label = 1}$). We reserve to future work the investigation of the efficacy of the model on transients of different nature, including quasars (QSOs), strong lensed systems, Tidal Disruption Events (TDEs), and supernovae of different types.  These transients may have characteristically different associations with the host galaxy, including preferences for different galaxy types and locations with respect to the galaxy center, compared to the simulated SNe events in our training set. 

Specifically thinking of Rubin LSST data, an additional source of variability may be introduced by Differential Chromatic Refraction effects \citep{Abbott_2018, richards2018leveraging}, or stars with significant proper motion which, due to exquisite image quality of the Rubin images, would be detectable effects.

While we demonstrated CNN model's potential in the detection of transients without DIA, we did not address the question of completeness as a function of \search\ or \temp\ depth or the potential for performing accurate photometry without DIA. 

\new{We trained multiple network architectures to attempt to improve our prediction performance, as discussed in \autoref{sec:method}. None of the models we trained achieved an accuracy that significantly outperformed the random forest-based \texttt{autoscan}, and none of the \nodia\ versions of our models compensated for the $\sim4\%$ loss induced by the removal of the \diff\ in input. Our efforts included training the original \nodia\ for many more epochs, tuning hyperparameters, adding convolutional layers, and adding layers after the last convolutional layer but before the flatten layers to reduce bottlenecks. We intend to continue investigating alternative architectures in future work. }%rotation invariance, 
%, {\it etc}., and we leave these tasks as future work beyond the proof-of-concept presented here. %The optimization of the architecture would be included for a future work. 

Finally, we note that our models did in fact implicitly leverage some of the information generated by the DIA even when they did not use the \diff\ image itself as input.  First, the \temp\ and \search\ images are de-warped. Since the transient alerts are generated from aligned DIA images, the transient source is always located at the image center in our data. We use postage stamps that were, however, not PSF matched or scaled to match the template brightness.
To move beyond a proof-of-concept, in future work we will re-train and apply our models to images whose alignment does not depend on the existence of a transient. %This would support strongly that the used of the \diff\ image is not necessary because nowadays the position of the source in the image-triplet is done based in the \diff\ image.

%% file: conclusion.tex
In this work, we have \new{measured the accuracy loss associated with the removal of the \diff\ image in input to a CNN trained in }
%demonstrated that high-accuracy models for 
classifying true astrophysical transients from artifacts and moving objects (a task generally known as ``real-bogus''). \new{We have demonstrated that, while a model with $\sim91.1\%$ accuracy} can be built without leveraging the results of a difference image analysis (DIA) pipeline that constructs a ``template-subtracted'' image, \new{there is a loss of performance of a few percentage points.}

Starting from the Dark Energy Survey dataset that supported the creation of the well known real-bogus \texttt{autoscan} model \citep{Goldstein_2015}, we first built a CNN-based model, dubbed \diabased, that uses a sky template (\temp), a nightly image (\search), and a template-subtracted version of the nightly image (\diff) that performs real-bogus classification at the level of \texttt{autoscan} model.
%state-of-the art models.  
Our \diabased\ model reaches a $97\%$ accuracy in the bogus classification with an Area-Under-the-Curve of 0.992 and does not require human decision in the feature engineering or extraction phase.

We then created \nodia, a model which uses only the \temp\ and \search\ images and can extract information that enables the identification of bogus transients with $91.1\%$ \new{accuracy. We attribute this performance decrease directly to the {loss of information} about the PSF of the original image since, in addition to what is contained in the \temp\ and \search\ pair, the \diff\ contains information about the PSF used to degrade the \temp\ to match the quality of the \search\ image  
 (\autoref{sec:stateofart}). Thus, the convolutional architecture has been unable to recover that information.} %Our original \nodia\ model architecture is purposefully kept close to the architecture of the \diabased\ model to quantify the impact of change in input, but subsequent attempt to accuracy even with a naive model architecture that, to enable a direct comparison, is purposefully designed to be minimally different from our high accuracy \diabased\ model.

We further investigated what information enables the real-bogus classification in both the \diabased\ and \nodia\ models and demonstrated that a CNN trained with the DIA output primarily uses the information in the \diff\ image to make the final classification, and that the model examines a \diff-\search-\temp\ image-set fundamentally differently in the cases where there is a transient, than in the cases where there is not one. The \nodia\ model, conversely, takes a more comprehensive look at both \temp\ and \search\ images, but relies primarily on \temp\ to enable the reconstruction of the information found in the \diff.

Implementation of this methodology in future surveys could reduce the time and computational cost required for classifying transients by entirely omitting the construction of the difference images.  

%% file: acknowledge.tex
TAC and FBB acknowledge the support of the LSST Corporation and the Enabling Science Grant program that partially supported TAC through Grant No. 2021-040, Universidad Nacional de Colombia and University of Delaware for the Beyond Research Program (2020) and Summer Research Exchange (2021). MS and HQ were supported by DOE grant DE-FOA-0002424 and NSF grant AST-2108094. 

TAC: Prepared the data, conducted the analysis, created and maintains the machine learning models, wrote the manuscript.
FBB: Advised on data selection,  curation, and preparation,  selection and design, revised the manuscript.
GD: Advised on data preparation and model selection and design, revised the manuscript.
MS: Advised on data selection, shared knowledge about the compilation and curation of the original dataset, revised the manuscript.
HQ: Advised on data preparation and model design, revised the manuscript.

 This paper has undergone internal review in the LSST Dark Energy Science Collaboration.
 The internal reviewers were: Viviana Acquaviva, Suhail Dhawan, and Michael Wood-Vasey.

% This work used some telescope which is operated/funded by some agency or consortium or foundation ...
% We acknowledge the use of An-External-Tool-like-NED-or-ADS.
The DESC acknowledges ongoing support from the Institut National de 
Physique Nucl\'eaire et de Physique des Particules in France; the 
Science \& Technology Facilities Council in the United Kingdom; and the
Department of Energy, the National Science Foundation, and the LSST 
Corporation in the United States.  DESC uses resources of the IN2P3 
Computing Center (CC-IN2P3--Lyon/Villeurbanne - France) funded by the 
Centre National de la Recherche Scientifique; the National Energy 
Research Scientific Computing Center, a DOE Office of Science User 
Facility supported by the Office of Science of the U.S.\ Department of
Energy under Contract No.\ DE-AC02-05CH11231; STFC DiRAC HPC Facilities, 
funded by UK BIS National E-infrastructure capital grants; and the UK 
particle physics grid, supported by the GridPP Collaboration.  This 
work was performed in part under DOE Contract DE-AC02-76SF00515.

 This project used public archival data from the Dark Energy Survey (DES). Funding for the DES Projects has been provided by the U.S. Department of Energy, the U.S. National Science Foundation, the Ministry of Science and Education of Spain, the Science and Technology Facilities Council of the United Kingdom, the Higher Education Funding Council for England, the National Center for Supercomputing Applications at the University of Illinois at Urbana–Champaign, the Kavli Institute of Cosmological Physics at the University of Chicago, the Center for Cosmology and Astro-Particle Physics at the Ohio State University, the Mitchell Institute for Fundamental Physics and Astronomy at Texas A\&M University, Financiadora de Estudos e Projetos, Fundação Carlos Chagas Filho de Amparo à Pesquisa do Estado do Rio de Janeiro, Conselho Nacional de Desenvolvimento Científico e Tecnológico and the Ministério da Ciência, Tecnologia e Inovação, the Deutsche Forschungsgemeinschaft and the Collaborating Institutions in the Dark Energy Survey. The Collaborating Institutions are Argonne National Laboratory, the University of California at Santa Cruz, the University of Cambridge, Centro de Investigaciones Enérgeticas, Medioambientales y Tecnológicas–Madrid, the University of Chicago, University College London, the DES-Brazil Consortium, the University of Edinburgh, the Eidgenössische Technische Hochschule (ETH) Zürich, Fermi National Accelerator Laboratory, the University of Illinois at Urbana-Champaign, the Institut de Ciències de l'Espai (IEEC/CSIC), the Institut de Física d'Altes Energies, Lawrence Berkeley National Laboratory, the Ludwig-Maximilians Universität München and the associated Excellence Cluster Universe, the University of Michigan, the National Optical Astronomy Observatory, the University of Nottingham, The Ohio State University, the OzDES Membership Consortium, the University of Pennsylvania, the University of Portsmouth, SLAC National Accelerator Laboratory, Stanford University, the University of Sussex, and Texas A\&M University.

Based in part on observations at Cerro Tololo Inter-American Observatory, National Optical Astronomy Observatory, which is operated by the Association of Universities for Research in Astronomy (AURA) under a cooperative agreement with the National Science Foundation. 

This research used resources of the National Energy Research Scientific Computing Center (NERSC), a U.S. Department of Energy Office of Science User Facility located at Lawrence Berkeley National Laboratory, operated under Contract No. DE-AC02-05CH1123.

\new{This research was supported in part through the use of DARWIN computing system: DARWIN – A Resource for Computational and Data-intensive Research at the University of Delaware and in the Delaware Region, which is supported by NSF under Grant Number: 1919839, Rudolf Eigenmann, Benjamin E. Bagozzi, Arthi Jayaraman, William Totten, and Cathy H. Wu, University of Delaware, 2021.}

This material is based upon work supported by the University of Delaware Graduate
College through the Unidel Distinguished Graduate Scholar Award. Any opinions,
findings, and conclusions or recommendations expressed in this material are those of
the author(s).

TAC thanks the LSSTC Data Science Fellowship Program, which is funded by LSSTC, NSF Cybertraining Grant \#1829740, the Brinson Foundation, and the Moore Foundation; her participation in the program has benefited this work.

All of our code is written in \texttt{Python} using the following packages:
\begin{itemize}
\item \texttt{TensorFlow \citep{tensorflow2015-whitepaper}}
\item \texttt{Keras \citep{chollet2015keras}}
\item \texttt{numpy \citep{harris2020array}}
\item \texttt{astropy \citep{astropy:2013, astropy:2018} }
\item \texttt{pandas \citep{mckinney-proc-scipy-2010, reback2020pandas}}
\item \texttt{matplotlib \citep{Hunter:2007}}
\item \texttt{seaborn \citep{Waskom2021}}
\end{itemize}

\section{Data availability}
The data underlying this article are available in 
\url{https://portal.nersc.gov/project/dessn/autoscan/#tdata},
and explain in more detail in \cite{Goldstein_2015}.

%% file: appendixa.tex
\begin{figure*}[ht]
    \centering
    \includegraphics[width=0.6\linewidth]{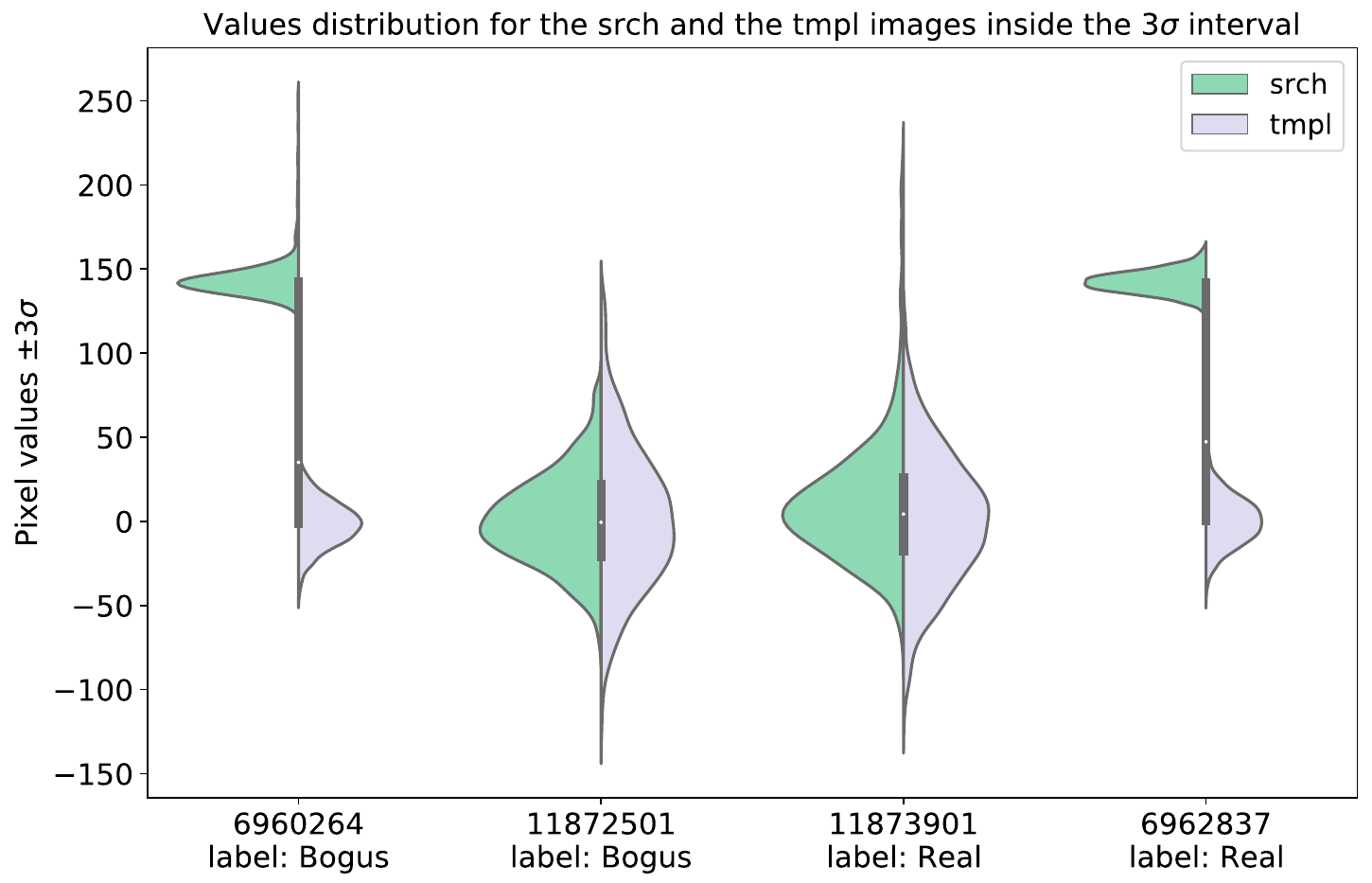}
    \caption{Violin plot of pixel values inside the $\mu \pm 3\sigma$ interval for \textit{srch} and \textit{temp} image of %Quantitative representation of 
    the data shown in \autoref{fig:examples_no_normalization} (before normalization). These values were scaled between $0$ and $1$. Left and right curve correspond to \textit{srch} and \textit{temp} distribution of values respectively. The distribution of values correspond to the same shown in \autoref{fig:histobeforenormalizarion} for \textit{srch} and \textit{temp}, the green and purple colors has its correspondence to \autoref{fig:histobeforenormalizarion}.}
    \label{fig:violinplot}
\end{figure*}
\begin{figure*}[ht]
    \centering
    \includegraphics[width=0.6\linewidth]{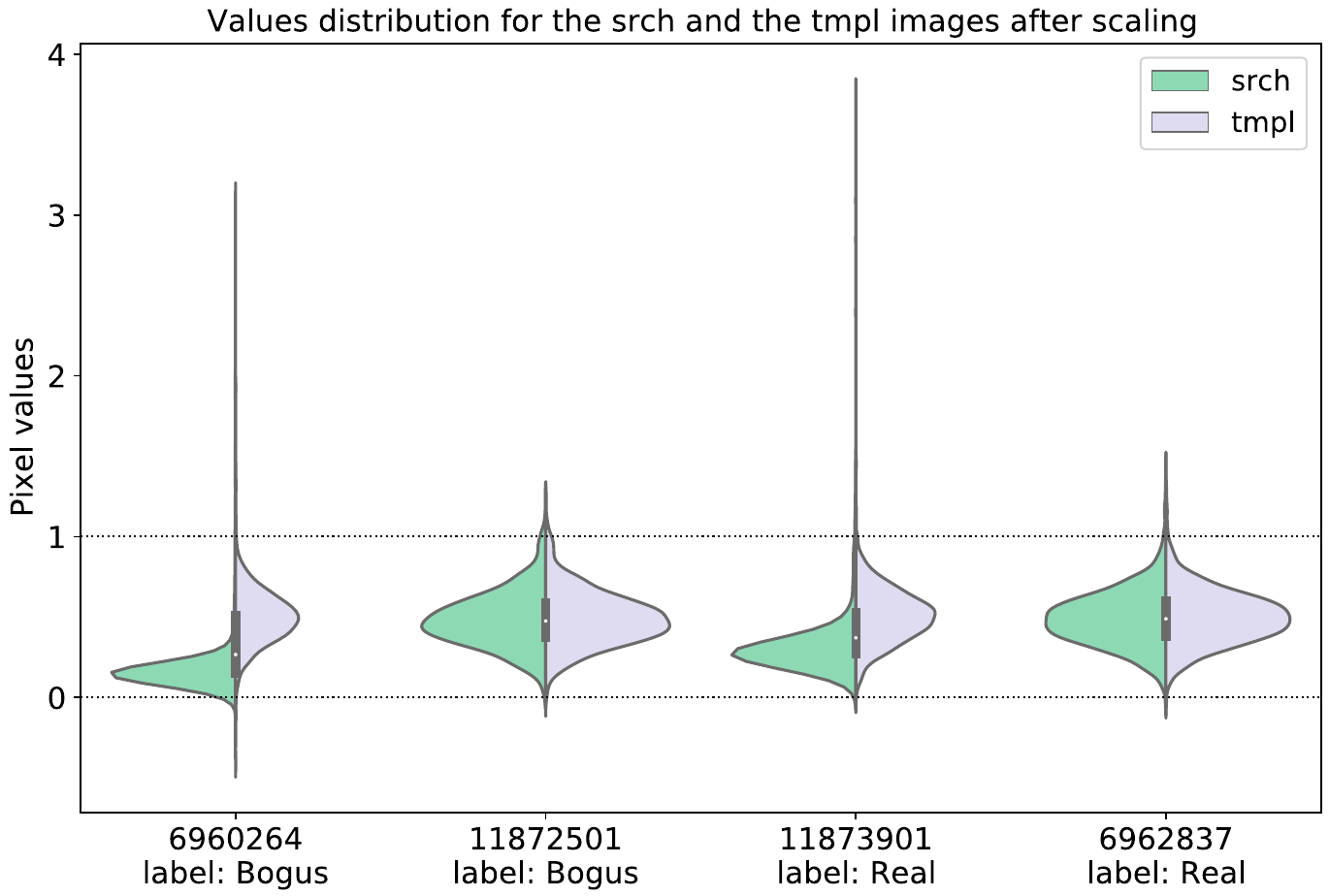}
    \caption{Violin plot of pixel values for \textit{srch} and \textit{temp} image of %Quantitative representation of 
    the data shown in \autoref{fig:examples_no_normalization} (after normalization). Values between the vertical lines in $0$ and $1$ were mapped to the values in \autoref{fig:violinplot}. Negative values and above $1$ correspond to the scaled values outside the $3\sigma$ clip. }
    \label{fig:violinplotscaling}
\end{figure*}

To visualize and compare the behaviour of the flux of the \search\ and \temp\ images, the value distribution for the four transients presented in \autoref{fig:examples_no_normalization} is shown as a violin plot in \autoref{fig:violinplot}. The first transient on the left is labeled as ``bogus'', the \search\ distribution pixel values are in general greater, positive and non-zero center than the values for \temp, the same behaviour is observed to the last transient on the right, but this one is labeled as ``real''. This same comparison can be applied to the transients plotted on the middle of \autoref{fig:violinplot}, both show similar distributions, however one is ``bogus'' and the other is ``real''. For the four transients presented the pixel values have long tails, outside the $\mu \pm 3\sigma$ values. The scaling of the \search\ and \temp\ images for the four transients, according to the description given in \autoref{subsec:scaling} is visualized in \autoref{fig:violinplotscaling}. 

%Due to the long tails presented in each image, the similarities observed in \autoref{fig:violinplot} are not longer visible after the scaling.

%% file: appendix_3channel.tex
\begin{figure*}[h]
    \centering
    \includegraphics[width=0.5\linewidth]{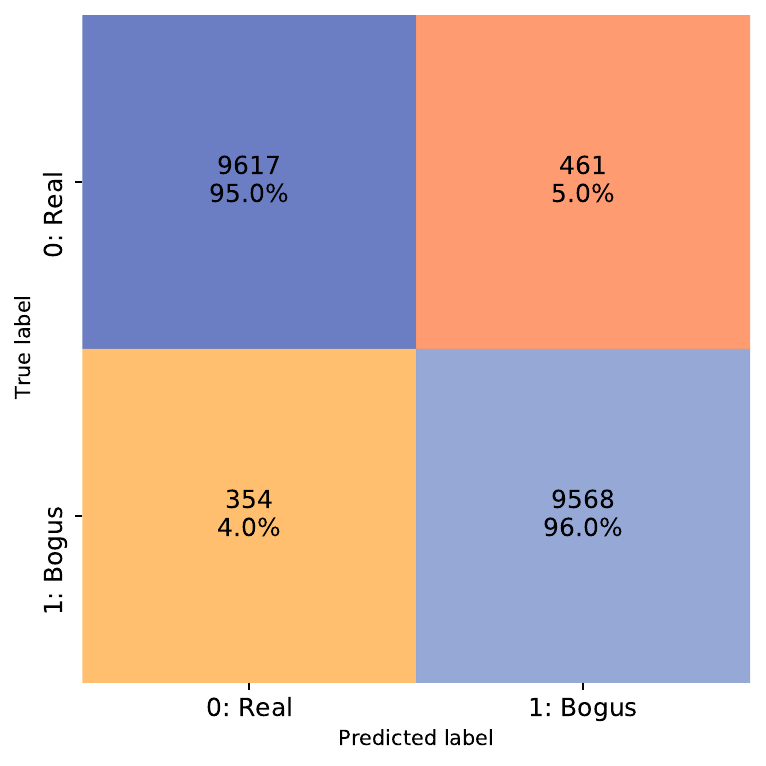}
    \caption{As {\it Left} \autoref{fig:confusiomatrix_models} but for a \diabased\ model which takes input data stacked in depth axis, \ie, 5$\times$151$\times$3. Overall, the performance of the model is not significantly affected by the choice of input data, with a $\sim1\%$ decrease in the TN rate.} 
    \label{fig:confusionechannel}
\end{figure*}

%% file: appendixb.tex
We designed a network of 12 layers using \texttt{tensorflow.keras} in python, for the \diabased\ case \autoref{tab:DIACNN_architecture} and a network of 11 layers using  \texttt{tensorflow.keras} in python, for the \nodia\ case \autoref{tab:noDIACNN_architecture}.
\newtwo{Here we show the details of the architectures of these models. \autoref{tab:compilation} shows the compilation hyperparameters, including optimizer, learning rate, loss function, etc, which are shared by all models.
In \autoref{tab:DIACNN_architecture} and \autoref{tab:noDIACNN_architecture} we show the number of neurons or filters in each layer, the size of the filters and padding choice in convolutional layers, and the activation functions. } 

\begin{table}[ht!]
    \centering
    \begin{tabular}{|c|c|c|c|c|}
    \hline
    
    \multicolumn{5}{|l|}{\bf{Compilation configuration}} \\\hline
    Optimizer&\multicolumn{4}{c|}{Stochastic Gradient Descent}\\\hline
    Learning rate&\multicolumn{4}{c|}{0.01}\\\hline
    Batch size&\multicolumn{4}{c|}{200}\\\hline
    Loss function&\multicolumn{4}{c|}{Sparse Categorical Crossentropy}\\\hline
    Metric&\multicolumn{4}{c|}{Accuracy}\\\hline
    Save best model&\multicolumn{4}{c|}{Max validation accuracy}\\\hline
    Stop training&\multicolumn{4}{c|}{After 100 epochs with no improvements in validation loss}\\
\hline    \end{tabular}
    \caption{Compilation hyperparamerters shared by all our models}
    \label{tab:compilation}
\end{table}

\begin{table}[ht!]
    \centering
    \begin{tabular}{|c|c|c|c|c|}
    \hline
    \bf{Type}&\bf{Filters}&\bf{Size}&\bf{Padding}&\bf{Activation}\\\hline
    Conv2D&16&(5,5)&valid&relu\\\hline
    MaxPooling&-&(2,2)&-&-\\\hline
    Dropout&-&0.4&-&-\\\hline
    Conv2D&32&(5,5)&valid&relu\\\hline
    MaxPooling&-&(2,2)&-&-\\\hline
    Dropout&-&0.4&-&-\\\hline
    Conv2D&64&(5,5)&valid&relu\\\hline
    MaxPooling&-&(2,2)&-&-\\\hline
    Dropout&-&0.4&-&-\\\hline
    Flatten&-&-&-&-\\\hline
    Dense&32&-&-&relu\\\hline
    Dense&2&-&-&softmax\\\hline
    \end{tabular}
    \caption{\diabased\  model architecture.}
    \label{tab:DIACNN_architecture}
\end{table}

\begin{table}
    \centering
    \begin{tabular}{|c|c|c|c|c|}
    \hline
    \bf{Type}&\bf{Filters}&\bf{Size}&\bf{Padding}&\bf{Activation}\\\hline
    Conv2D&1&(7,7)&valid&relu\\\hline
    MaxPooling&-&(2,2)&-&-\\\hline
    Conv2D&16&(3,3)&valid&relu\\\hline
    MaxPooling&-&(2,2)&-&-\\\hline
    Dropout&-&0.4&-&-\\\hline
    Conv2D&32&(3,3)&valid&relu\\\hline
    MaxPooling&-&(2,2)&-&-\\\hline
    Dropout&-&0.4&-&-\\\hline
    Flatten&-&-&-&-\\\hline
    Dense&32&-&-&relu\\\hline
    Dense&2&-&-&softmax\\\hline
    \end{tabular}
    \caption{\nodia\ model architecture.}
    \label{tab:noDIACNN_architecture}
\end{table}

\newpage
\subsection{\round{Architecture other models}}

\newtwo{In addition to the CNN \diabased\ and \nodia\ models presented in this paper, five alternative architectures were tested (see \autoref{sec:method}).  The tables in this appendix show the details of the architectures of these models. The structure of the table is the same as for \autoref{tab:DIACNN_architecture} and \autoref{tab:noDIACNN_architecture}.
\autoref{tab:DIACNNresnet_architecture} shows the architecture for ResNet 50-V2 \citep{RESNET50V2} and VGG16 \citep{VGG16}. Only the final non-convolutional layers are shown, as the convolutional elements are maintained as designed in the respective papers. \autoref{tab:noDIACNNextradenseend_architecture} through \autoref{tab:noDIACNNextracndense_architecture} describe modifications of the final \diabased\ and \nodia\ architecture described in \autoref{tab:DIACNN_architecture} and \autoref{tab:noDIACNN_architecture}; they include additional dense layers to assess the possible impact of bottlenecks in information caused by large jumps in the number of neurons between consecutive layers (\autoref{tab:noDIACNNextradenseend_architecture} through \autoref{tab:noDIACNNextradense_architecture}) and  deeper models with additional convolutional and  dense layers (\autoref{tab:noDIACNNextracn_architecture} and \autoref{tab:noDIACNNextracndense_architecture}), but none of these layers led to improvements in the model performance, thus the simpler versions were chosen as our final models. }

\begin{table}[!h]
    \centering
    \begin{tabular}{|c|c|c|c|c|}
    \hline
    \bf{Type}&\bf{Filters}&\bf{Size}&\bf{Padding}&\bf{Activation}\\\hline
    \multicolumn{5}{|c|}{specific convolutional architecture} \\\hline
    Dropout&-&0.4&-&-\\\hline
    Flatten&-&-&-&-\\\hline
    Dense&32&-&-&relu\\\hline
    Dense&2&-&-&softmax\\\hline
    \end{tabular}
    \caption{ResNet 50-V2 \citep{RESNET50V2} and VGG16 \citep{VGG16} architectures tested on the DIA data. Only the feed-forward layers are shown. For the specific concolutional architecture details please refer to the original papers.}
    \label{tab:DIACNNresnet_architecture}
\end{table}

\begin{table}[!h]
    \centering
    \begin{tabular}{|c|c|c|c|c|}
    \hline
    \bf{Type}&\bf{Filters}&\bf{Size}&\bf{Padding}&\bf{Activation}\\\hline
    Conv2D&1&(7,7)&valid&relu\\\hline
    MaxPooling&-&(2,2)&-&-\\\hline
    Conv2D&16&(3,3)&valid&relu\\\hline
    MaxPooling&-&(2,2)&-&-\\\hline
    Dropout&-&0.4&-&-\\\hline
    Conv2D&32&(3,3)&valid&relu\\\hline
    MaxPooling&-&(2,2)&-&-\\\hline
    Dropout&-&0.4&-&-\\\hline
    Flatten&-&-&-&-\\\hline
    Dense&32&-&-&relu\\\hline
    Dense&24&-&-&relu\\\hline
    Dense&16&-&-&relu\\\hline
    Dense&8&-&-&relu\\\hline
    Dense&2&-&-&softmax\\\hline
    \end{tabular}
    \caption{\nodia\ model with additional Dense Layers (size = [24, 16, 8]) after Dense Layer size 32.}
    \label{tab:noDIACNNextradenseend_architecture}
\end{table}

\begin{table}[!h]
    \centering
    \begin{tabular}{|c|c|c|c|c|}
    \hline
    \bf{Type}&\bf{Filters}&\bf{Size}&\bf{Padding}&\bf{Activation}\\\hline
    Conv2D&1&(7,7)&valid&relu\\\hline
    MaxPooling&-&(2,2)&-&-\\\hline
    Conv2D&16&(3,3)&valid&relu\\\hline
    MaxPooling&-&(2,2)&-&-\\\hline
    Dropout&-&0.4&-&-\\\hline
    Conv2D&32&(3,3)&valid&relu\\\hline
    MaxPooling&-&(2,2)&-&-\\\hline
    Dropout&-&0.4&-&-\\\hline
    Flatten&-&-&-&-\\\hline
    Dense&1200&-&-&relu\\\hline
    Dense&800&-&-&relu\\\hline
    Dense&400&-&-&relu\\\hline
    Dense&160&-&-&relu\\\hline
    Dense&80&-&-&relu\\\hline
    Dense&32&-&-&relu\\\hline
    Dense&24&-&-&relu\\\hline
    Dense&16&-&-&relu\\\hline
    Dense&8&-&-&relu\\\hline
    Dense&2&-&-&softmax\\\hline
    \end{tabular}
    \caption{\nodia\ model with additional Dense Layers (size = [1200, 800, 400, 160, 80]) after Flattening Layer and additional after Dense Layer size 32.}
    \label{tab:noDIACNNextradense_architecture}
\end{table}

\begin{table}[!h]
    \centering
    \begin{tabular}{|c|c|c|c|c|}
    \hline
    \bf{Type}&\bf{Filters}&\bf{Size}&\bf{Padding}&\bf{Activation}\\\hline
    Conv2D&1&(7,7)&valid&relu\\\hline
    Conv2D&8&(5,5)&valid&relu\\\hline
    Conv2D&16&(3,3)&valid&relu\\\hline
    Conv2D&24&(3,3)&valid&relu\\\hline
    Dropout&-&0.4&-&-\\\hline
    Conv2D&32&(3,3)&valid&relu\\\hline
    MaxPooling&-&(2,2)&-&-\\\hline
    Dropout&-&0.4&-&-\\\hline
    Flatten&-&-&-&-\\\hline
    Dense&32&-&-&relu\\\hline
    Dense&2&-&-&softmax\\\hline
    \end{tabular}
    \caption{\nodia\ model with additional Convolutional Layers.}
    \label{tab:noDIACNNextracn_architecture}
\end{table}

\begin{table}[!h]
    \centering
    \begin{tabular}{|c|c|c|c|c|}
    \hline
    \bf{Type}&\bf{Filters}&\bf{Size}&\bf{Padding}&\bf{Activation}\\\hline
    Conv2D&1&(7,7)&valid&relu\\\hline
    Conv2D&8&(5,5)&valid&relu\\\hline
    Conv2D&16&(3,3)&valid&relu\\\hline
    Conv2D&24&(3,3)&valid&relu\\\hline
    Dropout&-&0.4&-&-\\\hline
    Conv2D&32&(3,3)&valid&relu\\\hline
    MaxPooling&-&(2,2)&-&-\\\hline
    Dropout&-&0.4&-&-\\\hline
    Flatten&-&-&-&-\\\hline
    Dense&32&-&-&relu\\\hline
    Dense&24&-&-&relu\\\hline
    Dense&16&-&-&relu\\\hline
    Dense&8&-&-&relu\\\hline
    Dense&2&-&-&softmax\\\hline
    \end{tabular}
    \caption{\nodia\ model with additional Convolutional Layer and Dense Layers (size = [24, 16, 8]) after Dense Layer size 32.}
    \label{tab:noDIACNNextracndense_architecture}
\end{table}

\clearpage
\newpage
\subsection{\round{Hyperparameter Grid Search}}

\newtwo{The combination of hyperparameters tested for the \nodia\ model (see \autoref{sec:method}) and the respective testing accuracy. The hyperparameter grid search was implemented using \texttt{sklearn.model\_selection.RandomizedSearchCV}}

\begin{table}[!h]
    \centering
    \begin{tabular}{|c|c|c|c|c|}
    \hline
    \bf{Kernel layer 1}&\bf{Kernel layer 3}&\bf{Kernel layer 6}&\bf{Batch Size}&\bf{Test acc}\\\hline
    5&3&3&100&0.8980\\\hline
    5&3&3&200&0.8892\\\hline
    7&3&3&100&0.8968\\\hline
    % 7&3&3&200&0.91\\\hline
    7&5&3&100&0.9000\\\hline
    7&5&3&200&0.8958\\\hline
    7&5&5&100&0.9032\\\hline
    7&5&5&200&0.9004\\\hline
    10&3&3&100&0.8960\\\hline
    10&3&3&200&0.8853\\\hline
    10&5&3&100&0.9054\\\hline
    10&5&3&200&0.9008\\\hline
    10&5&5&100&0.9038\\\hline
    10&5&5&200&0.9012\\\hline
    15&3&3&100&0.8953\\\hline
    15&3&3&200&0.8784\\\hline
    15&5&3&100&0.8848\\\hline
    15&5&3&200&0.9023\\\hline
    15&5&5&100&0.8937\\\hline
    15&5&5&200&0.8854\\\hline
    \end{tabular}
    \caption{Grid search for \nodia\ model, varying kernel size of the Convolutional Layer 1, 3 and 5  (see \autoref{tab:noDIACNN_architecture}) and batchsize.}
    \label{tab:noDIAgridsearch}
\end{table}

% \begin{minted}{python}
% layer1 = keras.layers.Conv2D(1, kernel_size=(7, 7), padding="valid", activation="relu", 
%                              input_shape=(51,102,1))
% layer2 = keras.layers.MaxPooling2D((2, 2), strides=2)
% layer3 = keras.layers.Conv2D(16, kernel_size=(3, 3), padding="valid", activation="relu")
% layer4 = keras.layers.MaxPooling2D((2, 2), strides=2)
% layer5 = keras.layers.Dropout(0.4)
% layer6 = keras.layers.Conv2D(32, kernel_size=(3, 3), padding="valid", activation="relu")
% layer7 = keras.layers.MaxPooling2D((2, 2), strides=2)
% layer8 = keras.layers.Dropout(0.4)
% layer9 = keras.layers.Flatten()
% layer10 = keras.layers.Dense(32, activation="relu")
% layer11 = keras.layers.Dense(2, activation="softmax")
% \end{minted}
% To compile the models we used the optimizer \mintinline{c}/SGD/ (stochastic gradient descent) with a \mintinline{c}/learning_rate=0.01/ the \mintinline{c}/"sparse_categorical_crossentropy"/ loss function and the \mintinline{c}/"accuracy"/ metric. 
Our models are available in a dedicated GitHub repository \footnote{\url{https://github.com/taceroc/DIA_noDIA}.}

%% file: appendixc.tex
\begin{figure*}
    \centering
    \includegraphics[width=0.85\linewidth]{
    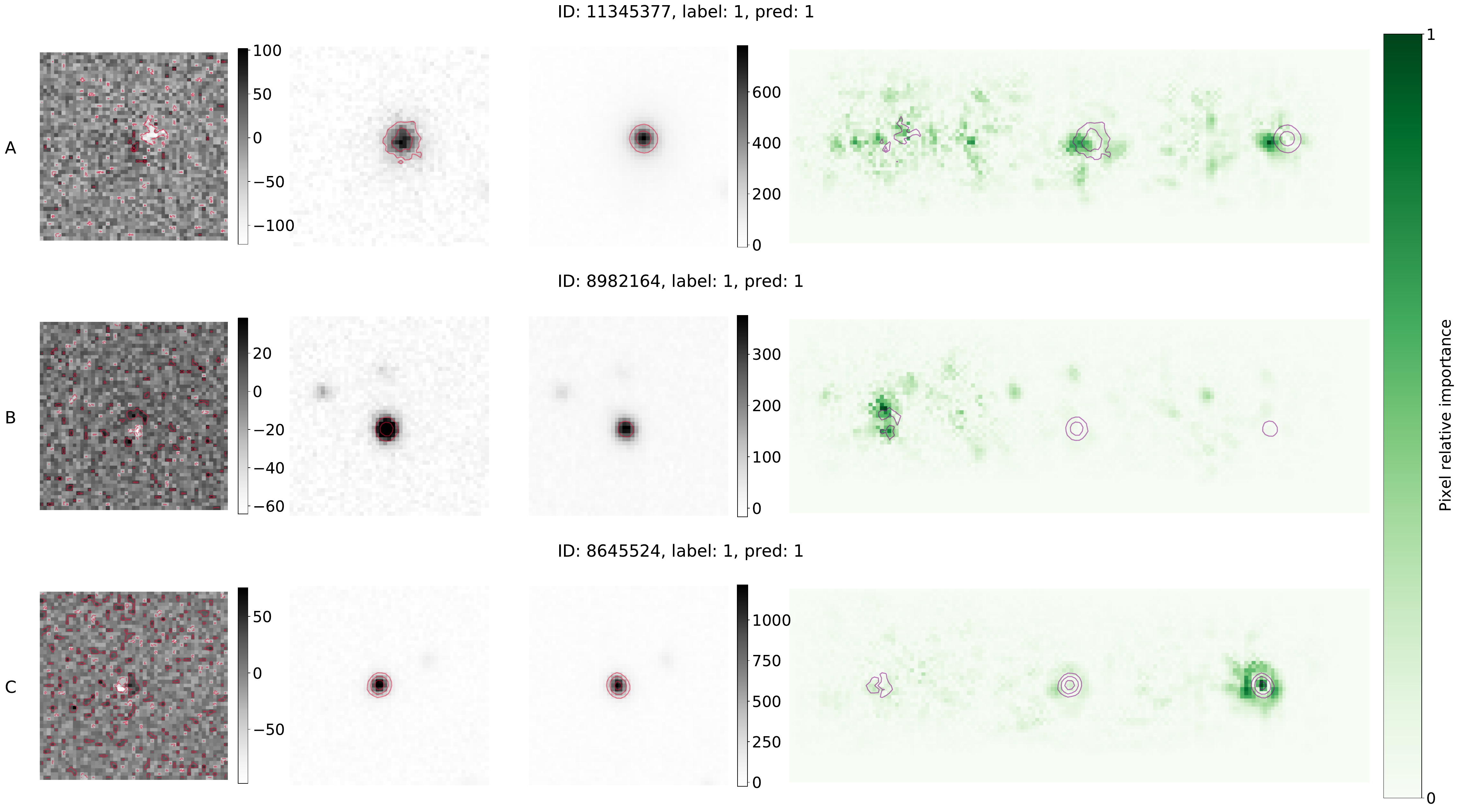}
    \caption{Transients (\diff-\search-\temp) and their respective saliency map for \diabased\ model True Negative predictions (correctly identified ``bogus'').  Contour plot of light intensity from the original images are overplotted, delineating the bright sources. In panel A: $I_\diff \sim 0.4$; B: $I_\diff\sim0.69$ and C $I_\temp\sim0.42$. In panel C, the saliency for the \temp\ portion of the image looks strikingly similar to the selection of pixels that is made in aperture photometry, with pixel values considered in the core of the source, ignored in a region immediately around the source, and again considered farther out to calculate the source's background. This figure is further discussed in \hyperref[sec:appendixc]{Appendix C}.}
    \label{fig:saliency_3id_contour}

    \includegraphics[width=0.65\linewidth]{
    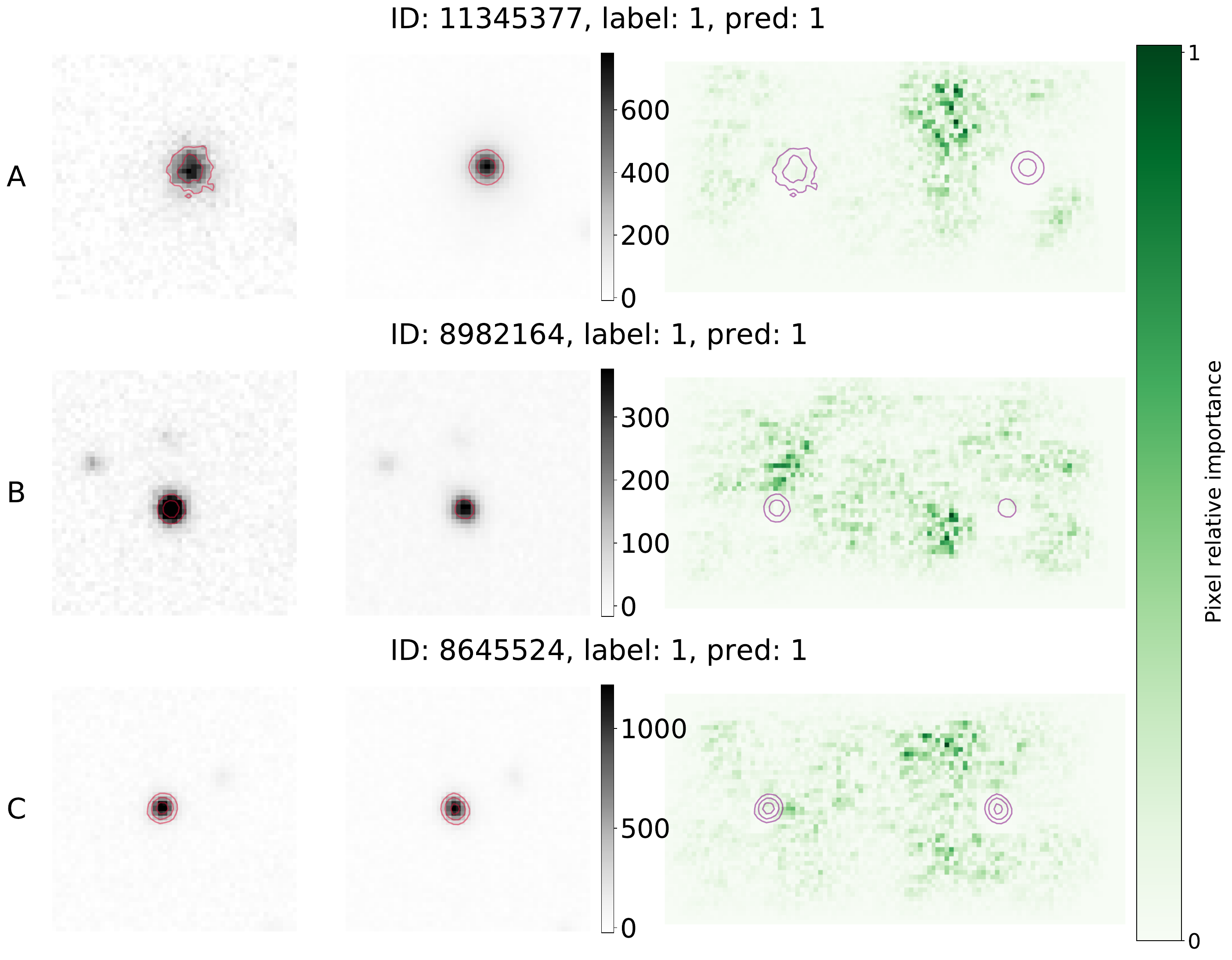}
    \caption{The same transients (\search-\temp) shown in \autoref{fig:saliency_3id_contour} and their respective saliency maps for \nodia\ model True Negatives (correctly identified ``bogus'').  Important pixels are found at nearly all locations in the image, rather than in a small region around the center. The model needs to learn properties of the image at large to enable a comparison of the \temp\ and \diff.  This figure is further discussed in \hyperref[sec:appendixc]{Appendix C}.}
    \label{fig:saliency_nodia3id_contour}
\end{figure*}
\begin{figure*}
    \centering
    \includegraphics[width=0.8\linewidth]{
    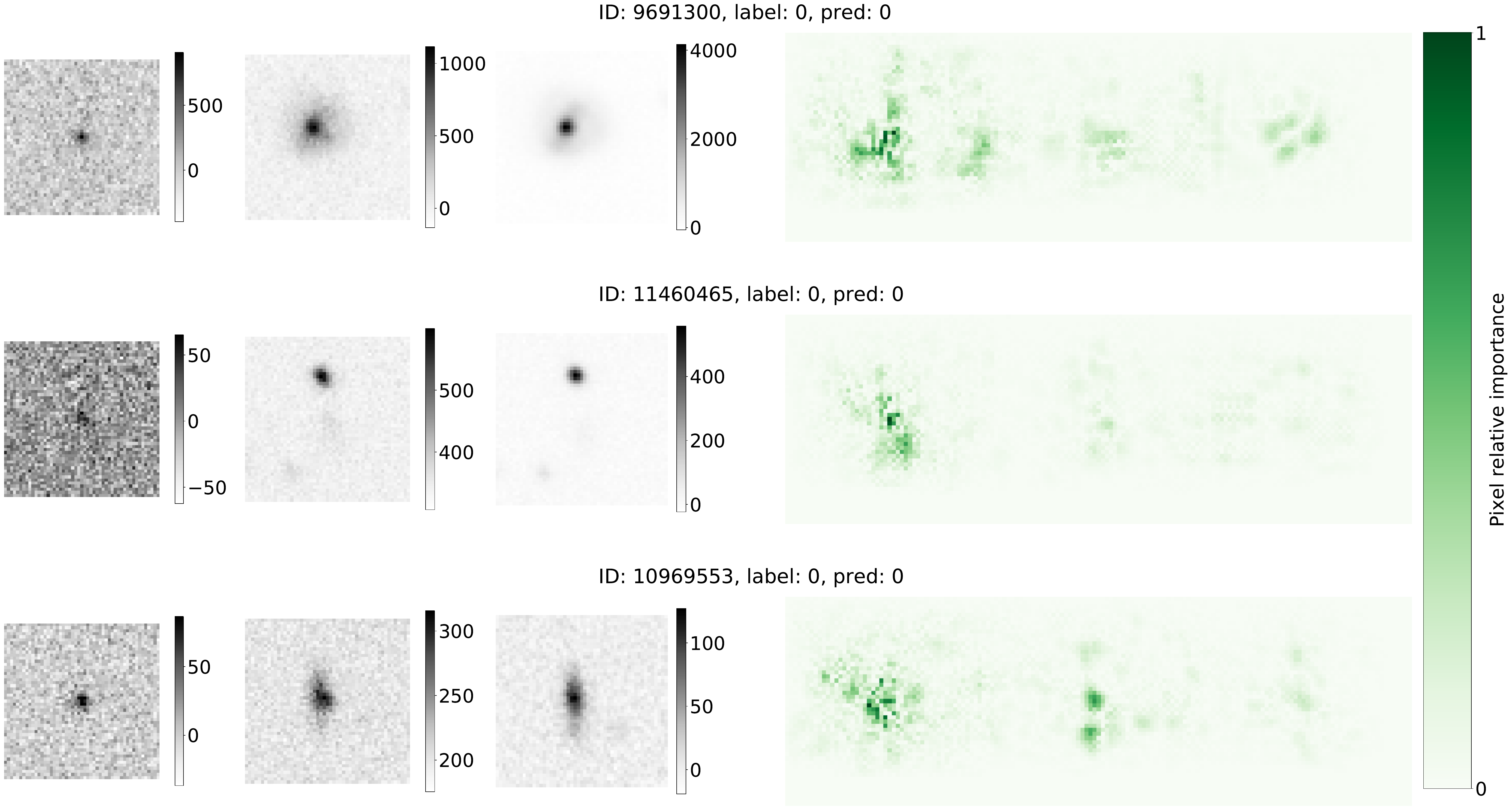}
    
    \includegraphics[width=0.8\linewidth]{
    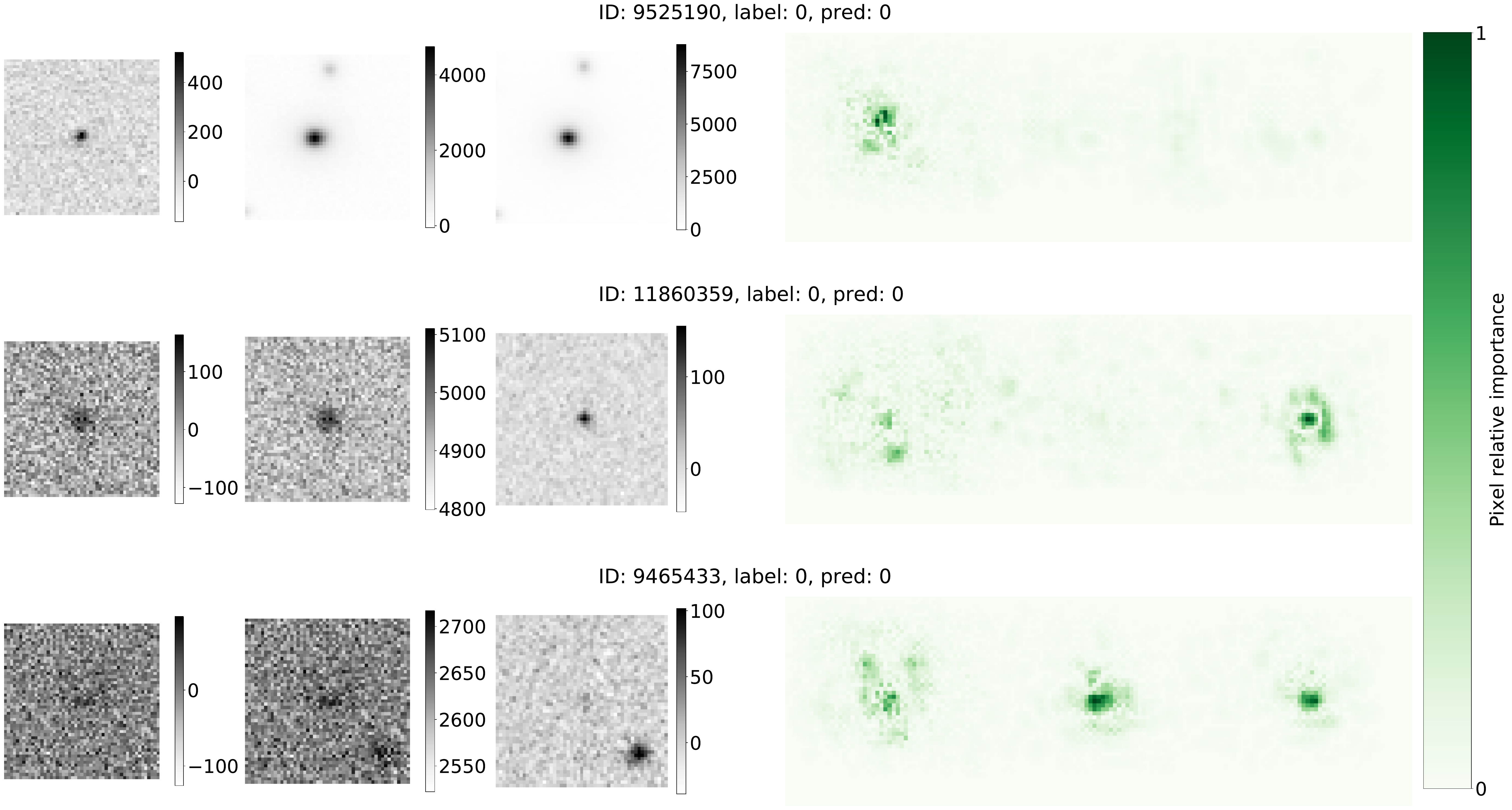}
    \caption{Transients (\diff-\search-\temp) and their respective saliency map  for \diabased\ model True Positives  (correctly identified ``real'' astrophysical transients). The important pixels are generally found in the \diff\ portion of the image for the \diabased\ model, as discussed in \autoref{sec:results}, but there are exceptions: here we show several cases of True Positives (real transients) classifications where the component of the image that was principally leveraged by the model was the \diff\ (the leftmost third) but two cases where the \nodia\ relied principally on the \search\ and/or \temp\ (bottom two panels).}
    \label{fig:saliency_dia3id_TP}
\end{figure*}
\begin{figure*}
    \centering
    \includegraphics[width=0.8\linewidth]{
    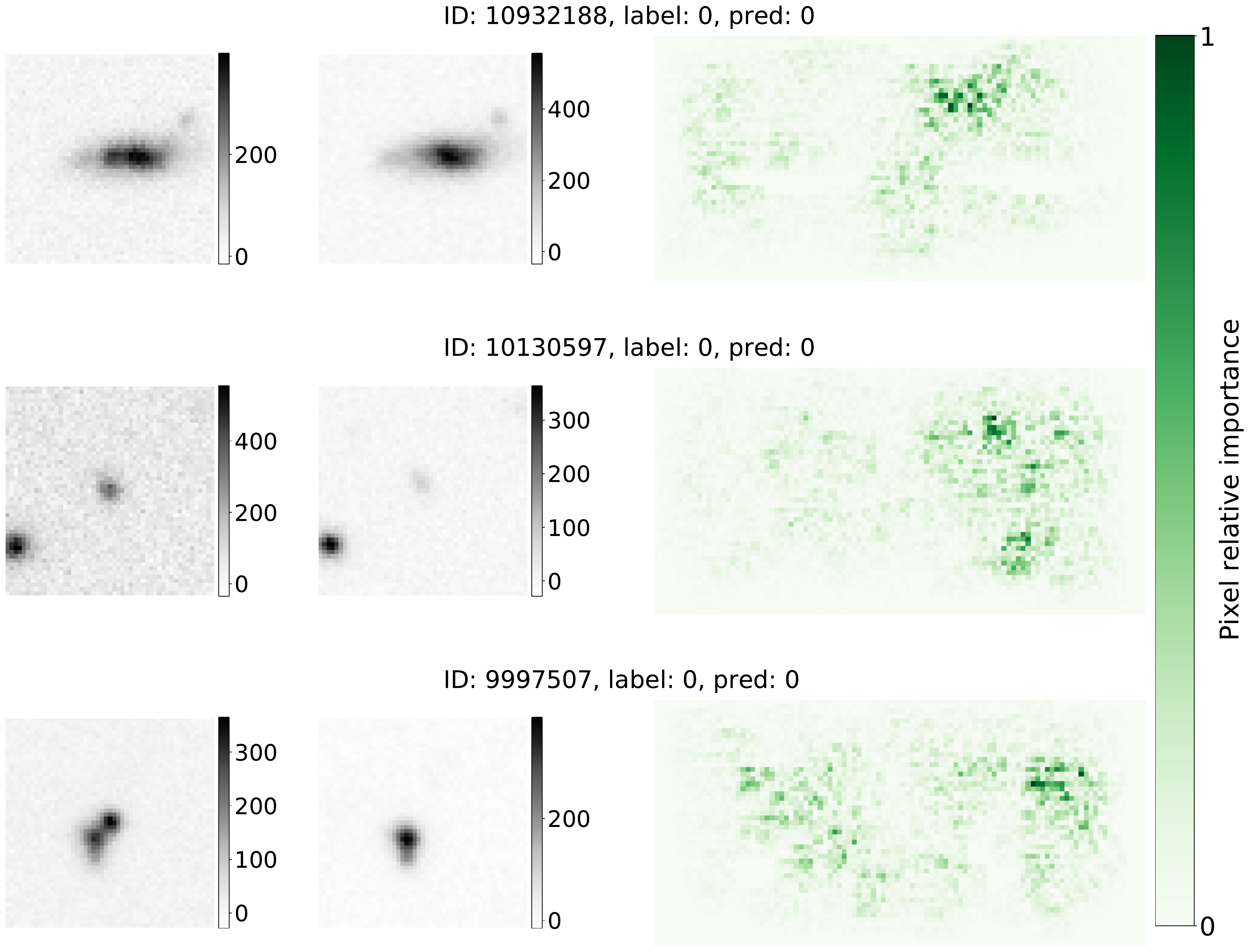}
    \includegraphics[width=0.8\linewidth]{
    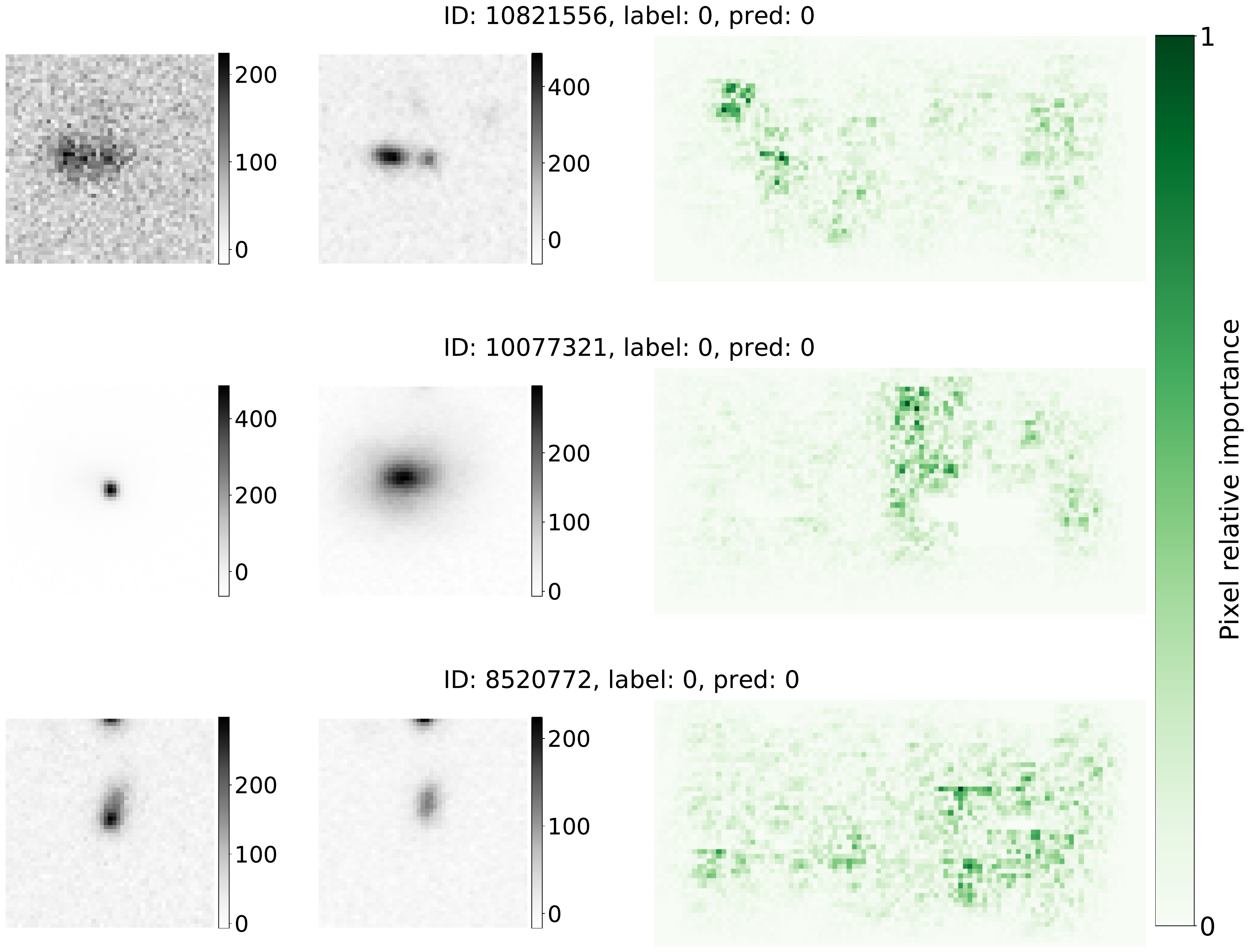}
   \caption{Transients (\search-\temp) and their respective saliency map for the \nodia\ model True Positives  (correctly identified ``real'' astrophysical transients). Important pixels are found everywhere in the image, as the CNN learns how to compare the \diff\ and \temp\ taking a synoptic look at the properties of each image component.}
    \label{fig:tpndia}
\end{figure*}
\begin{figure*}
    \centering
    \includegraphics[width=0.8\linewidth]{
    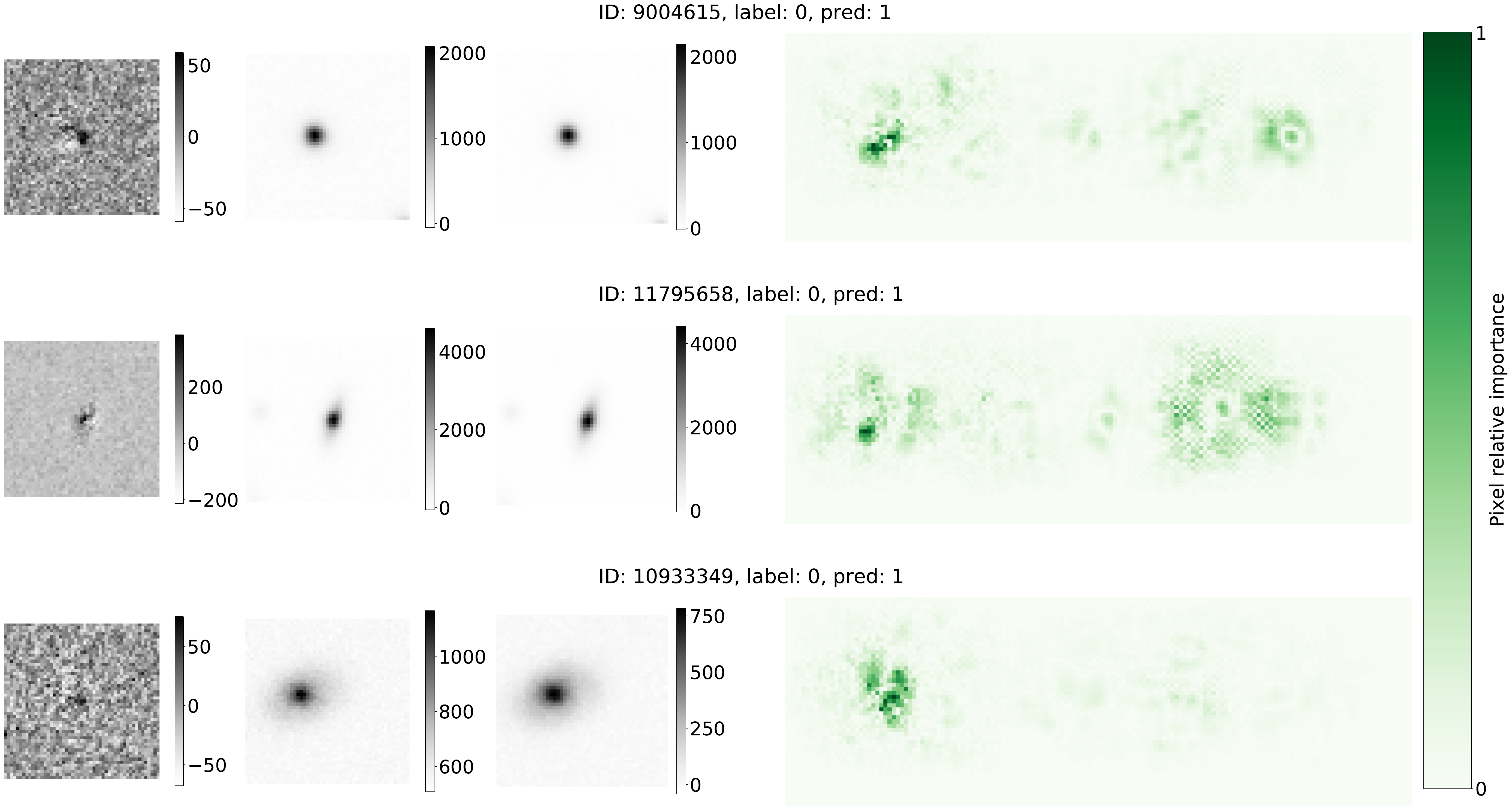}
    \includegraphics[width=0.8\linewidth]{
    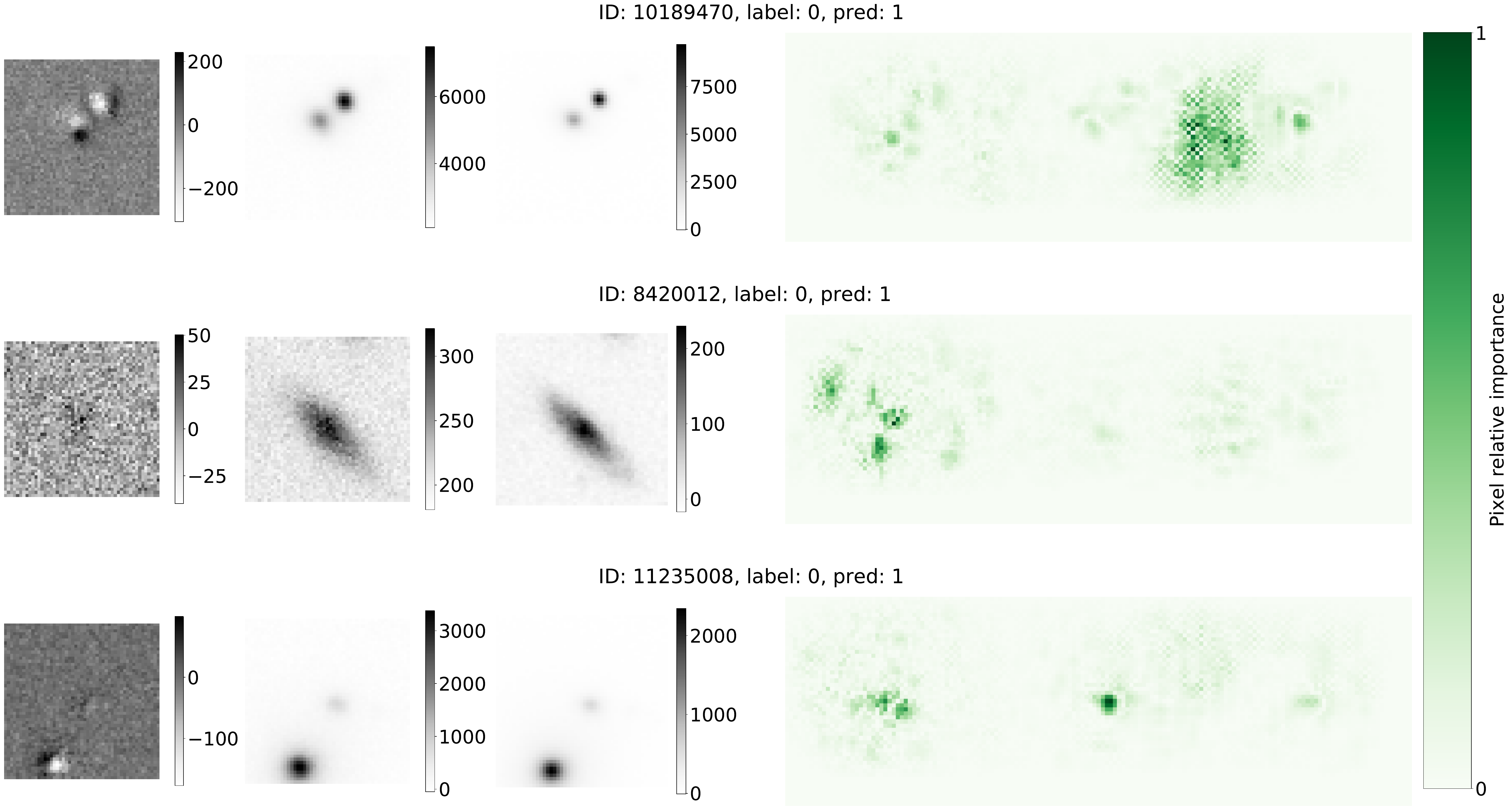}
    \caption{Transients (\diff-\search-\temp)  and their respective saliency map for \diabased\ model False Negatives (real transients identified as ``bogus'').
    We remind the reader that the labels are inherited from  \citet{Goldstein_2015} and cannot be verified. Some level of label inaccuracy is expected. The ``real'' transients in this dataset are implanted supernovae onto real DES images. However, in this collection, several transient display DIA inaccuracies (row 1, 2, 4, 6 show ``dipoles'', see \autoref{sec:DI}) that likely lead to the incorrect classification. Two very low signal-to-noise detections are missed (row 3 and 5) by our model.
    Important pixels are more commonly found in \diff\ portion of the image. In the \search\ saliency maps we see again that the core of the central source is used in the classification, as well as pixels that surround the source, but these two sets of important pixels are separate by it by a gap, again reminiscent of the typical aperture photometry technique (top two panels). } 
    \label{fig:saliency_dia3id_FN}
\end{figure*}
\begin{figure*}
    \centering
    \includegraphics[width=0.76\linewidth]{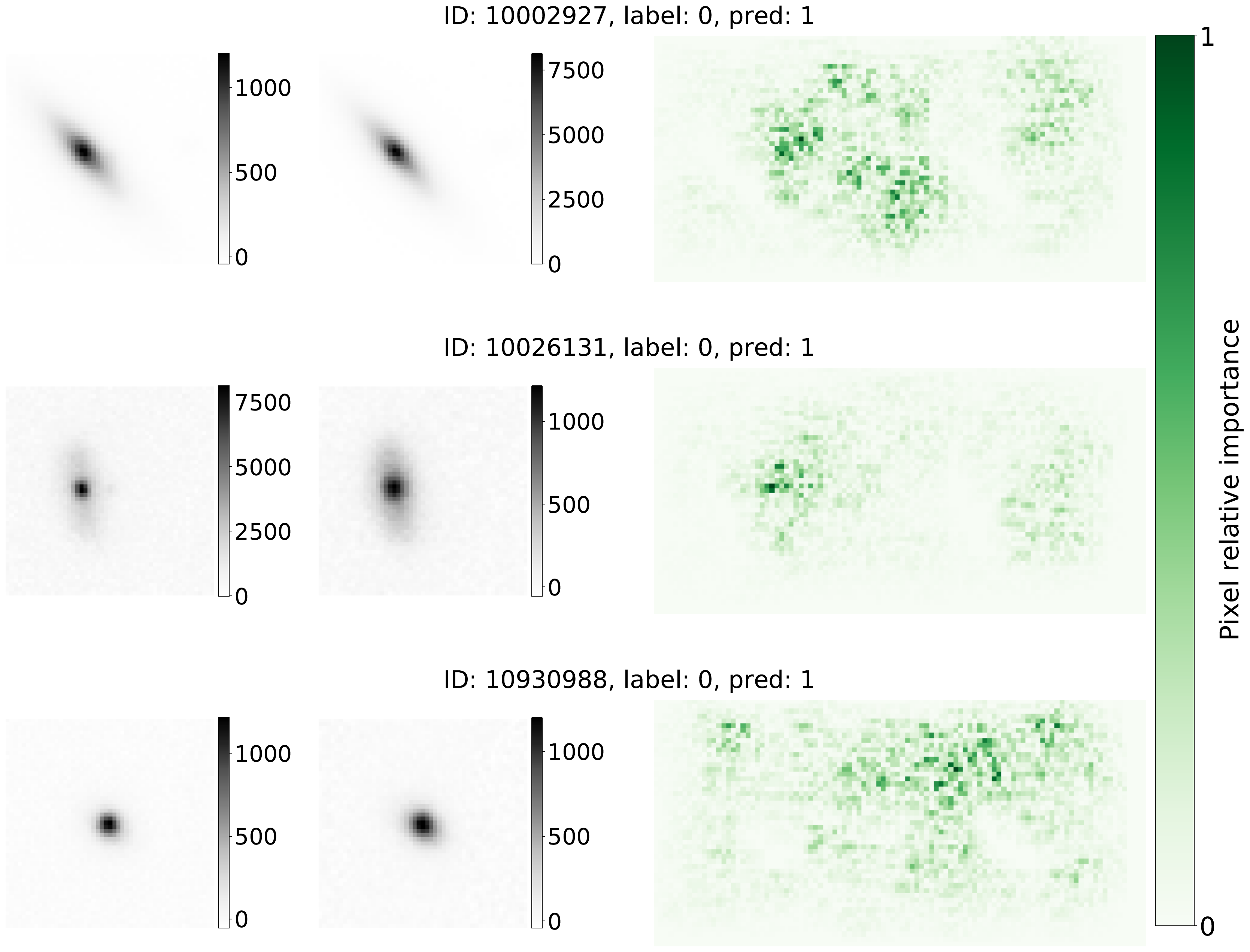}
    \includegraphics[width=0.76\linewidth]{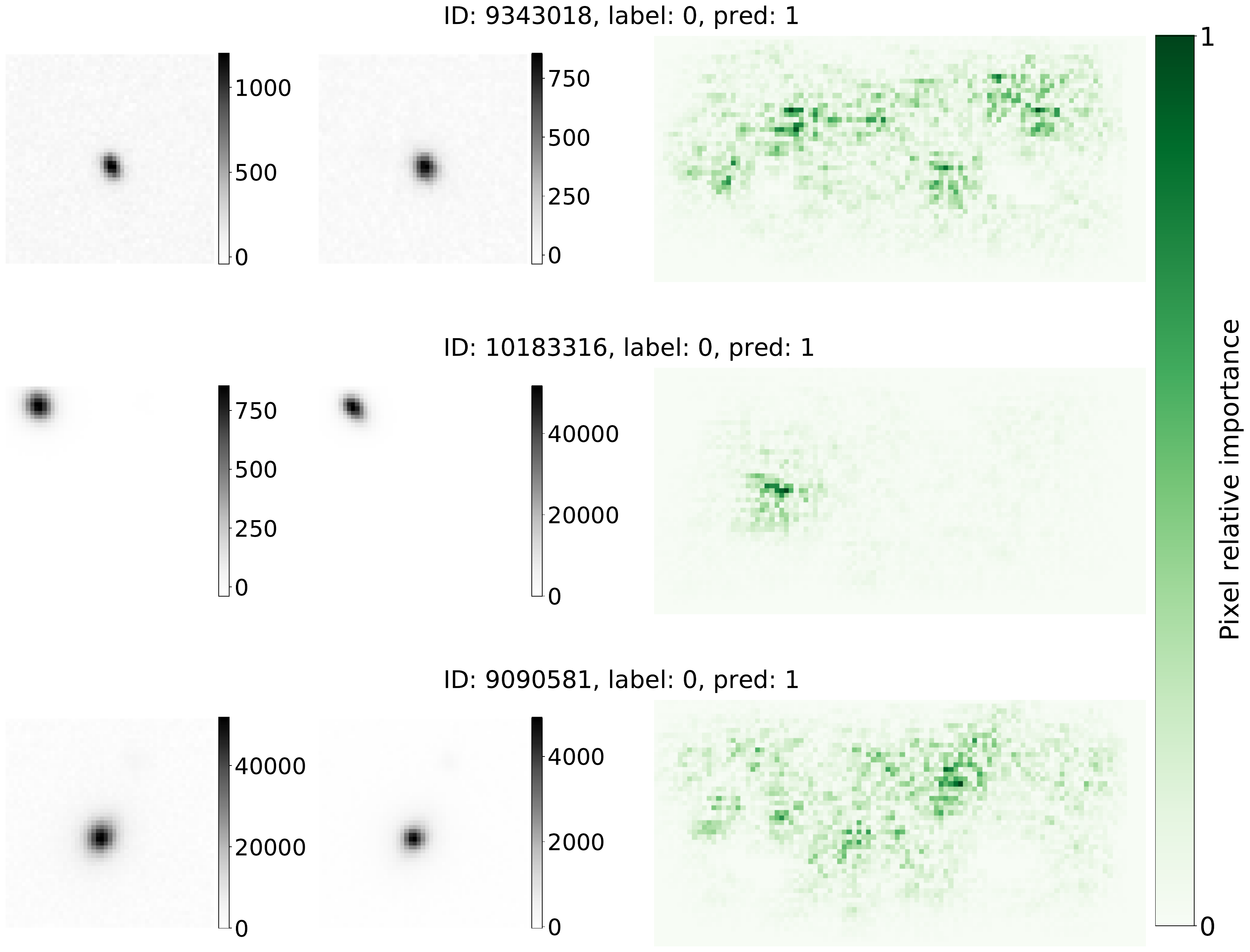}
    \caption{
    Transients (\search-\temp) and their respective saliency map for the \nodia\ model False Negatives (astrophysical sources classified as ``bogus''). In all cases but row 5 it is not clear why the classification fails. In row 5, another source dominates the image scaling (and pre-processing) reducing the visibility of the transient (that is completely missed by human inspection). We remind the reader that the ``real'' transients in this dataset are implanted supernovae onto real DES images. Important pixels are found everywhere in the image, as the CNN learns how to compare the \diff\ and \temp\ taking a synoptic look at the properties of each image component.}
    \label{fig:fnndia}
\end{figure*}
\begin{figure*}
    \centering
    \includegraphics[width=0.8\linewidth]{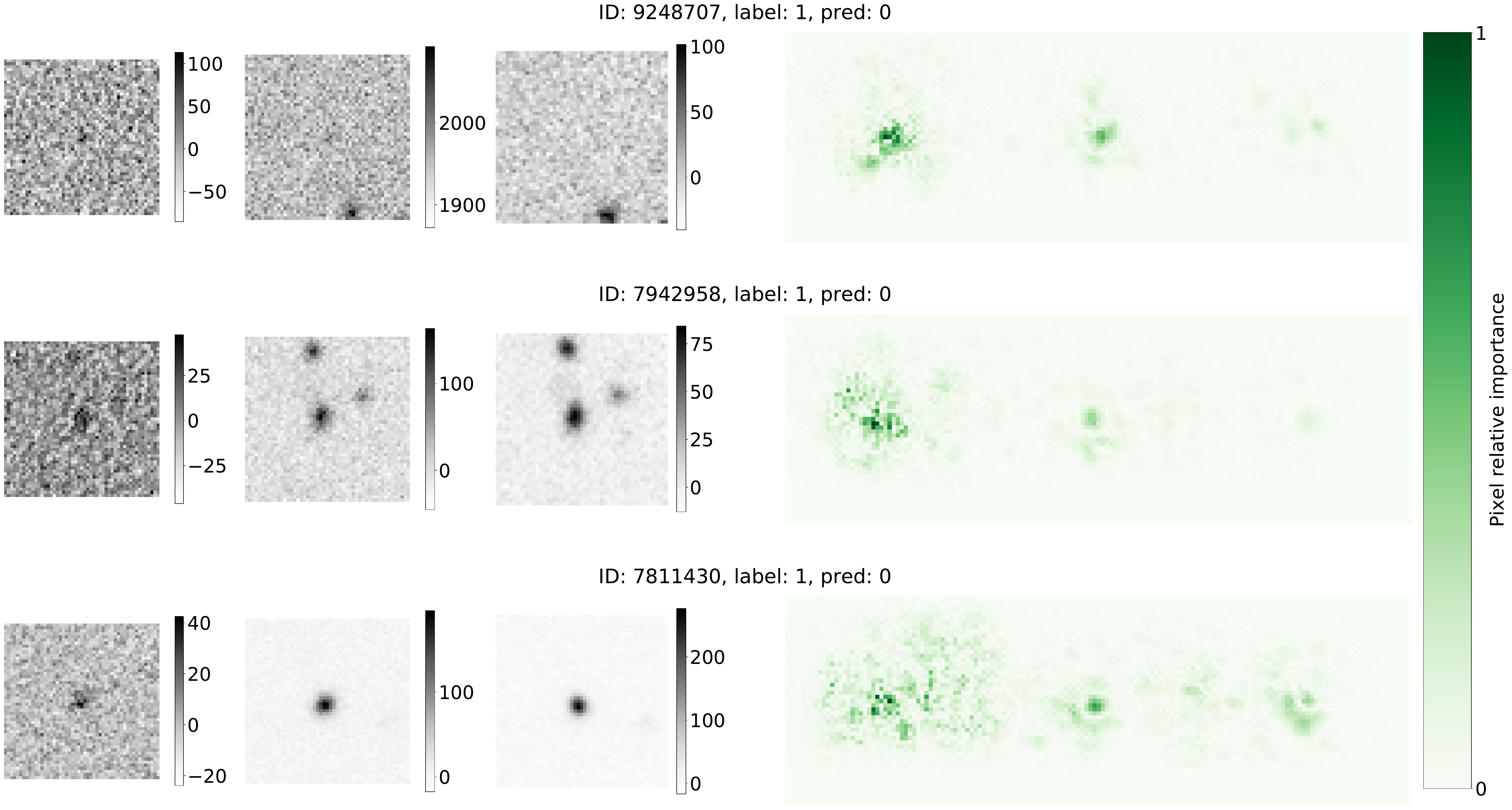}
    \includegraphics[width=0.8\linewidth]{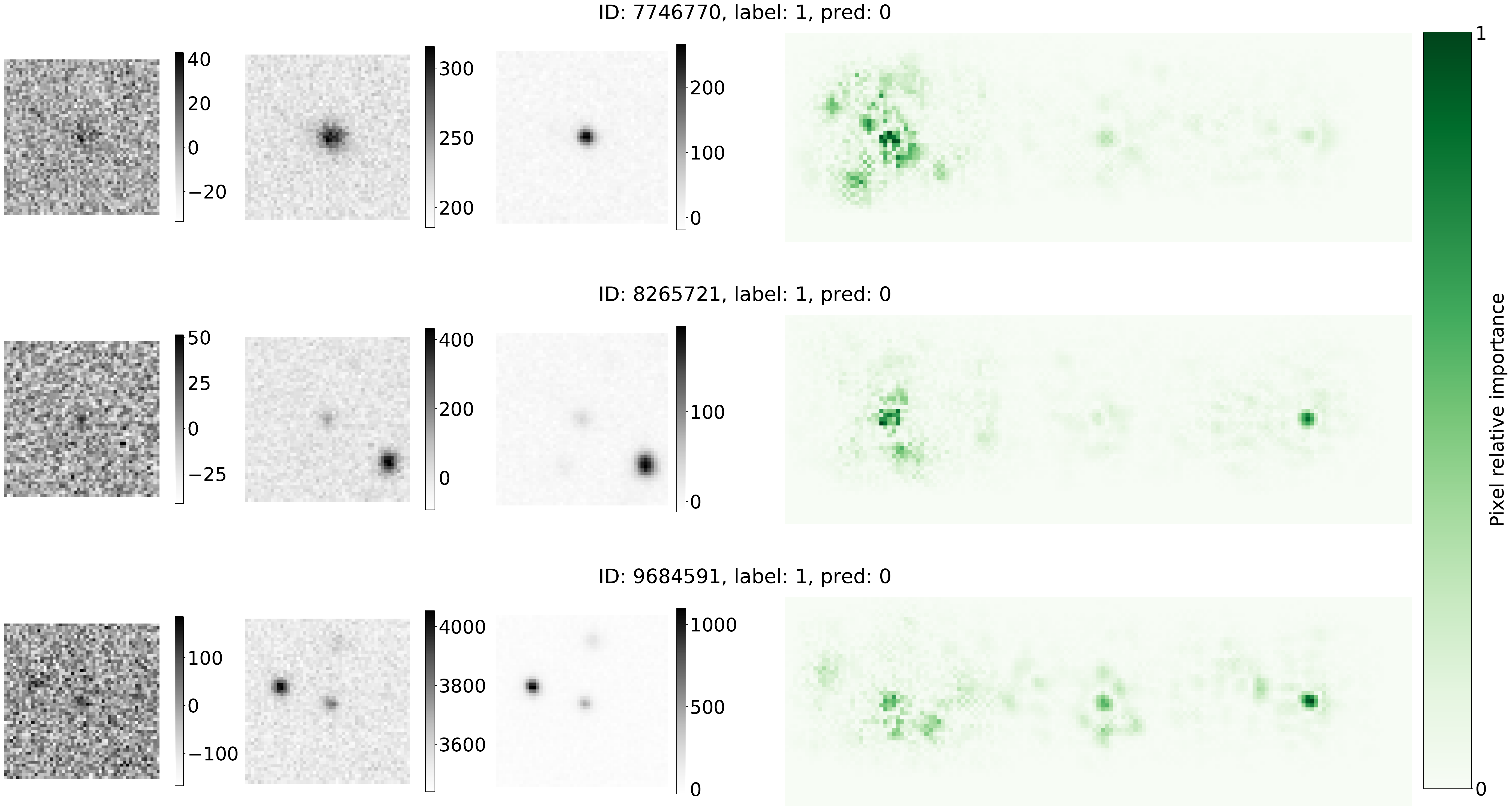}
    \caption{Transients (\diff-\search-\temp) and their respective saliency map for \diabased\ model False Positives (``bogus'' predicted as ``real'').  We remind the reader again that the labels are inherited from  \citet{Goldstein_2015} and cannot be verified.  Some level of label inaccuracy is expected. Bogus transients were labeled by human scanners among astrophysical images with detection. However, in this collection, we cannot verify the nature of the transient and we argue that in the cases presented here there is no obvious evidence of its ``bogus'' nature. Important pixels are most commonly found in the \diff, but in \temp\ and \search\ we see again the CNN analyzes the central source and its surrounding, but avoiding the tail of the central source, in a way similar to traditional aperture photometry techniques.}
    \label{fig:saliency_dia3id_FP}
\end{figure*}
\begin{figure*}
    \centering
    \includegraphics[width=0.76\linewidth]{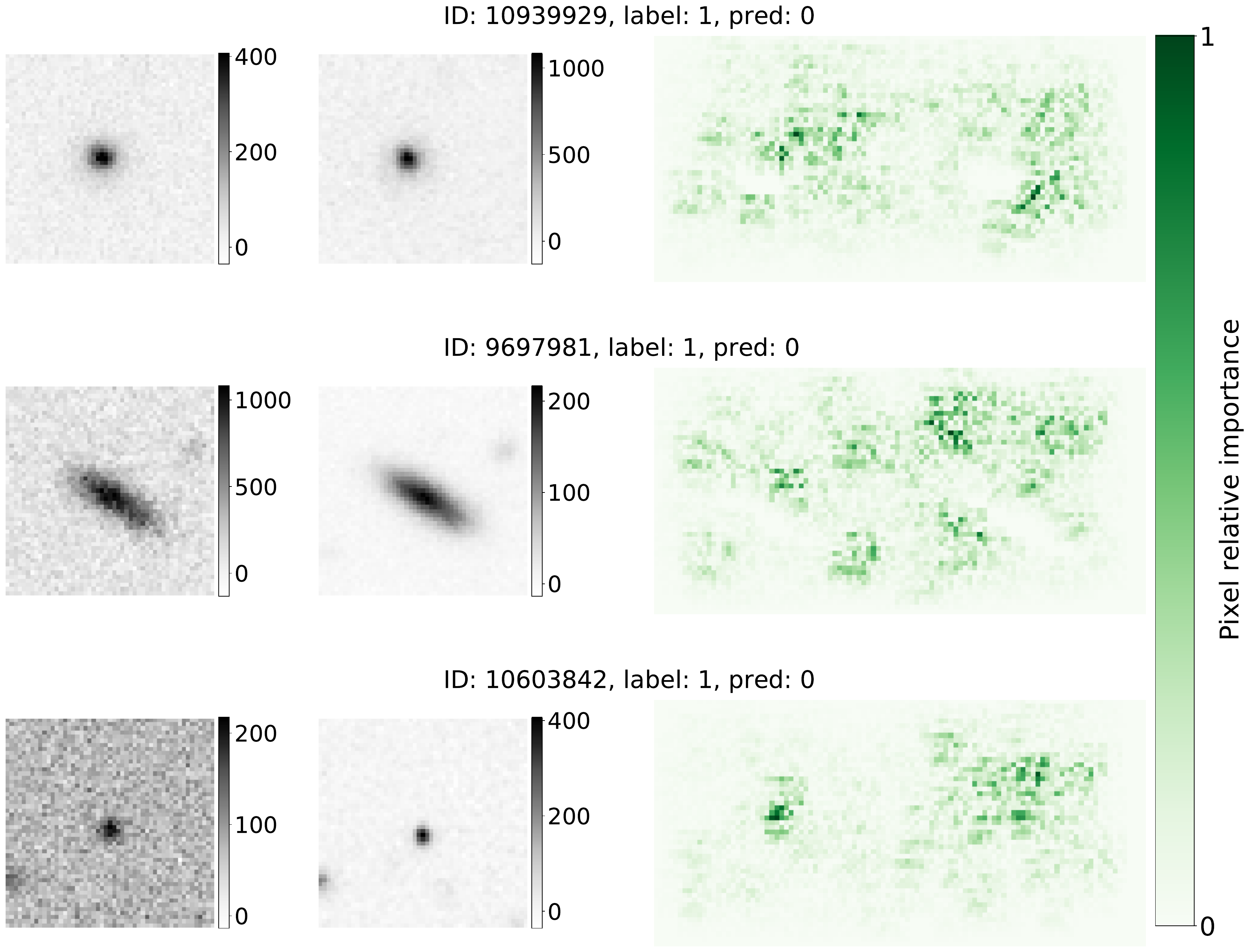}
    \includegraphics[width=0.76\linewidth]{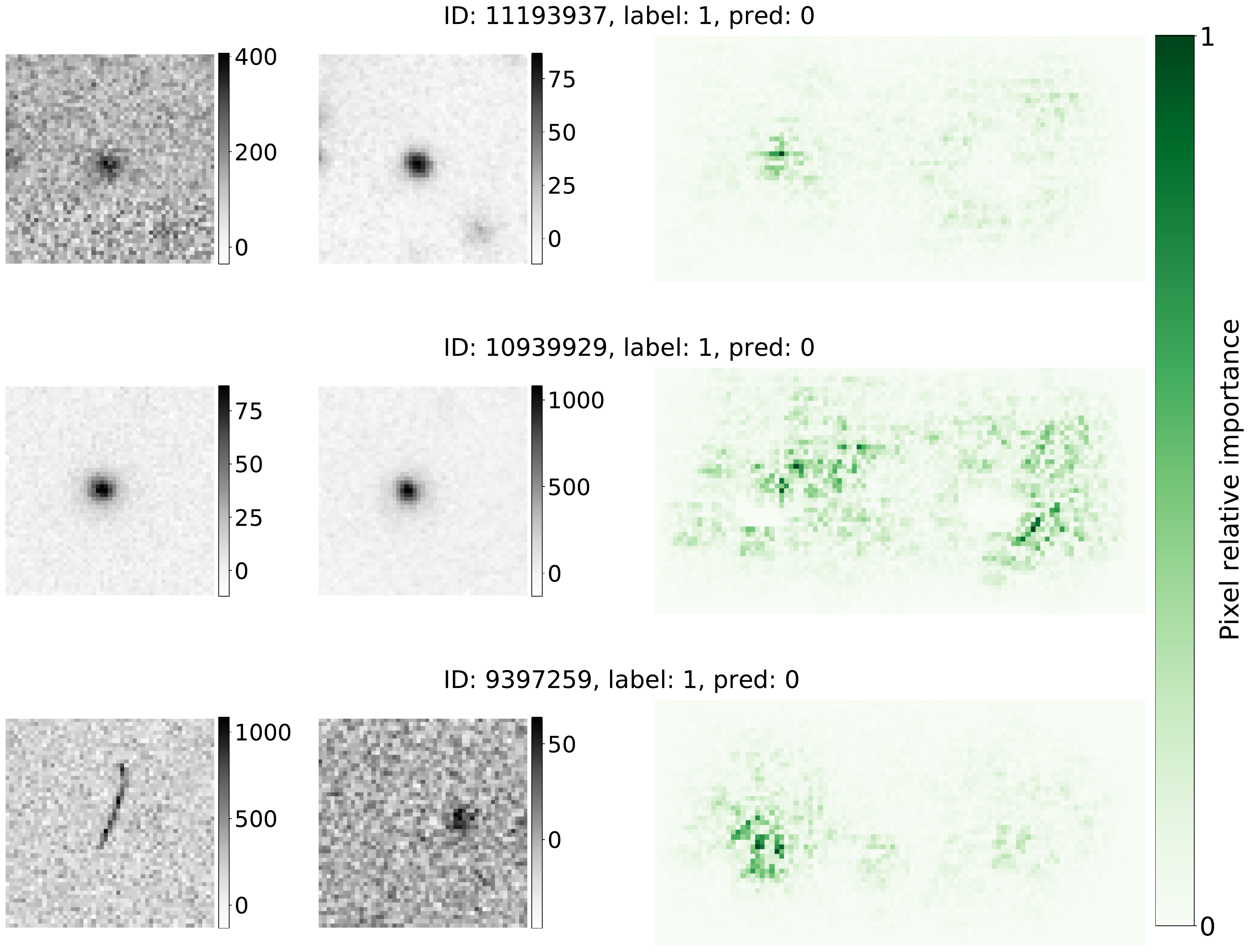}
    \caption{Transients (\search-\temp) and their respective saliency map for the \nodia\ model False Positives (``bogus''  classified as ``real''). Important pixels are found everywhere in the image, as the CNN learns how to compare the \diff\ and \temp\ taking a synoptic look at the properties of each image component.}
    \label{fig:fpndia}
\end{figure*}

We include a series of saliency maps  in the following 8 figures. Several interesting behavioral patterns can be observed. The figures are organized by model and by classification as follows: TN ---\autoref{fig:saliency_3id_contour} and \autoref{fig:saliency_nodia3id_contour}---, TP ---\autoref{fig:saliency_dia3id_TP} and \autoref{fig:tpndia}---, FN \autoref{fig:saliency_dia3id_FN} and \autoref{fig:fnndia}–––, and FP ---\autoref{fig:saliency_dia3id_FP} and \autoref{fig:fpndia}.

In  \autoref{fig:saliency_3id_contour} and \autoref{fig:saliency_nodia3id_contour} we include the transient's contours overplotted onto the saliency maps for the \diabased\ and \nodia\ model, respectively to guide the reader's eye. 
% We note a few intriguing features and speculate on their meaning. In \autoref{fig:saliency_3id_contour}A, a dipole, the important pixels are located off-center, in correspondence to the dark portion of the dipole (althought we note that this classification is incorrect \fed{can you add a dipole with correct classification?}). 
Notice the offset of the \diabased\ model's focus in  \autoref{fig:saliency_3id_contour}A with respect to the transient; the model is principally inspecting the \diff\ and \temp\ at the location corresponding to the bright patch in the \diff.
In  \autoref{fig:saliency_3id_contour}C a correctly predicted ``bogus'' is characterized by a dipole probably arising from a poorly centered DIA, the model inspect mainly the \temp\ and even seem to reproduce traditional aperture photometry, with pixel values measured at the core of the transients, and in the sky surrounding the transient. The behavior is principally different for the same transients when they are inspected, and also correctly predicted, by the \nodia\ model (\autoref{fig:saliency_nodia3id_contour}): the focus of the model is in all three cases away from the transients, and shifted to the surrounding: the model is learning the transients' context and extracting information to enable the comparison of \temp\ and \diff\ (essentially, to enable the image differencing). This behavior is generally seen throughout all examples in \autoref{fig:saliency_dia3id_TP}-\autoref{fig:fpndia}. In addition, for each figure we highlight potential reasons for failed predictions, and potential inaccuracies in the labeling that may lead to an artificial lowering of our measured accuracy.

\clearpage


%% file: appendix_relabel.tex
\round{\section{Visual inspection of Bogus images}

A total of 300 images originally label as bogus were visually inspected by our team, each inspected by 1-5 people. It was found that $\sim 3\%$ of them should be re-labeled as real and a classification disagreement persisted for $\sim 10\%$ of them.} %\autoref{fig:3sure-real} is an example of one of the $\sim 3\%$ visually re-label as real.}

%\begin{figure*}
%    \centering
 %   \includegraphics[width=0.5\linewidth]{
%    figures/10bogusreal.png}
 %   \caption{Object originally label as bogus that is visually label as real.}
  %  \label{fig:3sure-real}
%\end{figure*}

%\round{\autoref{fig:10bogus-real} is an example of one of the $\sim 10\%$ whose classification by visual inspection is undetermined.}

%\begin{figure*}
 %   \centering
    %\includegraphics[width=0.5\linewidth]{
    %figures/3real.png}
    %\caption{Object originally label as bogus that upon visual inspection the classification is undetermined.}
    %\label{fig:10bogus-real}
%\end{figure*}

\begin{figure*}
    \centering
    \includegraphics[width=0.8\linewidth]{
    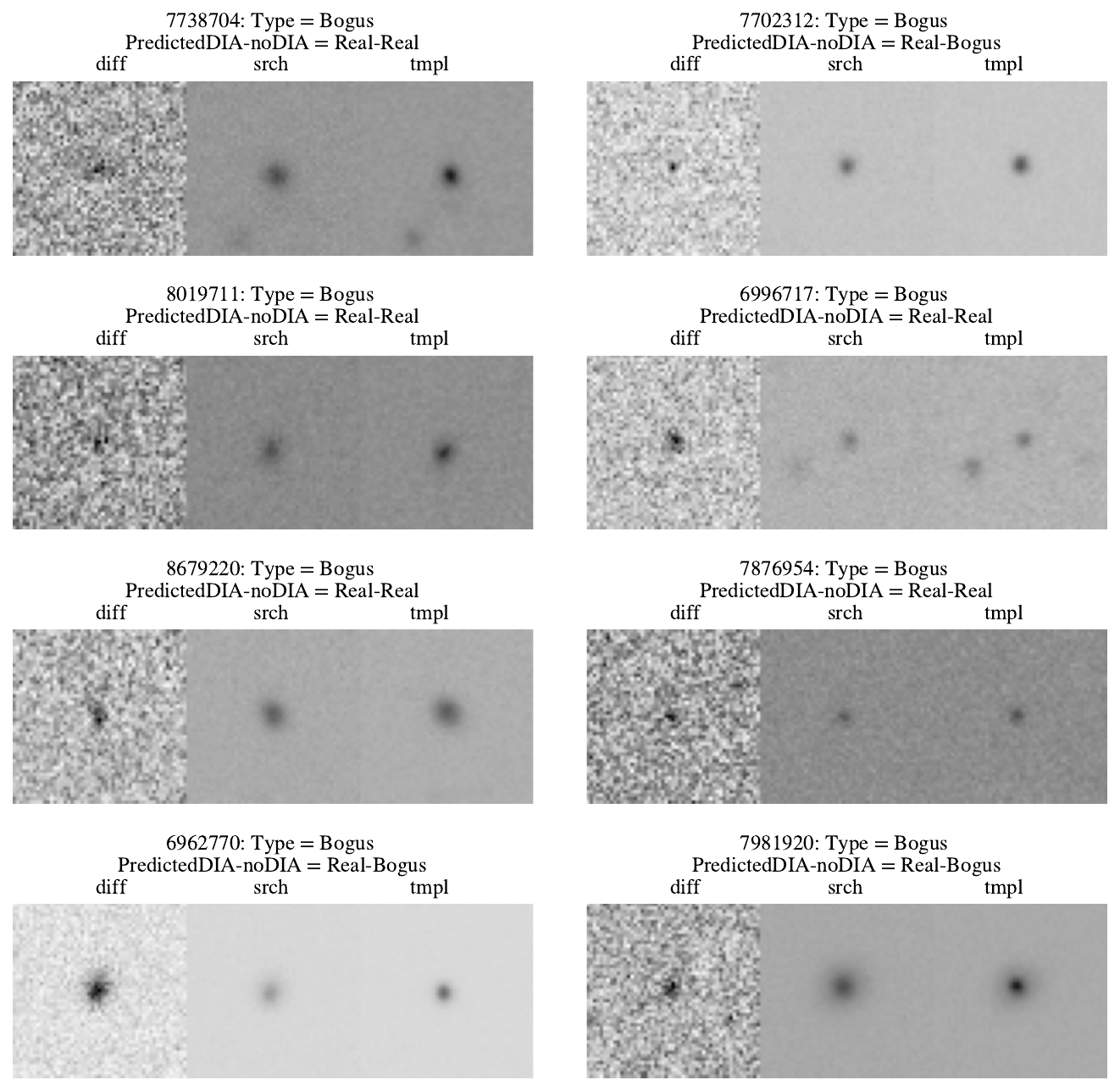}
    \caption{Visual inspection of some objects originally label as bogus that were classify as real for \diabased\ or \nodia.}
    \label{fig:FPvisual}
\end{figure*}

\round{Some of the models' false positives were also visually inspected. A few examples are shown in \autoref{fig:FPvisual}.}
